\documentclass[pdflatex,sn-basic]{sn-jnl}
\usepackage{tcolorbox}
\usepackage{graphicx}%
\usepackage{multirow}%
\usepackage{amsmath,amssymb,amsfonts}%
\usepackage{amsthm}%
\usepackage{mathrsfs}%
\usepackage[title]{appendix}%
\usepackage{xcolor}%
\usepackage{textcomp}%
\usepackage{manyfoot}%
\usepackage{booktabs}%
\usepackage{algorithm}%
\usepackage{pdflscape}
\usepackage{listings}%

\usepackage[british,UKenglish]{babel}
\usepackage{natbib}
\usepackage{hyperref}
\usepackage{array}
\usepackage{bm}
\usepackage[noend]{algpseudocode}
\graphicspath{ {images/} }
\usepackage{xparse}

\usepackage{hyphenat}
\PassOptionsToPackage{hyphens}{url}\usepackage{hyperref}

\definecolor{codegreen}{rgb}{0,0.6,0}
\definecolor{codegray}{rgb}{0.5,0.5,0.5}
\definecolor{codepurple}{rgb}{0.58,0,0.82}
\definecolor{backcolour}{rgb}{0.97,0.97,0.97}

\lstdefinestyle{python_jay}{
    backgroundcolor=\color{backcolour},
    commentstyle=\color{codegreen},
    keywordstyle=\color{blue},
    numberstyle=\tiny\color{codegray},
    stringstyle=\color{codepurple},
    basicstyle=\ttfamily\footnotesize,
    breakatwhitespace=false,
    breaklines=true,
    captionpos=b,
    keepspaces=true,
    numbers=left,
    numbersep=5pt,
    showspaces=false,
    showstringspaces=false,
    showtabs=false,
    tabsize=2
}

\NewDocumentCommand{\rot}{O{45} O{1em} m}{\makebox[#2][l]{\rotatebox{#1}{#3}}}%

\newcolumntype{x}[1]{>{\centering\let\newline\\\arraybackslash\hspace{0pt}}p{#1}}

\providecommand{\tabularnewline}{\\}

\theoremstyle{definition}
\newtheorem{definition}{Definition}

\begin{document}

\title[Bake off redux: a review and experimental evaluation of recent time series classification algorithms]{Bake off redux: a review and experimental evaluation of recent time series classification algorithms}

\author[1]{\fnm{Matthew} \sur{Middlehurst}}\email{m.b.middlehurst@soton.ac.uk}  
\author[2]{\fnm{Patrick} \sur{Sch\"afer}}\email{patrick.schaefer@hu-berlin.de}  
\author*[1,3]{\fnm{Anthony} \sur{Bagnall}}\email{a.j.bagnall@soton.ac.uk}  

\affil[1]{\orgdiv{School of Electronics and Computer Science}, \orgname{University of Southampton}, \orgaddress{\city{Southampton}, \country{United Kingdom}}}
\affil[2]{\orgname{Humboldt-Universit\"at zu Berlin}, \orgaddress{\city{Berlin}, \country{Germany}}}
\affil[3]{\orgdiv{School of Computing Sciences}, \orgname{University of East Anglia}, \orgaddress{\city{Norwich}, \country{United Kingdom}}}

\abstract{In 2017, a research paper~\citep{bagnall17bakeoff} compared 18 Time Series Classification (TSC) algorithms on 85 datasets from the University of California, Riverside (UCR) archive. This study, commonly referred to as a `bake off', identified that only nine algorithms performed significantly better than the Dynamic Time Warping (DTW) and Rotation Forest benchmarks that were used.  The study categorised each algorithm by the type of feature they extract from time series data, forming a taxonomy of five main algorithm types. This categorisation of algorithms alongside the provision of code and accessible results for reproducibility has helped fuel an increase in popularity of the TSC field. Over six years have passed since this bake off, the UCR archive has expanded to 112 datasets and there have been a large number of new algorithms proposed. We revisit the bake off, seeing how each of the proposed categories have advanced since the original publication, and evaluate the performance of newer algorithms against the previous best-of-category using an expanded UCR archive. We extend the taxonomy to include three new categories to reflect recent developments. Alongside the originally proposed distance, interval, shapelet, dictionary and hybrid based algorithms, we compare newer convolution and feature based algorithms as well as deep learning approaches. We introduce 30 classification datasets either recently donated to the archive or reformatted to the TSC format, and use these to further evaluate the best performing algorithm from each category. Overall, we find that two recently proposed algorithms, MultiROCKET+Hydra~\citep{dempster22hydra} and HIVE-COTEv2~\citep{middlehurst21hc2}, perform significantly better than other approaches on both the current and new TSC problems.}


\maketitle

\section{Introduction}

Time series classification (TSC) involves fitting a model from a continuous, ordered sequence of real valued observations (a time series) to a discrete response variable. Time series can be univariate (a single variable observed at each time point) or multivariate (multiple variables observed at each time point). For example, we could treat raw audio signals as a univariate time series in a problem such as classifying whale species from their calls and motion tracking co-ordinate data could be a three-dimensional multivariate time series in a human activity recognition (HAR) task. Where relevant, we distinguish between univariate time series classification (UTSC) and multivariate time series classification (MTSC). The ordering of the series does not have to be in time: we could transform audio into the frequency domain using a discrete Fourier transform or map one dimensional image outlines onto a one dimensional series using radial or linear scanning. Hence, some researchers refer to TSC as data series classification. We retain the term TSC for continuity with past research.

TSC problems arise in a wide variety of domains. Popular TSC archives\footnote{\url{https://timeseriesclassification.com}} contain classification problems using: electroencephalograms; electrocardiograms; HAR and other motion data; image outlines; spectrograms; light curves; audio; traffic and pedestrian levels; electricity usage; electrical penetration graph; lightning tracking; hemodynamics; and simulated data. The huge variation in problem domains characterises TSC research. The initial question when comparing algorithms for TSC is whether we can draw any indicative conclusions on performance across a wide range of problems without any prior knowledge as to the underlying common structure of the data. An  experimental evaluation of time series classification algorithms, which we henceforth refer to as the {\em bake off},  was conducted in 2016 and published in 2017~\citep{bagnall17bakeoff}. This bake off, coupled with a relaunch of time series classification archives~\citep{dau19ucr}, has helped increase the interest in TSC algorithms and applications. Our aim is to summarise the  significant developments since 2017. A new MTSC archive~\citep{ruiz21mtsc} has helped promote research in this field. A variety of new algorithms using different representations, including deep learners~\citep{fawaz19deep}, convolution based algorithms~\citep{dempster20rocket} and hierarchical meta ensembles~\citep{lines18hive}, have been proposed for TSC. Furthermore, the growth in popularity of TSC open source toolkits such as aeon\footnote{\url{https://www.aeon-toolkit.org}} and tslearn\footnote{\url{https://tslearn.readthedocs.io/}} have made comparison and reproduction easier. We extend and encompass recent experimental evaluations (e.g. ~\cite{ruiz21mtsc,bagnall20hivecote1,middlehurst21hc2,fawaz19deep}) to provide insights into the current state of the art in the field and highlight future directions. Our target audience is both researchers interested in extending TSC research and practitioners who have TSC problems. Our contributions can be summarised as follows:

\begin{enumerate}
    \item We describe a range of new algorithms for TSC and place them in the context of those described in the bake off.
    \item We compare performance of the new algorithms on the current UCR archive datasets in a univariate classification bake off redux.
    \item We release $30$ new univariate datasets donated by various researchers through the TSC GitHub repository and compare the best in category on these new datasets.
    \item We analyse the factors that drive performance and discuss the merits of different approaches.
\end{enumerate}

To select algorithms, we use the same criteria as the bake off. Firstly, the algorithm must have been published post bake off in a high quality conference or journal (or be an extension of such an algorithm). Secondly, it must have been evaluated on one of the UCR/UEA dataset releases, or on a subset thereof, with reasoning provided for any datasets that are missing. Thirdly, source code must be available and easily adaptable to the time series machine learning tools we use (i.e. usable or easily wrappable in a Java or Python environment). Further explanation on our tools and reproducing our experiments is available in Appendix~\ref{app:code}. We describe many algorithms which inevitably leads to many acronyms and possible naming confusion. We direct the reader to Table~\ref{tab:algorithms} for a summary of the algorithms used and the associated reference. Section~\ref{sec:definitions} describes the core terminology relating to TSC. Section~\ref{sec:proc} summarises how we conduct experimental evaluations of classifiers. We describe the latest TSC algorithms included in this bake off in Section~\ref{sec:algos}. This section also describes the first set of experiments that link to the previous bake off: for each category of algorithms we compare the latest classifiers with the best in class from~\cite{bagnall17bakeoff}. Section~\ref{sec:results} extends the experimental evaluation to include the new datasets. Section~\ref{Sec:analysis} investigates variation in performance in more detail. Finally, we conclude and discuss future direction in Section~\ref{sec:conc}.

\section{Definitions and  Terminology}
\label{sec:definitions}

We define the number of time series in a collection as $n$, the number of channels/dimensions of any observation as $d$ and length of a series as $m$.

\begin{definition}[Time Series (TS)] A time series $A = \left(a_1, a_2,\dots, a_m \right)$ is an ordered sequence of $m$ data points. We denote the $i$-th value of $A$ by $a_{i}$.
\end{definition}

In the above definition, if every point in $a_{i} \in A$ in the time series represents a single value ($a_{i} \in \mathbb{R}$), the series is a \emph{univariate time series (UTS)}. If each point represents the observation of multiple variables at the same time point (e.g., temperature, humidity, pressure, etc.) then each point itself is a vector $a_{i} \in \mathbb{R}^{d}$ of length $d$, and we call it a \emph{multivariate time series (MTS)}:

\begin{definition}[Multivariate Time Series (MTS)]
A multivariate time series $A=\left(a_1, \ldots ,a_m \right) \in \mathbb{R}^{(d \times m)}$ is a list of $m$ vectors with each $a_i$ being a vector of $d$ channels (sometimes referred to as dimensions).
We denote the $i$-th observation of the $k-th$ channel by the scalar $a_{k,i}\in \mathbb{R}$.
\end{definition}

Note that it is also possible to view a MTS as a set of $d$ time series, since in practice that is often how they are treated. However, the vector model makes it explicit that we assume that the dimensions are aligned, i.e. we assume that all observations in $a_i$ are observed at the same point in time or space.
In the context of supervised learning tasks such as classification, a dataset associates each time series with a label from a predefined set of classes.

\begin{definition}[Dataset]
A dataset $D=(X, Y)=\left(A^{(i)}, y^{(i)}\right)_{i \in [1, \dots, n]}$ is a collection of $n$ time series and a predefined set of discrete class labels $C$.  We denote the size of $D$ by $n$, and the $i^{th}$ instance by series and its label by $y^{(i)} \in C$.
\end{definition}

Many time series classification algorithms make use of subseries of the data.

\begin{definition}[Subseries] A subseries $A_{i,l}$ of a time series $A = (a_1, \dots, a_m)$, with $1 \leq i < i+l \leq m$, is a series of length $l$, consisting of the $l$ contiguous points from $A$ starting at offset i: $A_{i,l} =(a_i,a_{i+1},\dots,a_{i+l-1})$, i.e. all indices in the right-open interval $[i,i+l)$.
\end{definition}

We may extract subseries from a time series by the use of a sliding window.

\begin{definition}[Sliding Window]
A time series $A$ of length $m$ has $(m-l+1)$ sliding windows of length $l$ (when increment is $1$) given by: $$sliding\_windows(A)=\{A_{1,l},\dots,A_{(m-l+1),l}\}$$
\end{definition}

A common operation is the \emph{convolution operation}. A kernel (filter) is convolved with a time series through a sliding dot product.

\begin{definition}[Convolution (cross-correlation)]\label{def:convolution}
The result of applying a kernel $\omega$ of length $l$ to a given time series $A$ at position $i$ is given by:
$$M_i = (A_{i,l} * \omega) = \sum_{j=0}^{l-1} A_{i+j} \cdot w_j$$
\end{definition}

\begin{figure}[t]
    \centering
    \includegraphics[width=1\linewidth]{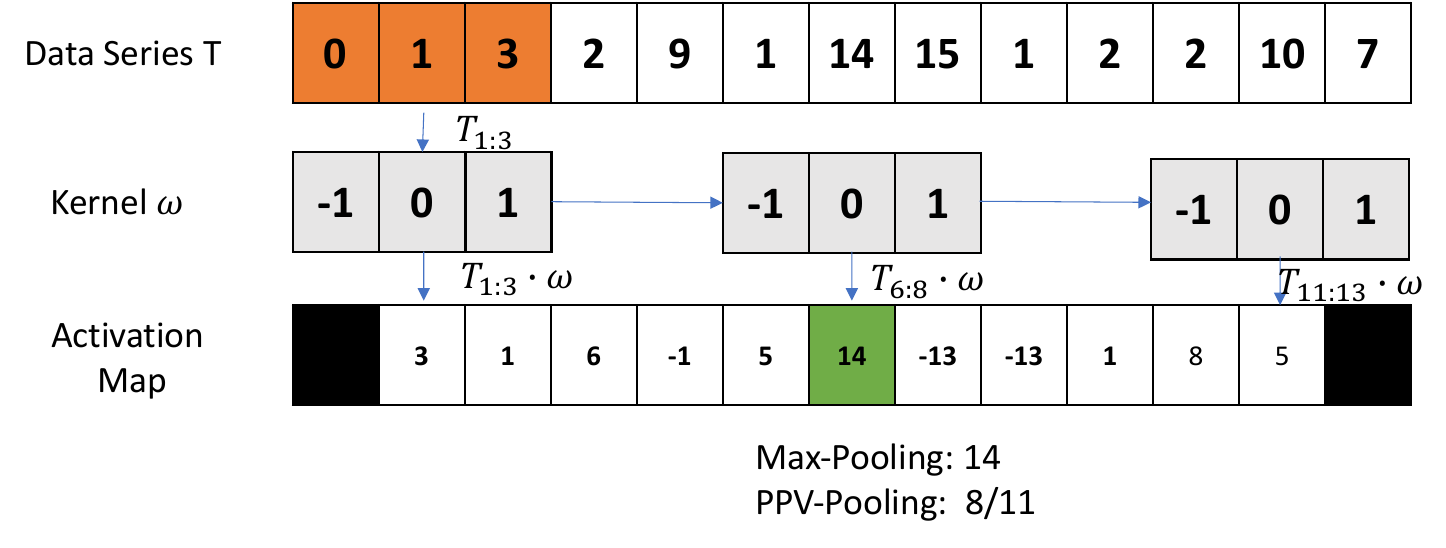}
    \caption{The convolution operation as a sliding dot-product. The kernel $\omega=[-1,0,1]$ is convolved with the input series, producing an activation map. Max-pooling extracts the maximum from this activation map.}
    \label{fig:rocket_convolution}
\end{figure}
The result of the operation is an activation map $M$. Figure~\ref{fig:rocket_convolution} shows the convolution for an input time series and kernel $\omega=[-1,0,1]$. The first entry of the activation map $M$ is the result of a dot-product between $A_{1:3} * \omega = A_{1:3} \cdot \omega = 0 + 0 + 3 = 3$. Each convolution creates a series to series transform from time series to activation map. Activation maps are used to create summary features.

The \emph{dilation technique} is a method that enables a filter, such as a sliding window or kernel, to cover a larger portion of the time series data by creating empty spaces between the entries in the filter. These spaces enable the filter to widen its receptive field while maintaining the total number of values constant. To illustrate, a dilation of $d=2$ would introduce a gap of $1$ between each pair of values. This effectively doubles the receptive field's size and enables the filter to analyse the data at various scales, akin to a down-sampling operation.

\begin{definition}[Dilated Subseries] A dilated subseries, denoted by $A_{i, l, d}$, is a sequence extracted from a time series $A = (a_1, \dots, a_n)$, with $1 \leq i < i + l \times d \leq m$. This subseries has  length $l$ and dilation factor $d$, and it includes $l$ non-contiguous points from $A$ starting at offset $i$ and taking every $d$-th value as follows:
$$A_{i,l,d} =(a_i,a_{i+d \times 1},\dots,a_{i + d \times(l-1)})$$
\end{definition}

The dilation technique is used in convolution-based, shapelet-based and dictionary-based models.

In real-world applications, series of $D$ are often unequal length. This is often treated as a preprocessing task, i.e. by appending tailing zeros, although some algorithms have the capability to internally handle this. We further typically assume that all time series of $D$ have the same sampling frequency, i.e., every $i^{th}$ data point of every series was measured at the same temporal distance from its predecessor.

\section{Experimental Procedure}
\label{sec:proc}

The bake off conducted experiments with the 85 UTSC datasets that were in the UCR archive relaunch of 2015. Each dataset was resampled 100 times for training and testing, and test accuracy was averaged over resamples. The evaluation began with 11 standard classifiers (such as Random Forest~\citep{breiman01randomforest}), then classifiers in each category were compared, including an evaluation of reproducibility. Finally, the best in class were compared to hybrids (combinations of categories).

We adapt this approach for the bake off redux to reflect the progression of the field. First, we take the previously used benchmark of Dynamic Time Warping using a one nearest neighbour classifier (1-NN DTW) and, if appropriate, the best of each category from the bake off and compare them to new algorithms of that type.
We do this stage of experimentation with the 112 equal length problems in the 2019 version of the UCR archive~\citep{dau19ucr}. Performance on these datasets, or some subset thereof, has been used to support every proposed approach, so this allows us to make a fair comparison of algorithms. We have regenerated all results for classifiers described both in the original bake off and this comparison.

Only a subset of the algorithms considered have been adapted for MTSC by their inventors. Furthermore, many algorithms have been proposed solely for MTSC, particularly in the deep learning field. Because of this and the considerable computational cost of including multivariate data, we restrict our attention to univariate classification only in this work.

We resample each pair of train/test data $30$ times for the redux, stratifying to retain the same class distribution. We do not adopt the bake off strategy of 100 resamples. We have found 30 resamples is sufficient to mitigate small changes in test accuracy over influencing ranks, and it is more computationally feasible. Resampling is seeded with the resample ID to aid with reproducibility. Resample 0 uses the original train and test split from the UCR archive.

Our primary performance measure is classification accuracy on the test set. We also compare predictive power with the balanced test set accuracy, to identify whether class imbalance is a problem for an algorithm. The quality of the probability estimates is measured with the negative log likelihood (NLL), also known as log loss. The ability to rank predictions is estimated by the area under the receiver operator characteristic curve (AUROC). For problems with two classes, we treat the minority class as a positive outcome. For multiclass problems, we calculate the AUROC for each class and weight it by the class frequency in the train data, as recommended in~\cite{provost03pet}. We present results with diagrams derived from the critical difference plots proposed by~\cite{demsar06comparisons}. We average ranks over all datasets and plot them on a line and group classifiers into cliques, within which there is no significant difference in rank. We replace the post-hoc Nemenyi test used to form cliques described in~\cite{demsar06comparisons} with a mechanism built on pairwise tests. We perform pairwise one-sided Wilcoxon signed-rank tests and form cliques using the Holm correction for multiple testing as described in~\cite{garcia08pairwise,benavoli16pairwise}.

Critical difference diagrams can be deceptive: they do not display the effective amount of differences, and the linear nature of clique finding can mask relationships between results. If, for example, three classifiers $A, B, C$ are ordered by rank $A>B>C$, and the test indicates $A$ is significantly better than $B$, and $B$ is significantly better than $C$, then we will form no cliques. However, it is entirely possible that $A$ is not significantly different to $C$, and the diagram cannot display this. Because of this, we expand our results to include pairwise plots, violin plots of accuracy distributions against a base line, tables of test accuracies and heatmap diagrams which include unadjusted p-values~\citep{ismail23multiple}.

\subsection{New Datasets}
\label{sec:new_data}

The 112 equal length TSC problems in the archive constitute a relatively large corpus of problems for comparing classifiers. However, they have been extensively used in algorithm development, and there is always the risk of an implicit overfitting resulting in conclusions that do not generalise well to new problems. Hence, we have gathered new datasets which we use to perform our final comparison of algorithms. These data come from direct donation to the TSC GitHub repository\footnote{\url{https://github.com/time-series-machine-learning/tsml-repo}}, discretised regression datasets\footnote{\url{http://tseregression.org/}}, a project on audio classification~\citep{flynn19classifying} and reformatting current datasets with unequal length or missing values. Submissions of new datasets to the associated repository are welcomed.

\begin{figure}[!ht]
    \centering
    \includegraphics[width=0.9\linewidth]{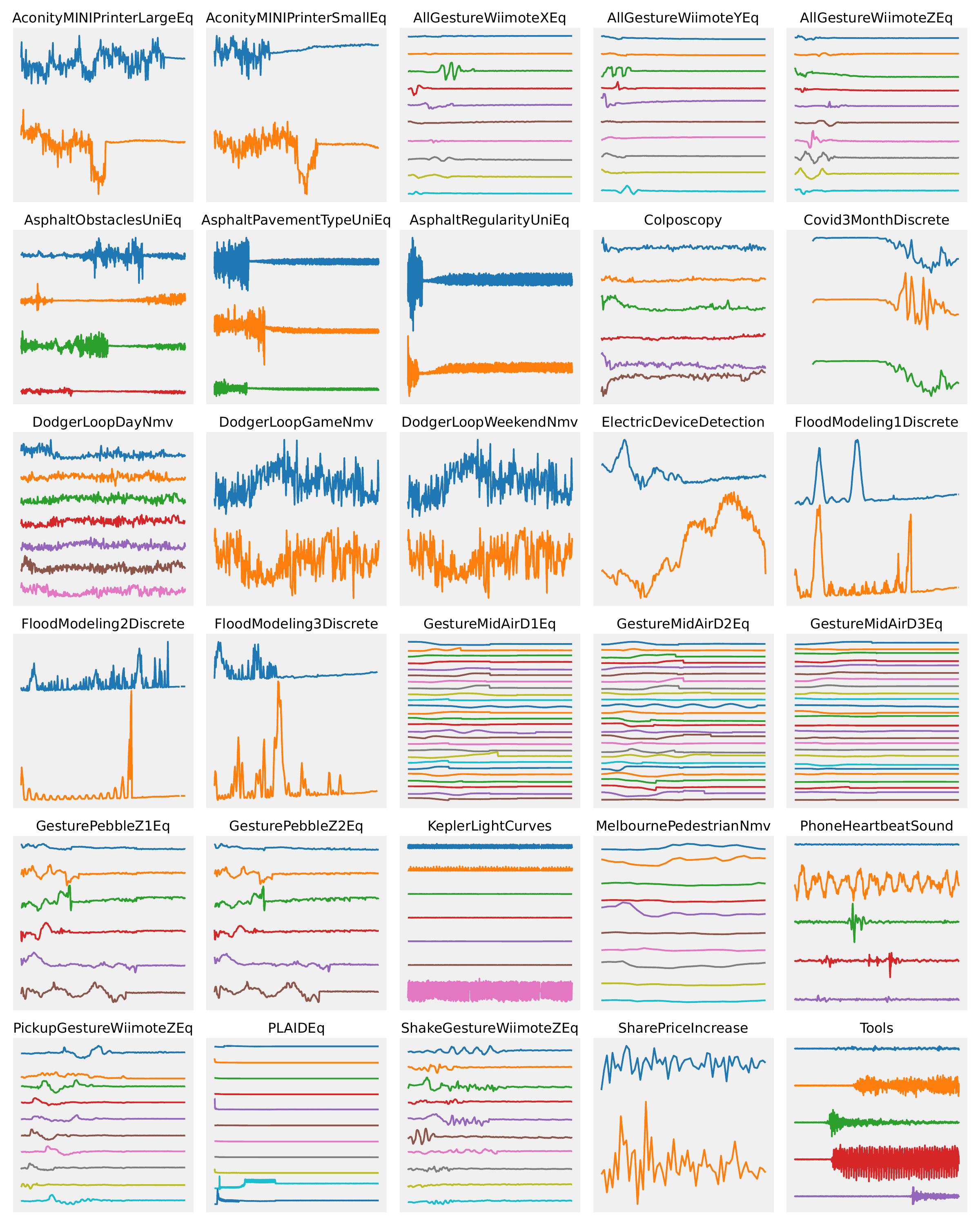}
    \caption{The $30$ new univariate datasets showing one representative series for each class.}
    \label{fig:new_datasets}
\end{figure}

In total, we have gathered 30 new datasets, summarised in Table~\ref{tab:new_data} and visualised in Figure~\ref{fig:new_datasets}. Datasets with the suffix \textbf{\_eq} are unequal length series made equal length through padding with the series mean perturbed by low level Gaussian noise. 11 of these problems (AllGestureWiimote versions, GestureMidAirD1, GesturePebbleZ, PickupGestureWiimoteZ, PLAID and ShakeGestureWiimoteZ) are already in the archive so need no further explanation.

\begin{table}[]
    \caption{A summary of the 30 new univariate datasets used in our experiments with suffix: \emph{\_eq, \_nmv, \_disc}}
    \label{tab:new_data}
    \begin{tabular}{m{0.32\linewidth} | x{0.08\linewidth} x{0.08\linewidth} x{0.08\linewidth} x{0.08\linewidth} x{0.14\linewidth}}
        \hline
        Dataset & Train size & Test size & Series length & No. Classes & Category \\
        \hline
        AconityMINIPrinterLarge\_eq & 2403 & 1184 & 300 & 2 & Sensor \\
        AconityMINIPrinterSmall\_eq & 589 & 292 & 300 & 2 & Sensor \\
        AllGestureWiimoteX\_eq & 300 & 700 & 500 & 10 & Motion \\
        AllGestureWiimoteY\_eq & 300 & 700 & 500 & 10 & Motion \\
        AllGestureWiimoteZ\_eq & 300 & 700 & 500 & 10 & Motion \\
        AsphaltObstaclesUni\_eq & 390 & 391 & 736 & 4 & Sensor \\
        AsphaltPavementTypeUni\_eq & 1055 & 1056 & 2371 & 3 & Sensor \\
        AsphaltRegularityUni\_eq & 751 & 751 & 4201 & 2 & Sensor \\
        Colposcopy & 99 & 101 & 180 & 6 & Image \\
        Covid3Month\_disc & 140 & 61 & 84 & 3 & Other \\
        DodgerLoopDay\_nmv & 67 & 77 & 288 & 7 & Sensor \\
        DodgerLoopGame\_nmv & 17 & 127 & 288 & 2 & Sensor \\
        DodgerLoopWeekend\_nmv & 18 & 126 & 288 & 2 & Sensor \\
        ElectricDeviceDetection & 624 & 3768 & 256 & 2 & Image \\
        FloodModeling1\_disc & 471 & 202 & 266 & 2 & Simulated \\
        FloodModeling2\_disc & 466 & 201 & 266 & 2 & Simulated \\
        FloodModeling3\_disc & 429 & 184 & 266 & 2 & Simulated \\
        GestureMidAirD1\_eq & 208 & 130 & 360 & 26 & Motion \\
        GestureMidAirD2\_eq & 208 & 130 & 360 & 26 & Motion \\
        GestureMidAirD3\_eq & 208 & 130 & 360 & 26 & Motion \\
        GesturePebbleZ1\_eq & 132 & 172 & 455 & 6 & Motion \\
        GesturePebbleZ2\_eq & 146 & 158 & 455 & 6 & Motion \\
        KeplerLightCurves & 920 & 399 & 4767 & 7 & Sensor \\
        MelbournePedestrian\_nmv & 1138 & 2319 & 24 & 10 & Sensor \\
        PhoneHeartbeatSound & 424 & 182 & 3053 & 5 & Other \\
        PickupGestureWiimoteZ\_eq & 50 & 50 & 361 & 10 & Motion \\
        PLAID\_eq & 537 & 537 & 1345 & 11 & Device \\
        ShakeGestureWiimoteZ\_eq & 50 & 50 & 385 & 10 & Motion \\
        SharePriceIncrease & 965 & 966 & 60 & 2 & Other \\
        Tools & 310 & 134 & 2926 & 5 & Other \\
        \hline
    \end{tabular}
\end{table}

Four problems with the suffix \textbf{\_nmv} (no missing values) are datasets where the original contains missing values. These are also from the current archive. We have used the simplest method for processing the data, and removed any cases which contain missing values for these problems (DodgerLoop variants and MelbournePedestrian). The number of cases removed per dataset amounts to 5-15\% of the original size for all four datasets which we deemed acceptable. While there have been imputation methods proposed for time series, the amount of missing values present and their pattern varies. The DodgerLoop datasets have large strings of missing values, while MelbournePedestrian has singular values or small groupings of missing data.

\begin{figure}[t]
	\centering
    \begin{tabular}{cc}
       \includegraphics[width=0.465\linewidth, trim={0cm 0cm 1.6cm 0.5cm},clip]{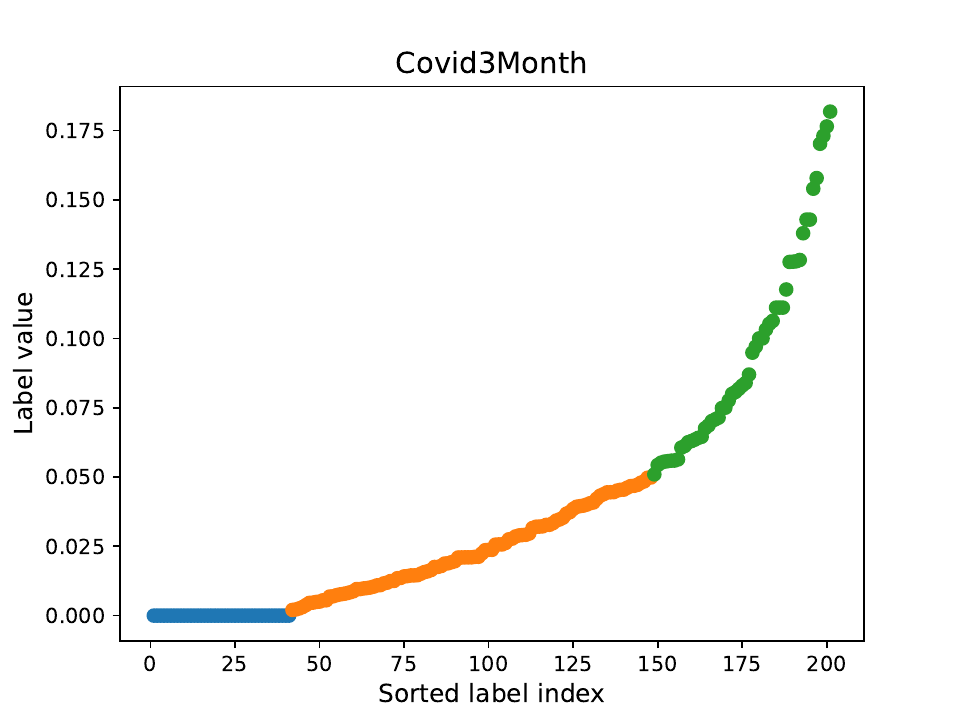} &
       \includegraphics[width=0.465\linewidth, trim={0cm 0cm 1.6cm 0.5cm},clip]{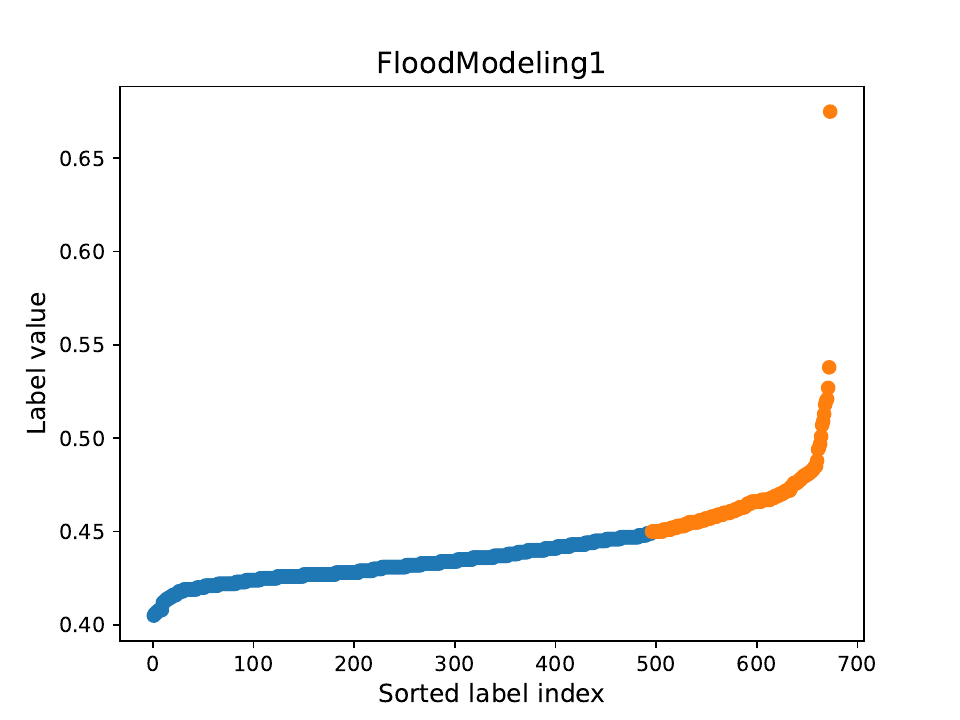}  \\
        \includegraphics[width=0.465\linewidth, trim={0cm 0cm 1.6cm 0.5cm},clip]{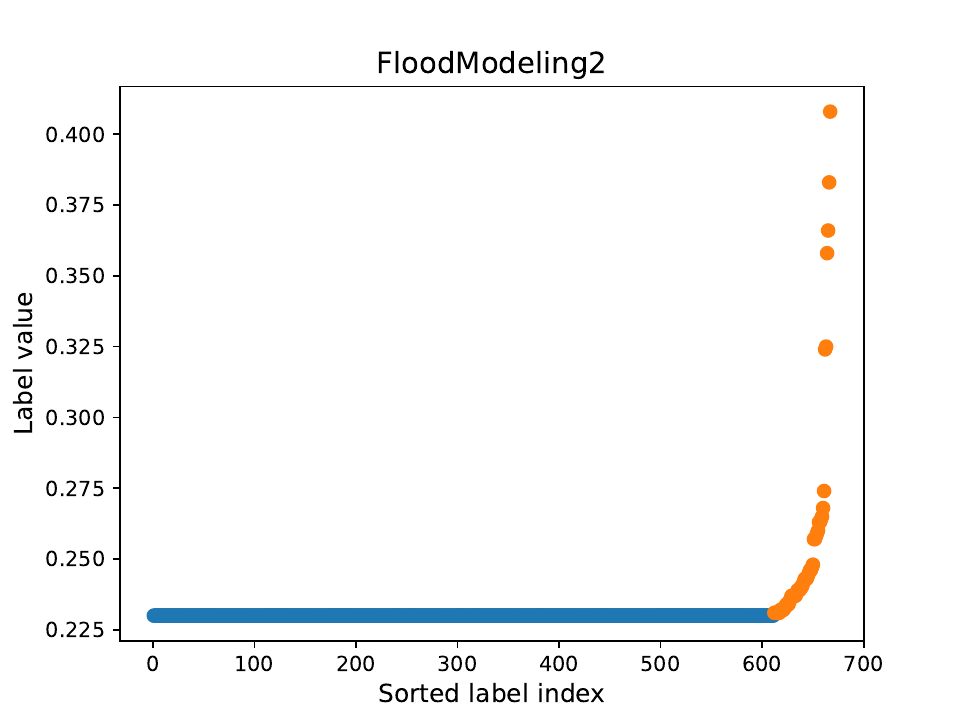} &
       \includegraphics[width=0.465\linewidth, trim={0cm 0cm 1.6cm 0.5cm},clip]{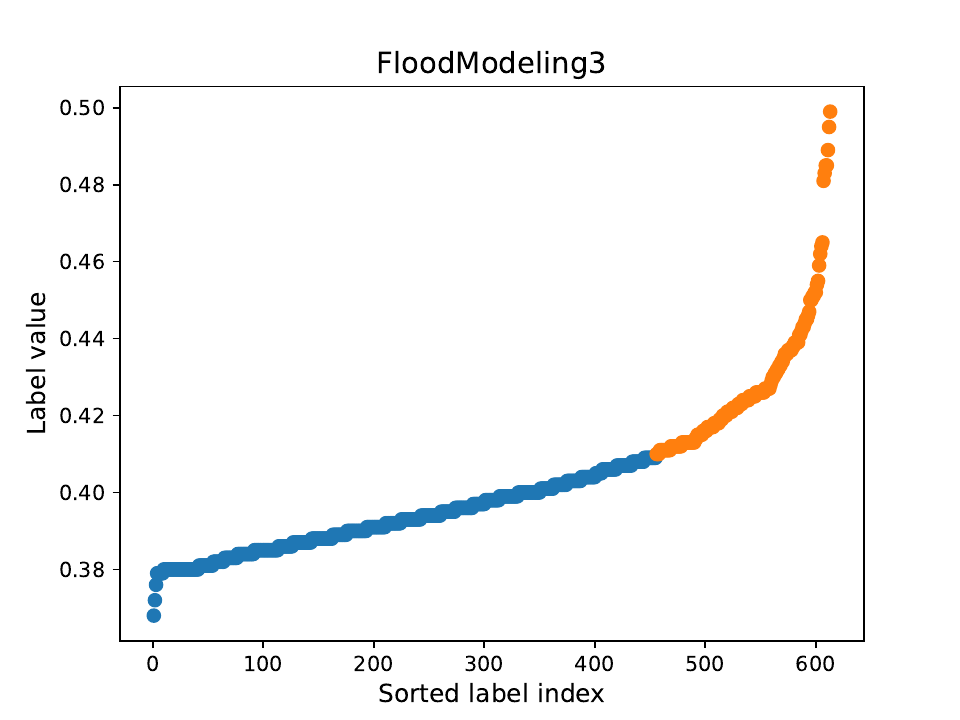}  \\
       \end{tabular}
       \caption{The sorted original label values for all discretised regression datasets. Each point is a label for a case, and its colour is the class it is part of for the new classification version.}
       \label{fig:regression_data}
\end{figure}

The four datasets ending with \textbf{\_disc} are taken from the TSER archive~\citep{tan21regression}. The continuous response variable was discretised manually for each dataset, the original continuous labels and new class values for each dataset are shown in Figure~\ref{fig:regression_data}. Both Covid3Month and FloodModeling2 had a minimum label value with many cases. For both of these, this minimum label value has been converted into its own class label. For problems where there are no obvious places where the label can be separated into classes by value (including Covid3Month where the value is greater than 0), a split point was manually selected taking into account the average label value and the number of cases in each class for a splitting point.

This leaves 11 datasets that are completely new to the archive. The two \textbf{AconityMINIPrinter} data sets are described in~\citep{mahato20printing} and donated by the authors of that paper. The data
 comes from the AconityMINI 3D printer during the manufacturing of stainless steel blocks with a designed cavity. The problem is to predict whether there is a void in the output of the printer. The time series are temperature data that comes from pyrometers that monitor melt pool temperature. The pyrometers track the scan of the laser to provide a time-series sampled at 100 Hz. The data is sampled from the mid-section of these blocks and is organized into two datasets (large and small). The large dataset covers cubes with large pores (0.4 mm, 0.5 mm, and 0.6 mm) and the small dataset covers cubes with small pores (0.05 mm and 0.1 mm).

The three \textbf{Asphalt} datasets were originally described in~\citep{souza18asphalt} and donated by the author of that paper. Accelerometer data was collected on a smartphone installed inside a vehicle using a flexible suction holder near the dashboard. The acceleration forces are given by the accelerometer sensor of the device and are the data used for the classification task. The class values for AsphaltObstacles classes are four common obstacles in the region of data collection: raised cross walk (160 cases); raised markers (187 cases); speed bump (212 cases); and vertical patch (222 cases); flexible pavement (816 cases); cobblestone street (527 cases); and dirt road (768 cases). AsphaltRegularity is a two class problem: Regular (762 cases), where the asphalt is even and the driver's comfort changes little over time; and Deteriorated (740 cases), where irregularities and unevenness in a damaged road surface are responsible for transmitting vibrations to the interior of the vehicle and affecting the driver's comfort.

The \textbf{Colposcopy} data is described in~\citep{gutierrez17colposcopy} and was donated to the repository by the authors\footnote{\url{https://github.com/KarinaGF/ColposcopyData}}. The task is to classify the nature of a diagnosis from a colposcopy. The time series represent the change in intensity values of a pixel region through a sequence of digital colposcopic images obtained during the colposcopy test that was performed on each patient included in the study.

The \textbf{ElectricDeviceDetection} data set~\citep{bagnall20bags} contains formatted image data for the problem of detecting whether a segment of a 3-D X-Ray contains an electric device or not. The data originates from an unsupervised segmentation of 3-D X-Rays. The data are histograms of intensities, not time series.

\textbf{KeplerLightCurves} was described in~\citep{barbara22kepler} and donated by the authors. Each case is a light curve (brightness of an object sampled over time) from NASA's Kepler mission (3-month-long series, sampled every 30 min). There are seven classes relating to the nature of the observed star.

The \textbf{SharePriceIncrease} data was formatted by Vladislavs Pazenuks as part of their 2018 undergraduate student project. The problem is to predict whether a share price will show an exceptional rise after quarterly announcement of the Earning Per Share based on the price movement of that share price on the preceding 60 days. Daily price data on NASDAQ 100 companies was extracted from a Kaggle data set\footnote{\url{https://www.kaggle.com/code/jacksoncrow/download-nasdaq-historical-data}}. Each data represents the percentage change of the closing price from the day before. Each case is a series of 60 days data. The target class is defined as $0$ if the price did not increase after company report release by more than five percent or $1$ else-wise.

\textbf{PhoneHeartbeatSound} and \textbf{Tools} are audio datasets. Tools contains the sound of a chainsaw, drill, hammer, horn and sword, with the task being to match which tool the audio belongs to. PhoneHeartbeatSound contains sounds of the heartbeats recorded on a phone using a digital stethoscope gathered for the 2011 PASCAL classifying heart sounds challenge\footnote{\url{http://www.peterjbentley.com/heartchallenge/index.html}}. The time series represent the change in amplitude over time during an examination of patients suffering from common arrhythmias. The classes are Artifact (40 cases), ExtraStole (46 cases), Murmur (129 cases), Normal (351 cases) and ExtraHLS (40 cases).

\begin{figure}[t]
	\centering
    \includegraphics[height=0.70\linewidth]{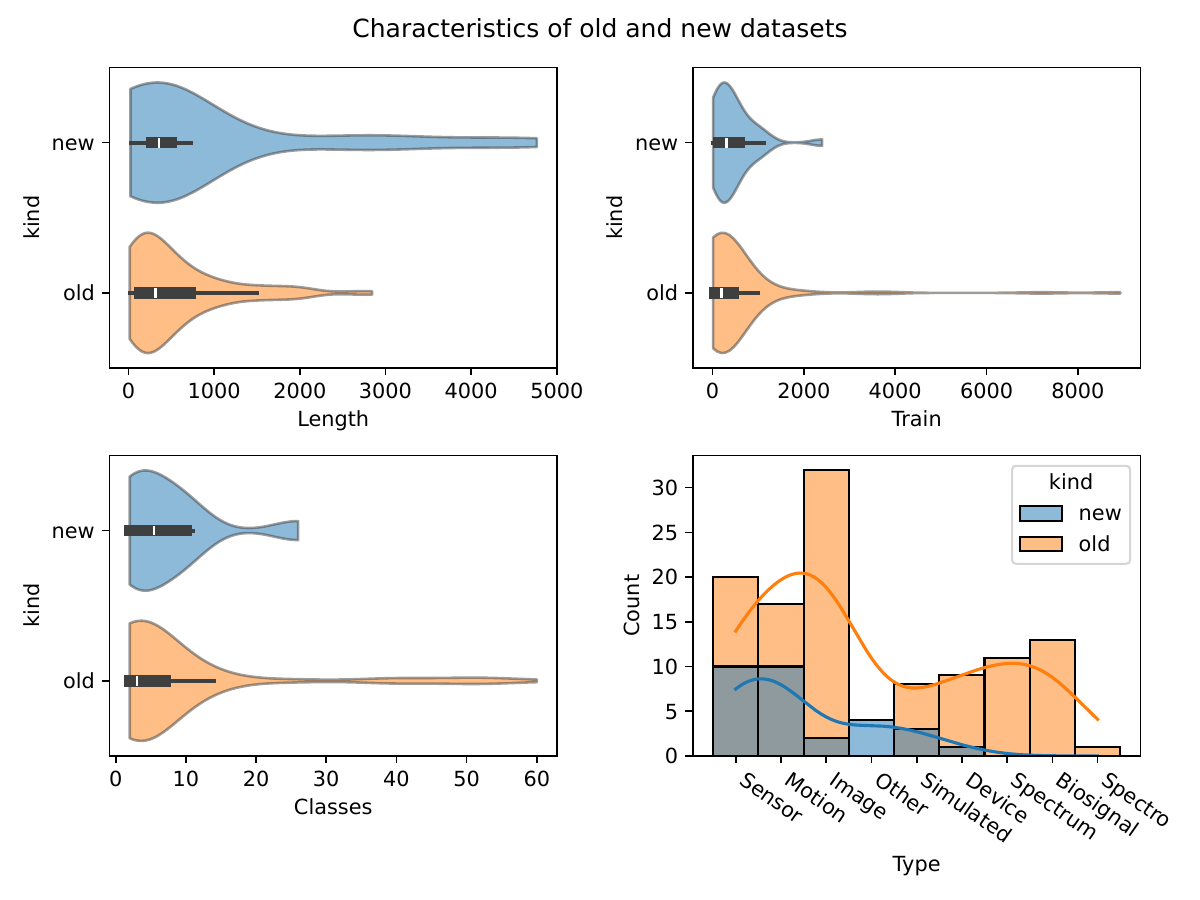}
    \caption{Comparison of distribution of the $30$ newly acquired to the existing $112$ UCR UTSC datasets across dimensions including length, train set size, number of classes, and data type.}
    \label{fig:dataset_comparison}
\end{figure}

Figure~\ref{fig:dataset_comparison} shows the characteristics of the $30$ new datasets when compared to the existing $112$ UCR UTSC datasets, across different dimensions including length, train set size, number of classes, and data type. Findings reveal that the new datasets exhibit a broader range of lengths compared to old ones, while showing similar train set size and similar number of classes. It is worth noting that there seems to be a slight bias towards datasets derived from sensor and motion data in the new collection, whereas the majority of older datasets are sourced from the domain of image outlines.

\subsection{Reproducibility}

The majority of the classifiers described are available in the aeon time series machine learning toolkit (see Footnote 2) and all datasets are available for download (see Footnote 1). Appendix~\ref{app:code} gives detailed code examples on how to reproduce these experiments, including parameters used, if they differ from the default. All results are available from the TSC website and can be directly loaded from there using aeon. Further guidance on reproducibility, parameterisation of the algorithms used in our experiments and our results files are available in an accompanying webpage\footnote{\url{https://tsml-eval.readthedocs.io/en/latest/publications/2023/tsc_bakeoff/tsc_bakeoff_2023.html}}. With the exception of three algorithms which only meet our usage criteria with a Java implementation, all the algorithms used in our experiments are runnable using the Python software and guides linked in the webpage.

\section{Time Series Classification Algorithms}
\label{sec:algos}

The bake off introduced a taxonomy of algorithms based on the representation of the data at the heart of the algorithm. TSC algorithms were classified as either whole series, interval based, shapelet based, dictionary based, combinations or model based. We extend and refine this taxonomy to reflect recent developments.

\begin{enumerate}
    \item Distance based: classification is based on some time series specific distance measure between whole series (Section~\ref{sec:distance}).
    \item Feature based: global features are extracted and passed to a standard classifier in a simple pipeline (Section~\ref{sec:features}).
    \item Interval based: features are derived from selected phase dependent intervals in an ensemble of pipelines (Section~\ref{sec:intervals}).
    \item Shapelet based: phase independent discriminatory subseries form the basis for classification (Section~\ref{sec:shapelets}).
    \item Dictionary based: histograms of counts of repeating patterns are the features for a classifier (Section~\ref{sec:dictionary}).
    \item Convolution based: convolutions and pooling operations create the feature space for classification (Section~\ref{sec:conv}).
    \item Deep learning based: neural network based classification (Section~\ref{sec:deep}).
    \item Hybrid approaches combine two or more of the above approaches (Section~\ref{sec:hybrid}).
\end{enumerate}

As well as the type of feature extracted, another defining characteristic is the design of the TSC algorithm. The simplest design pattern involves single pipelines where transformation of the series into discriminatory features is followed by the application of a standard machine learning classifier. These algorithms tend to involve an over-production and selection strategy: a large number of features are created, and the classifier determines which features are most useful. The transform can remove time dependency, e.g. by calculating summary features. We call this type series-to-vector transformations. Alternatively, they may be series-to-series, transforming into an alternative time series representation where we hope the task becomes more easily tractable, e.g. transforming to the frequency domain of the series.

The second transformation based design pattern involves ensembles of pipelines, where each base pipeline consists of making repeated, different, transforms and using a homogeneous base classifier. TSC ensembles can also be heterogeneous, collating the classifications from transformation pipelines and ensembles of differing representations of the time series.

The third common pattern involves transformations embedded inside a classifier structure. For example, a decision tree where the data is transformed at each node fits this pattern.

A common theme to all categories of algorithm is ensembling. Another popular method seen in multiple classifiers are transformation pipelines ending with a linear classifier. The most accurate classifiers we find all form homogeneous or heterogeneous ensembles, or extract features prior to a linear ridge classifier.

To try and capture the commonality and differences between algorithms we provide a Table in the Appendix~\ref{app:algorithms} (Table~\ref{tab:algorithm_characteristics}) that groups algorithms by whether they employ the following design characteristics: dilation; discretisation; differences/derivatives; frequency domain; ensemble; and linear classification.

We review each category of algorithms by providing an overview of the approach, review selected classifiers and describe the pattern they use, starting with the best of class from the bake off. We perform a comparison of performance within category on the 112 equal length UTSC problems currently in the UCR archive using 1-NN DTW as a benchmark. More detailed evaluation is delayed until Section~\ref{sec:results}.

\subsection{Distance Based}\label{sec:distance}

\begin{figure}[tb]
    \centering
    \includegraphics[width=1\linewidth]{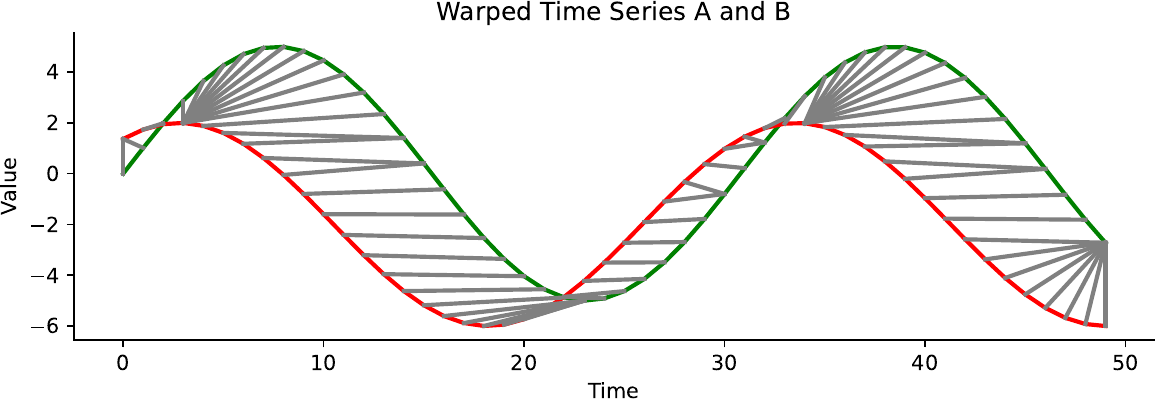}
    \caption{An example of how DTW compensates for phase shift by realigning two series (in red at the bottom and in green at the top).}
    \label{fig:dtw}
\end{figure}

\textit{Distance based} classifiers use a distance function to measure the similarity between whole time series. Historically, distance functions have been mostly used with nearest neighbour (NN) classifiers. Alternative uses of time series distances are described in~\cite{abanda19distance}. Prior to the bake off, 1-NN with DTW was considered state of the art for TSC~\citep{rakthanmanon13trillions}. Figure~\ref{fig:dtw} shows an example of how DTW attempts to align two series, depicted in red and green, to minimise their distance.

In addition to DTW, a wide range of alternative elastic distance measures (distance measures that compensate for possible misalignment between series) have been proposed. These use combinations of warping and editing on series and the derivatives of series. See~\cite{holder22clustering} for an overview of elastic distances. Previous studies~\citep{lines14elastic} have shown there is little difference in performance between 1-NN classifiers with different elastic distances.

\begin{figure}[tb]
    \centering
    \includegraphics[width=.8\linewidth,trim={2cm 3cm 2cm 2cm},clip]{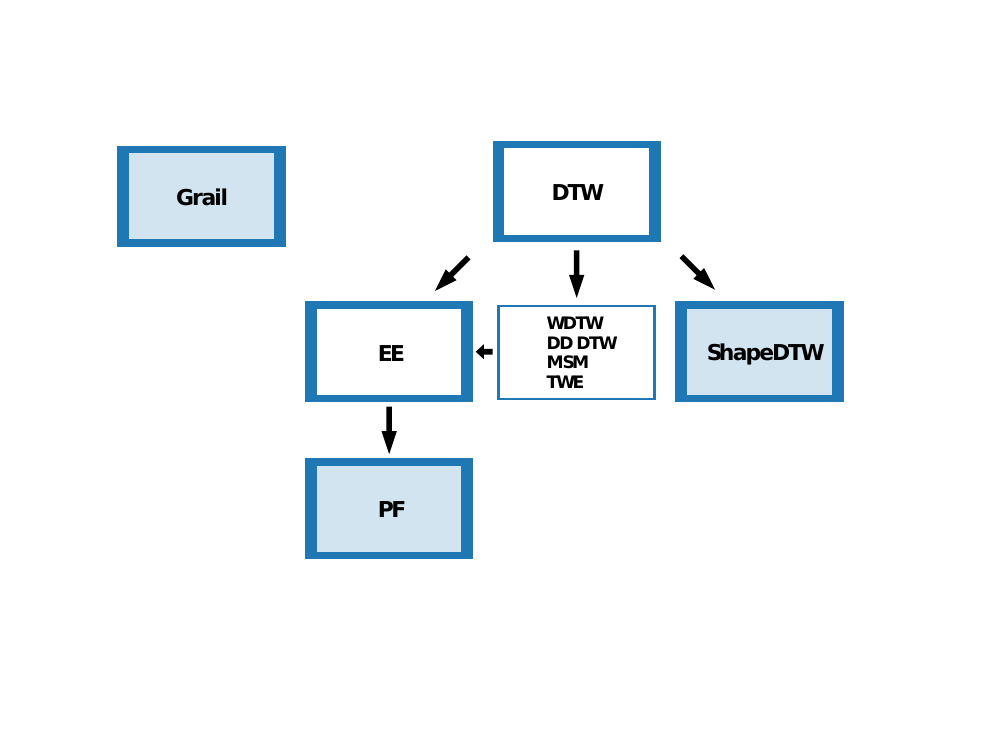}
    \caption{An overview of distance based classifiers and the relationship between them. Filled algorithms were released after the 2017 bake off~\citep{bagnall17bakeoff} and algorithms with a thin border are not included in our experiments.}
    \label{fig:distance_flow}
\end{figure}

The flowchart in Figure~\ref{fig:distance_flow} visualises the distance based algorithms described in this section and the relation between them. Algorithms following another are not necessarily better than the predecessor, but are either direct extensions or heavily draw inspiration from it.

\subsubsection{Elastic Ensemble (EE)}

The first algorithm to significantly outperform 1-NN DTW on the UCR data was the Elastic Ensemble (EE)~\citep{lines15elastic}. EE is a weighted ensemble of 11 1-NN classifiers with a range of elastic distance measures. It was the best performing distance based classifier in the bake off. Elastic distances can be slow, and EE requires cross validation to find the weights of each classifier in the ensemble. A caching mechanism was proposed to help speed up fitting the classifier~\citep{tan20fastee} and alternative speed ups were described in~\cite{oastler19significantly}.  The latter speed up is the version of EE we use in our experiments.

\subsubsection{Proximity Forest (PF)}

Proximity Forest (PF)~\citep{lucas19proximity} is an ensemble of Proximity Tree based classifiers. PF uses the same $11$ distance functions used by EE, but is more accurate and more scalable than the original EE algorithm. At every node of a tree, one of the 11 distances is selected to be applied with a fixed hyperparameter value. An exemplar single series is selected randomly for each class label. At every node, $r$ combinations of distance function, parameter value and class exemplars are randomly selected, and the combination with the highest Gini index split measure is selected. Series are passed down the branch with the exemplar that has the lowest distance to it, and the tree grows recursively until a node is pure.

\subsubsection{ShapeDTW}

Shape based DTW (ShapeDTW)~\citep{zhao18shape} works by extracting a set of shape descriptors over sliding windows of each series. The descriptors include slope, wavelet transforms and piecewise approximations. Based on the results presented~\citep{zhao18shape} we use the raw and derivative subsequences. The output data of these series-to-series transformations is then used with a 1-NN classifier with DTW.

\subsubsection{Generic RepresentAtIon Learning (GRAIL)}

The Generic RepresentAtIon Learning (GRAIL)~\citep{paparrizos19grail} paper focuses on efficient learning of time series representations that uphold bespoke distance function constraints. GRAIL harnesses kernel methods, particularly the Nystr\"om method, to learn precise representations within these constraints. The construction of representations involves expressing each time series as a linear combination of expressive landmarks, identified through cluster centroids. This approach gives rise to the Shift-Invariant Kernel (SINK) kernel function, which employs the Fast Fourier Transform to compare time series under shift invariance. GRAIL can be used to multiple time series related tasks, but for classification GRAIL and the SINK kernel are evaluated using a linear SVM.

\subsubsection{Comparison of Distance Based Approaches}

\begin{figure}[htb]
    \centering
    \includegraphics[width=1\linewidth]{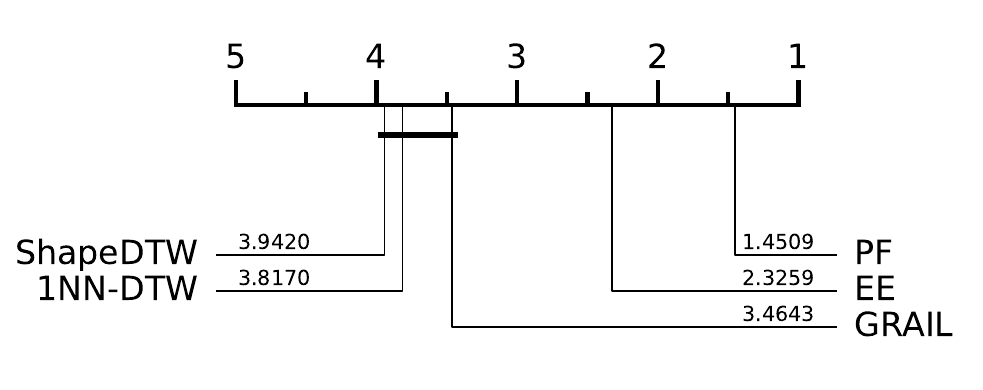}
    \caption{Ranked test accuracy of four distance based classifiers on 112 UCR UTSC problems. Accuracies are averaged over $30$ resamples of train and test splits.}
    \label{fig:distance}
\end{figure}

Figure~\ref{fig:distance} shows the relative rank test accuracies of the five distance based classifiers we discuss here, and Table~\ref{tab:DistanceBased} summarises four performance measures over these datasets. The results broadly validate previous findings. EE is significantly better than 1-NN DTW and PF is significantly better than EE. GRAIL performs slightly worse than expected. We have used the authors implementation\footnote{\url{https://github.com/TheDatumOrg/grail-python}} but have made some modifications to prevent the test set from being visible during the initial clustering, as it is incompatible with our experimental procedure. This would introduce bias, and may explain the discrepancy with published results alongside the different datasets and data resampling used.

Table~\ref{tab:DistanceBased} shows PF is over 2.5\% better in test accuracy and balanced test accuracy, has higher AUROC and lower NLL. Hence, we take PF as best of the distance based category.

\begin{table}[htb]
    \centering
    \caption{Summary performance measures for distance based classifiers on $30$ resamples of 112 UTSC problems. Best in bold.}
    \begin{tabular}{l|ccccc}
            & ACC   & BALACC & AUROC & NLL & F1 \\ \hline
    PF      & \textbf{0.837 (1)} & \textbf{0.819 (1)} & \textbf{0.942 (1)}  & \textbf{0.692 (1)} & \textbf{0.833 (1)}\\
    EE      & 0.811 (2) & 0.793 (2) & 0.918 (2) & 1.97 (3) & 0.806 (2)\\
    GRAIL   & 0.727 (5) & 0.699 (5) & 0.864 (3) & 0.788 (2) & 0.706 (5)\\
    1NN-DTW & 0.756 (3) & 0.739 (3) & 0.820 (4) & 8.812 (4) & 0.752 (3)\\
    ShapeDTW& 0.742 (4) & 0.726 (4) & 0.812 (5) & 9.282 (5) & 0.739 (4)\\
    \hline
    \end{tabular}
    \label{tab:DistanceBased}
\end{table}

\subsection{Feature Based}\label{sec:features}
\label{sec:feature}

\begin{figure}[hb]
    \centering
    \includegraphics[width=0.9\linewidth]{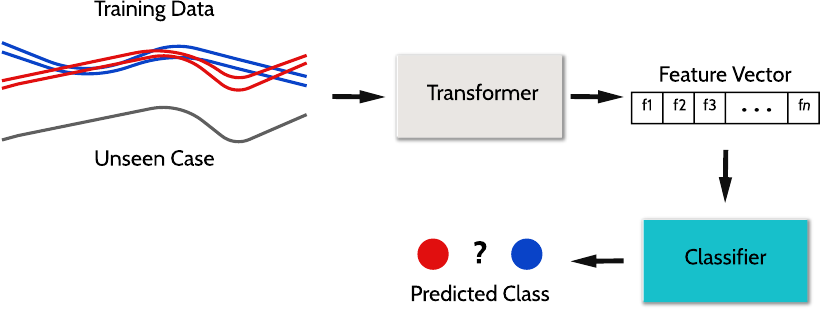}
    \caption{Visualisation of a pipeline classifier involving feature extraction followed by classification.}
    \label{fig:features}
\end{figure}

\emph{Feature based} classifiers are a popular recent theme. These extract descriptive statistics as features from time series to be used in classifiers. Typically, these features summarise the whole series, so we characterise these as series-to-vector transforms. Most commonly, these features are used in a simple pipeline of transformation followed by a classifier (see Figure~\ref{fig:features}). Several toolkits exist for extracting features.

 \begin{figure}[htb]
    \centering
    \includegraphics[width=.8\linewidth,trim={2cm 5.2cm 2cm 2cm},clip]{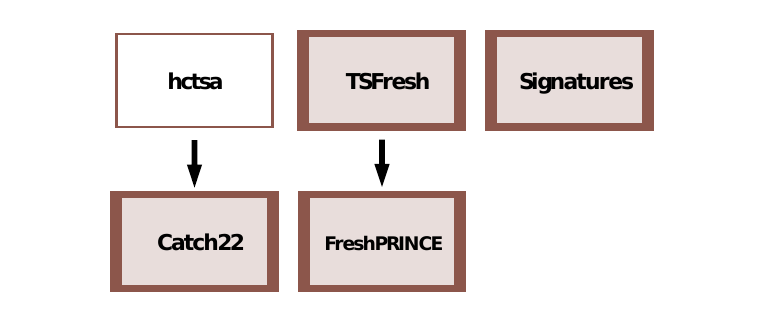}
    \caption{An overview of feature based classifiers and the relationship between them. Filled algorithms were released after the 2017 bake off~\citep{bagnall17bakeoff} and algorithms with a thin border are not included in our experiments.}
    \label{fig:feature_flow}
\end{figure}
The flowchart in Figure~\ref{fig:feature_flow} displays the feature based algorithms described in this section and related algorithms.

\subsubsection{The Canonical Time Series Characteristics (Catch22)}

The highly comparative time-series analysis (\textit{hctsa})~\citep{fulcher17hctsa} toolbox can create over $7700$ features for exploratory time series analysis. The canonical time series characteristics (Catch22)~\citep{lubba19catch22} are $22$ \textit{hctsa} features determined to be the most discriminatory of the full set. The Catch22 features were chosen by an evaluation on the UCR datasets. The \textit{hctsa} features were initially pruned, removing those which are sensitive to the series mean and variance and those that could not be calculated on over $80\%$ of the UCR datasets. A feature evaluation was then performed based on predictive performance. Any features which performed below a threshold were removed. For the remaining features, a hierarchical clustering was performed on the correlation matrix to remove redundancy. From each of the 22 clusters formed, a single feature was selected, taking into account balanced accuracy, computational efficiency and interpretability. The Catch22 features cover a wide range of concepts such as basic statistics of time series values, linear correlations, and entropy. Reported results for Catch22 are based on training a decision tree classifier after applying the transform to each time series~\citep{lubba19catch22}, the implementation we use builds a Random Forest classifier.

\subsubsection{Time Series Feature Extraction based on Scalable Hypothesis Tests (TSFresh)}

TSFresh~\citep{christ18time} is a collection of just under 800 features extracted from time series. While the features can be used on their own, a feature selection method called FRESH is provided to remove irrelevant features. FRESH considered each feature using multiple hypotheses tests, including Fisher's exact test~\citep{fisher1922interpretation}, the Kolmogorov-Smirnov~\citep{massey1951kolmogorov} test and the Kendal rank test~\citep{kendall1938new}. The Benjamini-Yekutieli procedure~\citep{benjamini2001control} is then used to control the false discovery rate caused by comparing multiple hypotheses and features simultaneously.

Results for the base features and after using the FRESH algorithm are reported using both a Random Forest and AdaBoost~\citep{freund96experiments} classifier. A comparison of alternative pipelines of feature extractor and classifier found that the most effective approach was the full set of TSFresh features with no feature selection applied, and combined with a Rotation Forest classifier~\citep{rodriguez06rotf}. This pipeline was called the FreshPRINCE~\citep{middlehurst22freshprince}. We include both TSFresh with feature selection using a Random Forest and the FreshPRINCE classifier in our comparison.

\subsubsection{Generalised Signatures}

Generalised signatures are a set of feature extraction techniques based on rough path theory. The generalised signature method~\citep{morrill20generalised} and the accompanying canonical signature pipeline can be used as a transformation for classification. Signatures are collections of ordered cross-moments. The pipeline begins by applying two augmentations. The basepoint augmentation simply adds a zero at the beginning of the time series, making the signature sensitive to translations of the time series. The time augmentation adds the series timestamps as an extra coordinate to guarantee that each signature is unique and obtain information about the parameterisation of the time series. A hierarchical window is run over the two augmented series, with the signature transform being applied to each window. The output for each window is then concatenated into a feature vector. The features are used to build a Random Forest classifier. The transformation was primarily developed for MTSC, but can be applied to univariate series.

\subsubsection{Comparison of Feature Based Approaches}

\begin{figure}[htb]
    \centering
    \includegraphics[width=1\linewidth]{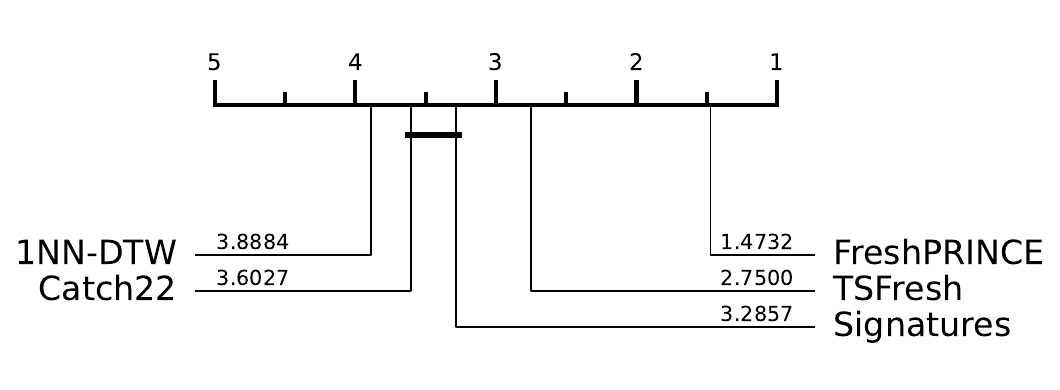}
    \caption{Ranked test accuracy of four feature based classifiers and the benchmark 1NN-DTW on 112 UCR UTSC problems. Accuracies are averaged over $30$ resamples of train and test splits.}
    \label{fig:cd_feature_based}
\end{figure}

Figure~\ref{fig:cd_feature_based} shows the relative rank performance, and Table~\ref{tab:FeatureBased} summarises the overall performance statistics. All four pipelines are significantly more accurate than 1-NN DTW. Excluding feature extraction and using Rotation Forest rather than Random Forest with TSFresh increases accuracy by over 0.05. This reinforces the findings that Rotation Forest is the most effective classifier for problems with continuous features~\citep{bagnall18rotf}.

\begin{table}[htb]
    \centering
    \caption{Summary performance measures for feature based classifiers on $30$ resamples of 112 UTSC problems. Best in bold.}
    \begin{tabular}{l|ccccc}
     & ACC & BALACC             & AUROC & NLL   & F1\\ \hline
    FreshPRINCE & \textbf{0.855 (1)} & \textbf{0.834 (1)} & \textbf{0.958 (1)} & \textbf{0.501 (1)} & \textbf{0.850 (1)} \\
    TSFresh    & 0.799 (2) & 0.772 (2)  & 0.902 (4) & 2.350 (4) & 0.778 (4)\\
    Signatures & 0.787 (4) & 0.763 (4)  & 0.92 (3)  & 0.730 (3) & 0.780 (3)\\
    Catch22    & 0.795 (3) & 0.771 (3)  & 0.929 (2) & 0.658 (2) & 0.788 (2)\\
    \hline
    \end{tabular}
    \label{tab:FeatureBased}
\end{table}

\subsection{Interval Based}\label{sec:intervals}

\begin{figure}[hb]
    \centering
    \includegraphics[width=.9\linewidth,trim={1cm 1cm 1.5cm 1.5cm},clip]{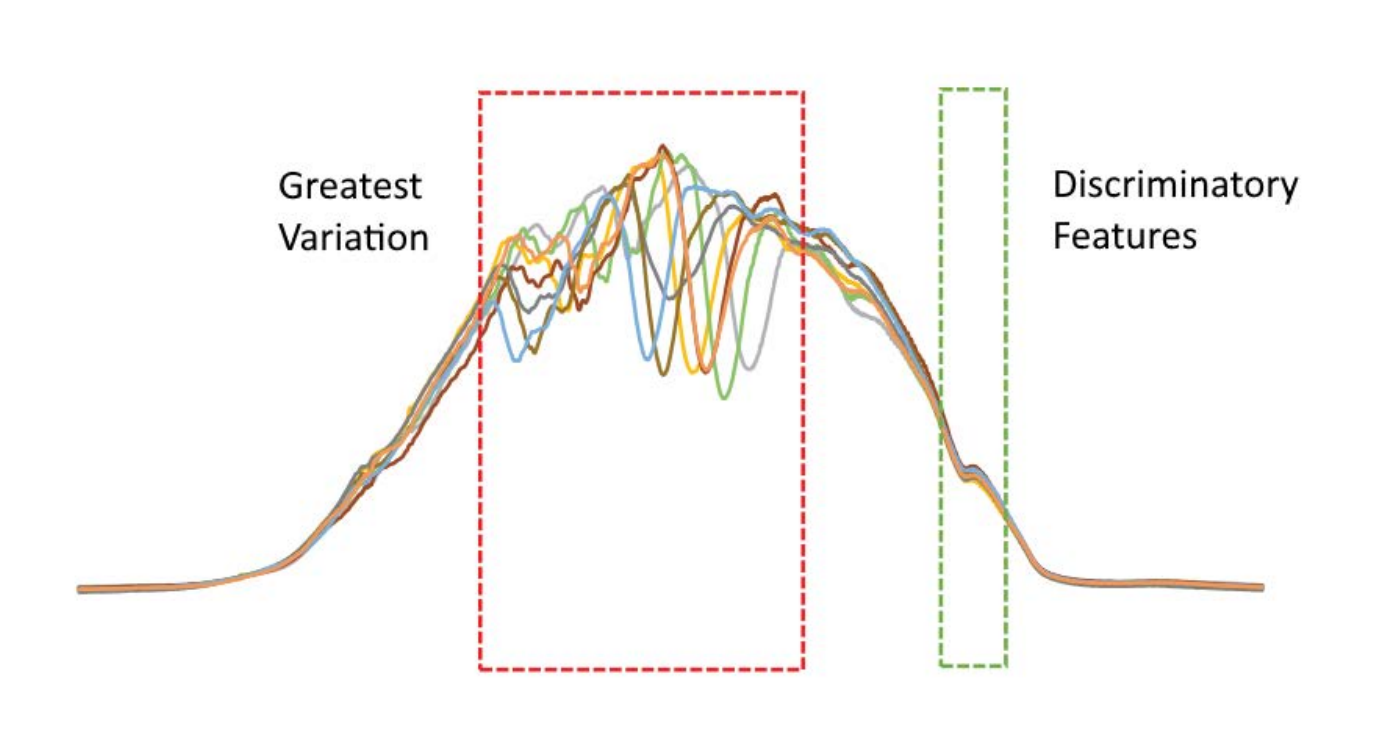}
    \caption{An example of a problem where interval based approaches may be superior. Each series is a spectrogram from a bottle of alcohol with a different concentration of ethanol. The discriminatory features are in the near infrared interval (green box to the right). However, the confounding factors such as bottle shape, labelling and colouring cause variation in the visible range (red box to the left). Using intervals containing just the near infrared features is likely to make classification easier.
    Image taken from~\citep{bagnall17bakeoff} with permission.}
    \label{fig:ethanol}
\end{figure}

\textit{Interval based} classifiers~\citep{deng13forest} extract phase dependent intervals of fixed offsets and compute (summary) statistics on these intervals. A majority of approaches include some form of random selection for choosing intervals, where  the same random interval locations are used across every series. Many of the interval based classifiers combine features from multiple random intervals. The motivation for taking intervals is to mitigate for confounding noise. Figure~\ref{fig:ethanol} shows an example problem where taking intervals will be better than using features derived from the whole series.

\begin{figure}[htb]
    \centering
    \includegraphics[width=0.65\linewidth]{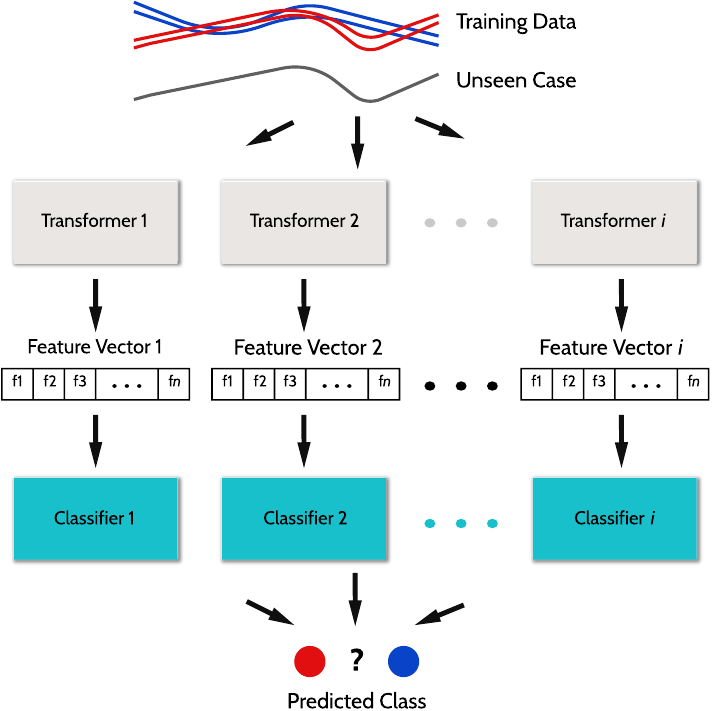}
    \caption{Visualisation of an ensemble of pipeline classifiers, as used in interval classifiers.}
    \label{fig:ensemble}
\end{figure}

\begin{figure}[tb]
    \centering
    \includegraphics[width=.8\linewidth,trim={2cm 3cm 2cm 2cm},clip]{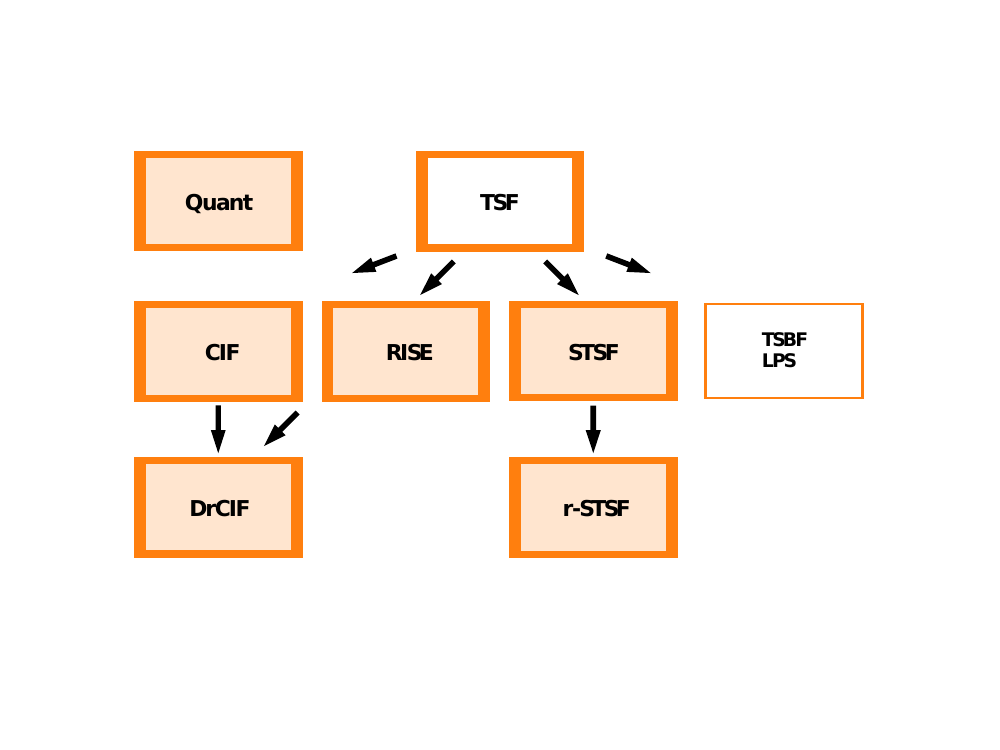}
    \caption{An overview of interval based classifiers and the relationship between them. Filled algorithms were released after the 2017 bake off~\citep{bagnall17bakeoff} and algorithms with a thin border are not included in our experiments.}
    \label{fig:interval_flow}
\end{figure}

Most recent interval based classifiers adopt a random forest ensemble model, where each base classifier is a pipeline of transformation and a tree classifier (visualised in Figure~\ref{fig:ensemble}). Diversity is injected through randomising the intervals for each tree. The relation flowchart for interval based algorithms is shown in Figure~\ref{fig:interval_flow}.

\subsubsection{Time Series Forest (TSF)}

The Time Series Forest (TSF)~\citep{deng13forest} is the simplest interval based tree based ensemble. For each tree, $\sqrt{m}$ (following the notation from Chapter~\ref{sec:definitions}, where $m$ is the length of the series and $d$ is the number of dimensions) intervals are selected with a random position and length. The same interval offsets are applied to all series. For each interval, three summary statistics (the mean, variance and slope) are extracted  and concatenated into a feature vector. This feature vector is used to build the tree, and features extracted from the same intervals are used to make predictions. The ensemble makes the prediction using a majority vote of base classifiers. The TSF base classifier is a modified decision tree classifier referred to as a time series tree, which considers all attributes at each node and uses a metric called margin gain to break ties.

\subsubsection{Random Interval Spectral Ensemble (RISE)}

First developed for the HIVE-COTE ensemble (described in Section~\ref{sec:hybrid}), the Random Interval Spectral Ensemble (RISE)~\citep{flynn19contract} is an interval based tree ensemble that uses spectral features. Unlike TSF, RISE selects a single random interval for each base classifier. The periodogram and auto-regression function are calculated over each randomly selected interval, and these features are concatenated into a feature vector, from which a tree is built. RISE was primarily designed for use with audio problems, where spectral features are more likely to be discriminatory.

\subsubsection{STSF and R-STSF}

\textbf{Supervised Time Series Forest (STSF)}~\citep{cabello20fast} is an interval based tree ensemble that includes a supervised method for extracting intervals. Intervals are found and extracted for a periodogram and the first order differences representation as well as the base series. STSF introduces bagging for each tree and extracts seven simple summary statistics from each interval. For each tree, an initial split point for the series is randomly selected. For both of these splits, the remaining subseries is cut in half, and the half with the higher Fisher score is retained as an interval. This process is then run recursively using higher scored intervals until the series is smaller than a threshold. This is repeated for each of the seven summary statistic features, with the extracted statistic being used to calculate the Fisher score.

\textbf{Randomised STSF (RSTSF)}~\citep{cabello21fast} is an extension of STSF, altering its components with more randomised elements. The split points for interval selection are selected randomly instead of splitting each candidate in half after the first. Intervals extracted from an autoregressive representation are included alongside the previous additions. Features are extracted multiple times from each representation into a single pool. Rather than extract different features for each tree in an ensemble, the features are used in a pipeline to build an Extra Trees~\cite{geurts06extremely} classifier.

\subsubsection{CIF and DrCIF}

\textbf{The Canonical Interval Forest (CIF)}~\citep{middlehurst20canonical} is another extension of TSF, that improves accuracy by integrating more informative features and by increasing diversity. Like other interval approaches, CIF is an ensemble of decision tree classifiers built on features extracted from phase dependent intervals. Alongside the mean, standard deviation and slope, CIF also extracts the Catch22 features described in Section~\ref{sec:feature}.  Intervals remain randomly generated, with each tree selecting $k=\sqrt{m}\sqrt{d}$ intervals. To add additional diversity to the ensemble, $a$ attributes out of the pool of $25$ are randomly selected for each tree. The extracted features are concatenated into a $k \cdot a$ length vector for each time series and used to build the tree. For multivariate data, CIF randomly selects the dimension used for each interval.

\textbf{The Diverse Representation Canonical Interval Forest (DrCIF)}~\citep{middlehurst21hc2} incorporates two new series representations: the periodograms (also used by RISE and STSF) and first order differences (also used by STSF). For each of the three representations, $(4 + \sqrt{r}\sqrt{d})/3$ phase dependent intervals are randomly selected and concatenated into a feature vector, where $r$ is the length of the series for a representation.

\subsubsection{QUANT}

QUANT~\citep{dempster23quant} employs a singular feature type, quantiles, to encapsulate the distribution of a given time series. The method combines four distinct representations, namely raw time series, first-order differences, Fourier coefficients, and second-order differences. The extraction process involves fixed, dyadic intervals derived from the time series. These disjoint intervals are constructed through a pyramid structure, where each level successively halves the interval length. At depths greater than one, an identical set of intervals, shifted by half the interval length, is also included. The total count of intervals is calculated as $2^{(d-1)} \times 4 - 2 - d$ for a depth of $d=\min(6, \log_2 n + 1)$. Each representation can have up to $120$ intervals, resulting in a total of $480$ intervals across all four representations. The concatenated feature vector is used to build an Extra Trees classifier.

\subsubsection{Comparison of Interval Based Approaches}

\begin{figure}[htb]
    \centering
    \includegraphics[width=1\linewidth]{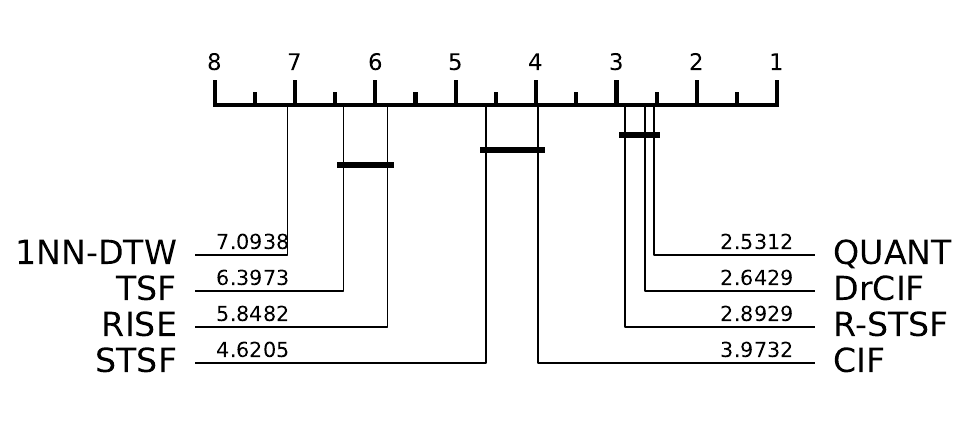}
    \caption{Ranked test accuracy of seven interval based classifiers on 112 UCR UTSC problems. Accuracies are averaged over $30$ resamples of train and test splits.}
    \label{fig:interval}
\end{figure}

Figure~\ref{fig:interval} shows the relative ranks of seven interval classifiers, with summary performance measures presented in Table~\ref{tab:IntervalBased}.

\begin{table}[htb]
\centering
    \caption{Summary performance measures for interval based classifiers on $30$ resamples of 112 UTSC problems. Best in bold.}
    \begin{tabular}{l|ccccc}
          & ACC    & BALACC & AUROC  & NLL  & F1    \\ \hline
    QUANT & \textbf{0.867 (1)}  & \textbf{0.845 (1)}  & \textbf{0.962 (1)}  &  0.496 (3) & \textbf{0.862 (1)} \\
    DrCIF & 0.864 (2) & 0.841 (3) & \textbf{0.962 (1)}  & \textbf{0.489 (1)} & 0.858 (2)\\
    R-STSF & 0.864 (2) & 0.842 (2) & 0.961 (3) & 0.493 (2) & 0.858 (2)\\
    CIF   & 0.848 (4) & 0.824 (5) & 0.954 (5) & 0.539 (4) & 0.842 (4)\\
    STSF  & 0.846 (5) & 0.827 (4) & 0.955 (4) & 0.555 (5) & 0.841 (5)\\
    RISE  & 0.806 (6) & 0.777 (7) & 0.937 (6) & 0.758 (7) & 0.796 (6)\\
    TSF   & 0.802 (7) & 0.780 (6) & 0.930 (7) & 0.615 (6) & 0.796 (6)\\
    \hline
    \end{tabular}
    \label{tab:IntervalBased}
\end{table}

\begin{figure}[tb]
    \centering
    \begin{tabular}{c c}
        \includegraphics[width=0.5\linewidth]{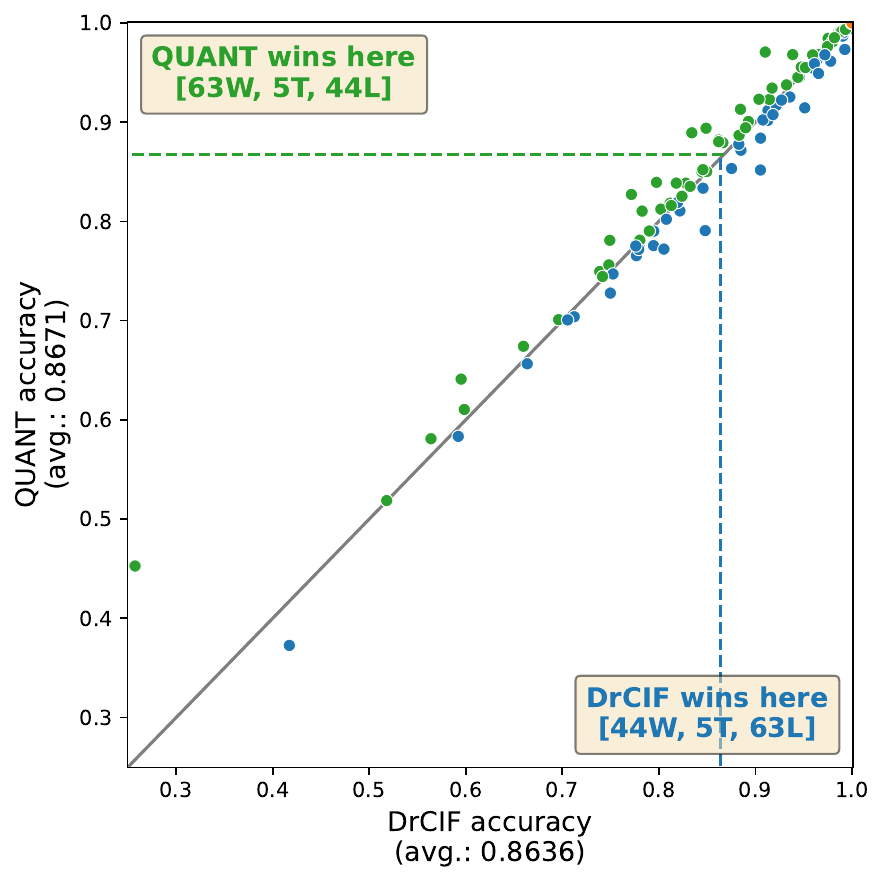} &
        \includegraphics[width=0.5\linewidth]{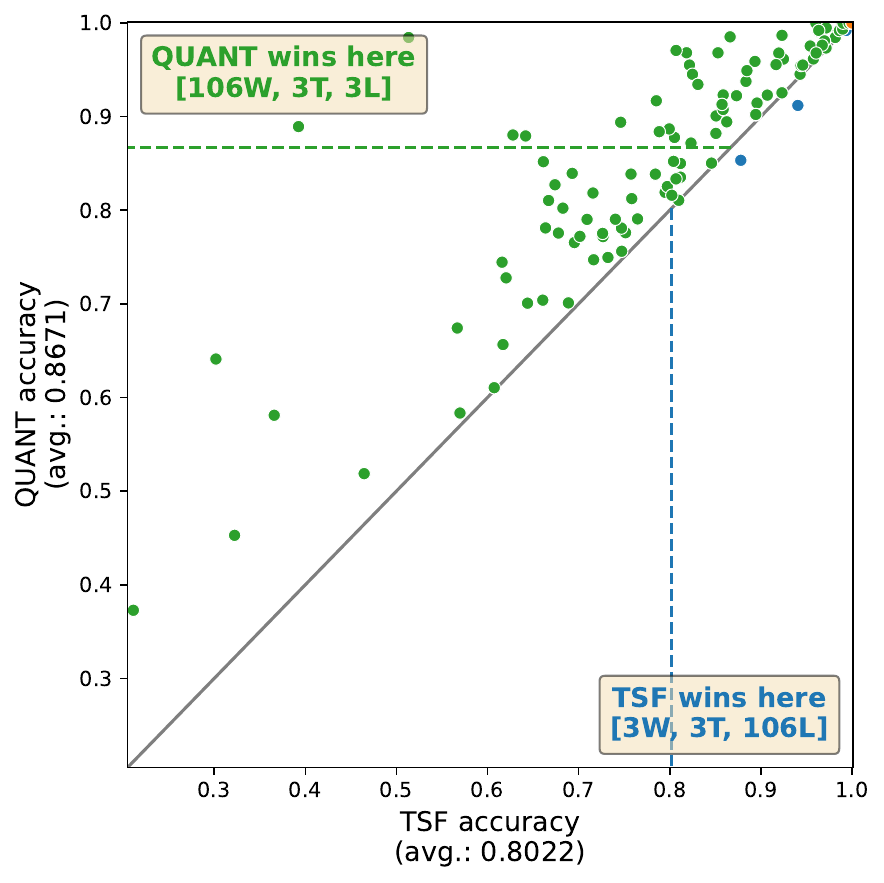} \\
        (a) QUANT vs DrCIF & (b) QUANT vs TSF
    \end{tabular}
    \caption{Scatter plot of test accuracies of DrCIF against RSTSF and TSF. TSF is better than DrCIF on just 6 of the 112 datasets. }
    \label{fig:intervalscatter}
\end{figure}

There is no significant difference between QUANT, DrCIF and RSTSF nor between their precursors CIF and STSF. All are significantly better than TSF, the best in class in the bake off. Figure~\ref{fig:intervalscatter}(a) shows the scatter plot of QUANT vs DrCIF. QUANT wins on $63$, draws $5$ and loses $44$. Overall, the two algorithms produce very similar results (the test accuracies have a correlation of $98.1\%$).

We choose QUANT as the best in class because it is significantly faster than DrCIF and RSTSF. Figure~\ref{fig:intervalscatter}(b) shows QUANT against TSF in order to confirm that QUANT, DrCIF and RSTSF represent genuine improvements to this type of algorithm over the previous best. Table~\ref{tab:IntervalBased} confirms that on average over 112 problems, the accuracy of the top clique is over $0.06$ higher than TSF.

\subsection{Shapelet Based}\label{sec:shapelets}

\begin{figure}
    \centering
    \includegraphics[width=1.0\linewidth]{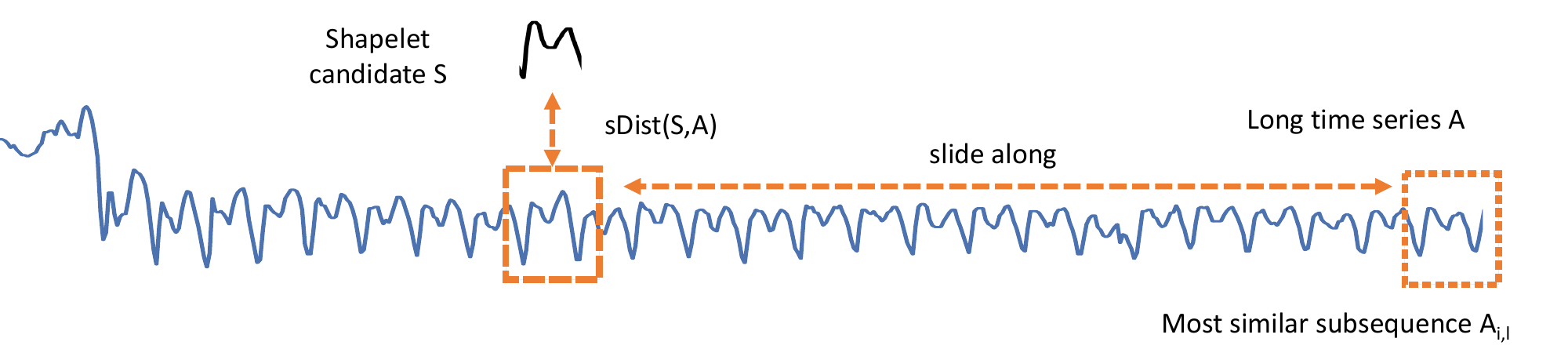}
    \caption{Visualisation of the shapelet distance operation $sDist()$ between a shapelet $S$ and a series $A$, which finds the closest distances to the shapelet from all possible subseries of the same length.}
    \label{fig:sdist}
\end{figure}

\emph{Shapelets} are subseries from the training data that are independent of the phase and can be used to discriminate between classes of time series based on their presence or absence. To evaluate a shapelet, the subseries is slid across the time series, and the z-normalised Euclidean distance between the shapelet and the underlying window is calculated. The distance between a shapelet and any series, $sDist()$, is the minimum distance over all such windows.  Figure~\ref{fig:sdist} shows a visualisation of the $sDist()$ process. The shapelet $S$ is shifted along the time series $A$, and the most similar offset and distance in $A$ are recorded. The distance between a shapelet and the training series is then used as a feature to evaluate the quality of the shapelet.

\begin{figure}[tb]
    \centering
    \includegraphics[width=.8\linewidth,trim={5cm 1cm 5cm 1cm},clip]{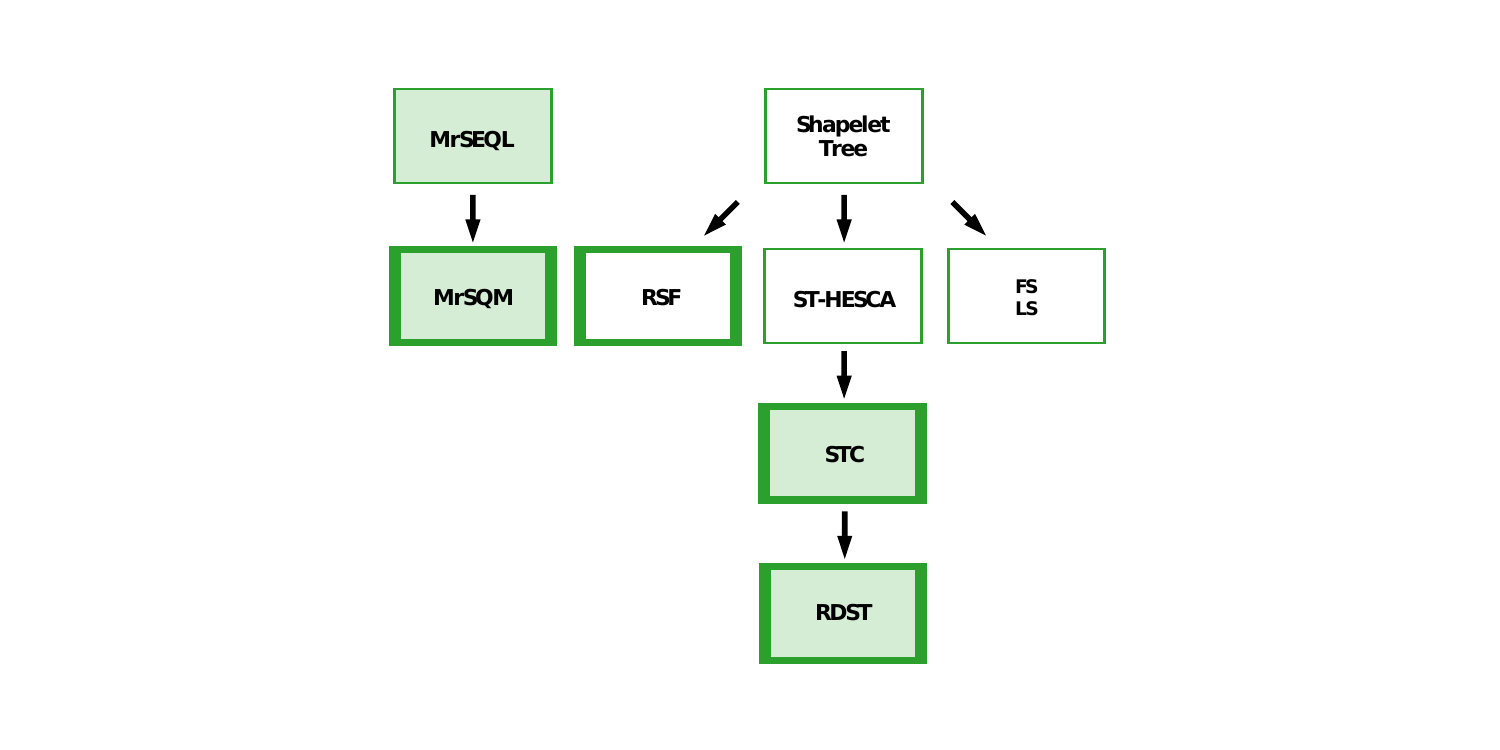}
    \caption{An overview of shapelet based classifiers and the relationship between them. Filled algorithms were released after the 2017 bake off~\citep{bagnall17bakeoff} and algorithms with a thin border are not included in our experiments.}
    \label{fig:shapelet_flow}
\end{figure}

Shapelets were first proposed as a primitive in~\cite{ye11shapelets}, and were embedded in a decision tree classifier. There have been four important themes in shapelet research post bake off: The first has concentrated on finding the best way to use shapelets to maximise classification accuracy. The second has focused on overcoming the shortcomings of the original shapelet discovery which required full enumeration of the search space and has cubic complexity in the time series length; the third theme is the progress toward unifying research with convolutions and shapelets; and the fourth theme is the balance between optimisation, randomisation and interpretability when finding shapelets. The relation flowchart for shapelet based algorithms is shown in Figure~\ref{fig:shapelet_flow}.

\subsubsection{The Shapelet Transform Classifier (STC)}

The Shapelet Transform Classifier (STC)~\citep{hills14shapelet} is a pipeline classifier which searches the training data for shapelets, transforms series to vectors of $sDist()$ distances to a filtered set of selected shapelets based on information gain, then builds a classifier on the latter. This is in contrast to the decision tree based approaches, which search for the best shapelet at each tree node. The first version of STC performed a full enumeration of all shapelets from all train cases before selecting the top $k$. The base classifier used was HESCA (later renamed CAWPE, ~\cite{large19cawpe}) ensemble of classifiers, a weighted heterogeneous ensemble of 8 classifiers including a diverse set of linear, tree based and Bayesian classifiers. Due to its full enumeration and large pool of base classifiers requiring weights, the algorithm does not scale well. We call the original full enumeration version \textbf{ST-HESCA} to differentiate it from the version described below which we simply call \textbf{STC}. It was the best performing shapelet based classifier in the bake off.

The following incremental changes have been made to the STC pipeline, described in~\cite{bostrom16evaluating,bostrom17binary}:
\begin{enumerate}
    \item Search has been randomised, and the number of shapelets sampled is now a parameter, which defaults to 10,000. This does not lead to significantly worse performance on the UCR datasets.
    \item Shapelets are now binary, in that they represent the class of the origin series and are evaluated against all other classes as a single class using one hot encoding. This facilitates greater use of the early abandon of the order line creation (described in~\citep{ye11shapelets}), and makes evaluation of split points faster.
    \item The heterogeneous ensemble of base classifiers in HESCA has been replaced with a single Rotation Forest~\citep{rodriguez06rotf} classifier, making STC a simple pipeline classifier.
\end{enumerate}

\subsubsection{The Generalised Random Shapelet Forest (RSF)}

The Random Shapelet Forest (RSF)~\citep{karlsson16generalized} is a bagging based tree ensemble that attempts to improve the computational efficiency and predictive accuracy of the Shapelet Tree through randomisation and ensembling. At each node of each tree $r$ univariate shapelets are selected from the training set at random. Each shapelet has a randomly selected length between predefined upper and lower limits. The quality of a shapelets is measured in the standard way with $sDist()$ and information gain, and the best is selected. The data is split, and a tree is recursively built until a stopping condition is met. New samples are predicted by a majority vote on the tree's predictions and multiple trees are ensembled.

\subsubsection{MrSEQL and MrSQM}

The Multiple Representation Sequence Learner (MrSEQL)~\citep{nguyen19interpretable}, is an ensemble classifier that extends previous adaptations of the SEQL classifier~\citep{nguyen17sequence}. MrSEQL looks for the presence or absence of a pattern (shapelet) in the data. Rather than using a distance based approach to measure the presence or not of a shapelet, MrSEQL discretises subseries into words. Words are generated through two symbolic representations, using SAX~\citep{lin07sax} for time domain and SFA~\citep{schaefer12sfa} for frequency domain. A set of discriminative words is selected through Sequence Learner (SEQL) and the output of training is a logistic regression model, which in concept is a vector of relevant subseries and their weights. Diversification is achieved through the two different symbolic representations and varying the window size.

MrSQM~\citep{nguyen22mrsqm} extends MrSEQL. It also combines two symbolic transformations to create words from subseries and trains a logistic regression classifier. What sets it apart is its innovative strategy for selecting features (substrings).

To begin with, MrSQM uses SFA and SAX to discretise time series subseries into words. It then utilizes a trie to store and rank frequent substrings, and applies either (a) a supervised chi-squared test to identify discriminative words or (b) an unsupervised random substring sampling method to prevent overestimating highly correlated substrings that are likely to be redundant. MrSQM establishes the number of learned representations (SFA or SAX) based on the length of the time series and utilizes an exponential scale for the window size parameter.

\subsubsection{Random Dilated Shapelet Transform (RDST)}

The Random Dilated Shapelet Transform (RDST)~\citep{guillaume22rdst} is a shapelet-based algorithm that adopts many of the techniques of convolution approaches described in Section~\ref{sec:conv}. While traditional shapelet algorithms search for the best shapelets from the train dataset, RDST takes a different approach by randomly selecting a large number of shapelets from the train data, typically ranging from thousands to tens of thousands, then training a linear Ridge classifier on features derived from these shapelets.

RDST employs dilation with shapelets. Dilation is a form of down sampling, in that it defines spaces between time points. Hence, a shapelet with dilation $d$ is compared to time points $d$ steps apart when calculating the distance. RDST also uses two features in addition to $sDist()$: it encodes the position of the minimum distance, and records a measure of the frequency of occurrences of the shapelet based on a threshold. Hence the transformed data has $3k$ features for $k$ shapelets.

\subsubsection{Comparison of Shapelet Based Approaches}

Table~\ref{tab:shapelet_family} highlights the key differences between the shapelet-based approaches.

\begin{table}
    \small
    \begin{centering}
    \caption{Key differences in shapelet based TSC algorithms.}
\label{tab:shapelet_family}
    \begin{tabular}{|p{1.8cm}|>{\centering}p{2.0cm}|>{\centering}p{2.0cm}|>{\centering}p{2.0cm}|>{\centering}p{2.0cm}|}
    \hline
     & \textbf{STC} & \textbf{RDST} & \textbf{RSF} & \textbf{MrSQM}\tabularnewline
    \hline
    Shapelet Discovery & Random Subsquences & Random Subsquences & Random subseries & Frequent Substrings\tabularnewline
    \hline
    Supervised shapelets & yes & no & yes & no \tabularnewline \hline
    Dilation & no & yes & no & no \tabularnewline
    \hline
    Discretisation & no & no & no & yes (SAX/SFA) \tabularnewline
    \hline
    Classification & Rotation Forest & Ridge Classifier CV & Random Tree Ensemble & Ridge Classifier CV \tabularnewline
    \hline
    \end{tabular}
    \par\end{centering}
\end{table}

 \begin{figure}[htb]
    \centering
    \includegraphics[width=1\linewidth,trim={0cm 0.7cm 0cm 0.7cm},clip]{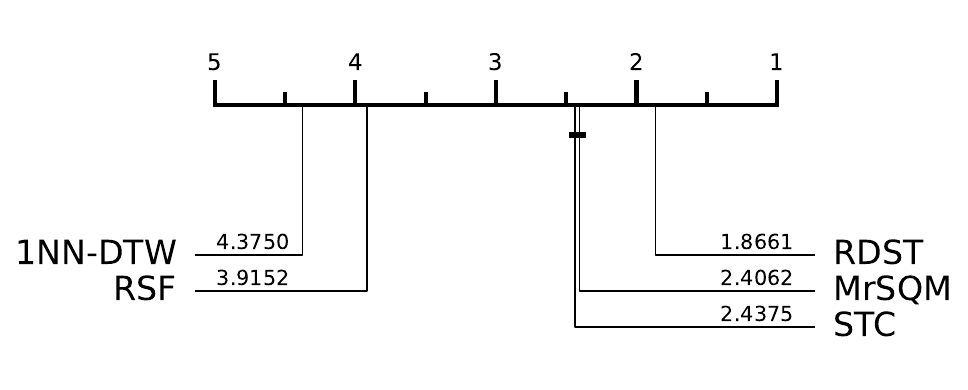}
    \caption{Ranked test accuracy of four shapelet based classifiers and the benchmark 1NN-DTW on 112 UCR UTSC problems. Accuracies are averaged over $30$ resamples of train and test splits.}
    \label{fig:shapelets}
\end{figure}

\begin{figure}[htb]
    \centering
    \begin{tabular}{c c}
        \includegraphics[width=0.5\linewidth]{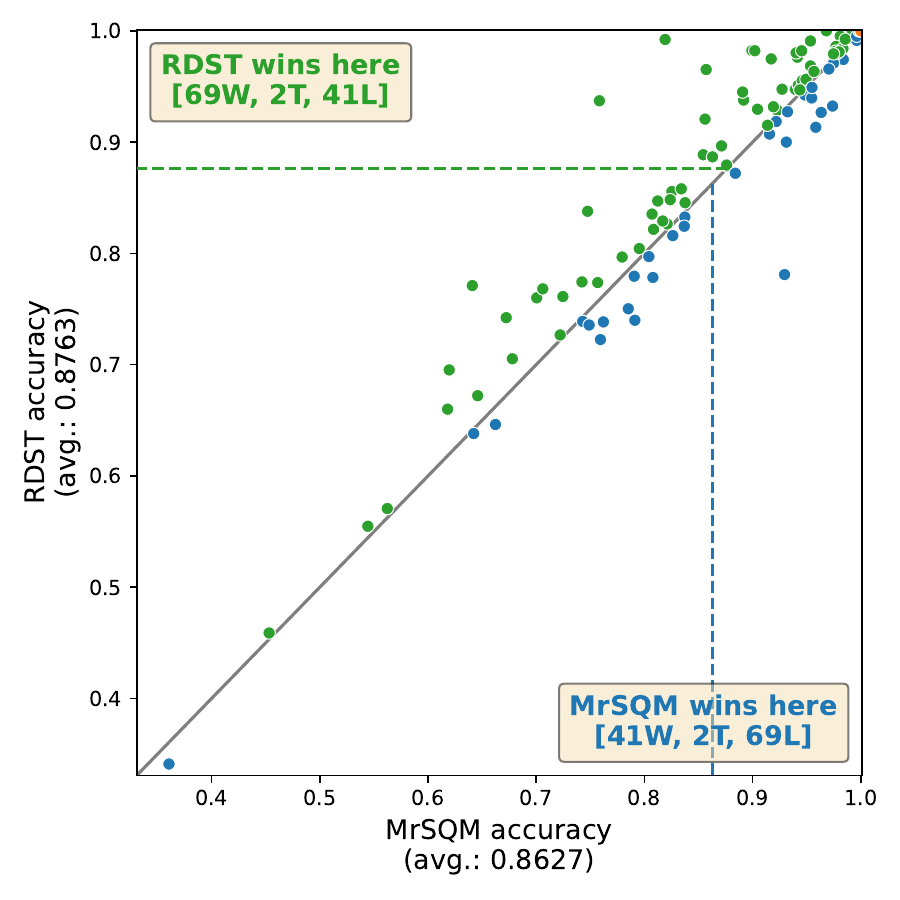} &
        \includegraphics[width=0.5\linewidth]{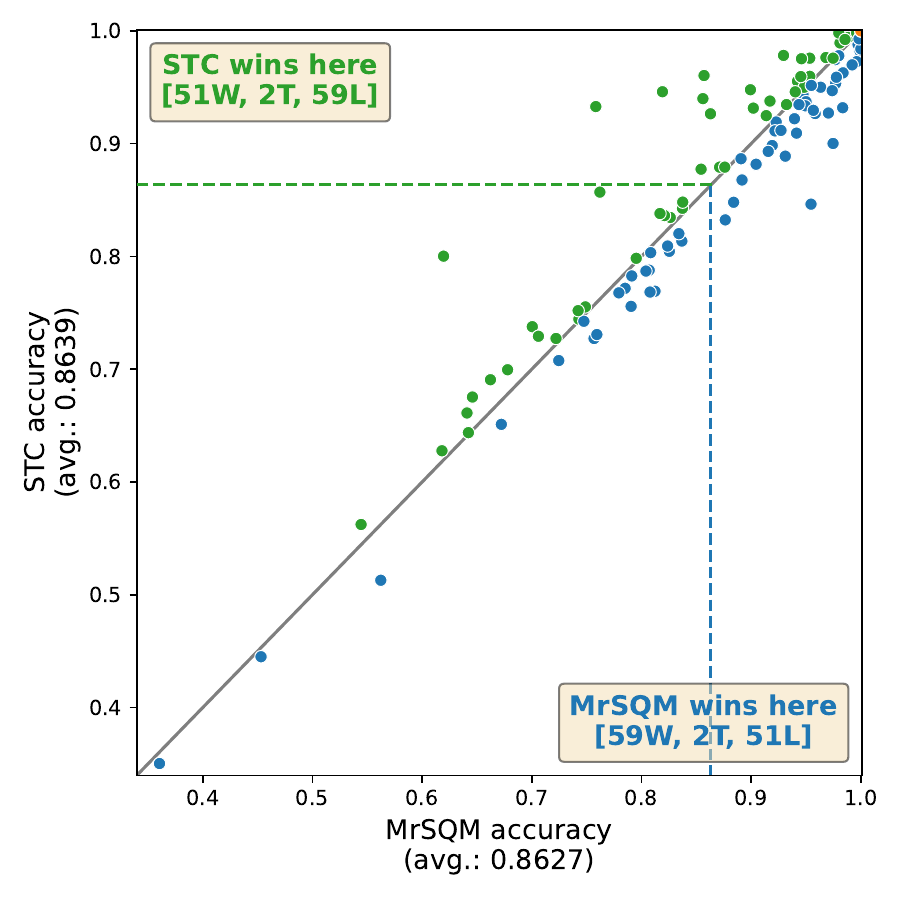}\\
        (a) RDST vs MrSQM & (b) STC vs MrSQM
    \end{tabular}
    \caption{Scatter plot of test accuracies of shapelet based classifiers.}
    \label{fig:shapelet_scatter}
\end{figure}

Figure~\ref{fig:shapelets} shows the relative ranks of the four shapelet classifiers. RDST is the clear winner. Table~\ref{tab:ShapeletBased} shows it is, on average more than 1\% more accurate than MrSQM, the second-best algorithm. The shapelet based algorithms are more fundamentally different in design than, for example, interval classifiers. This is demonstrated by the spread of test accuracies shown in Figure~\ref{fig:shapelet_scatter} of the top three algorithms. The grouping may become redundant: RDST is more similar to convolution based algorithms (Section~\ref{sec:conv}) in design than STC, and MrSQM has structure in common with dictionary based classifiers (Section~\ref{sec:dictionary}). However, they still retain the key characteristics that, unlike convolutions, they use the training data to find subseries and, unlike dictionary based algorithms, their features include the presence or absence of a pattern.

\begin{table}[htb]
    \centering
    \caption{Summary performance measures for shapelet based classifiers on $30$ resamples of 112 UTSC problems. Best in bold.}
    \begin{tabular}{l|ccccc}
          & ACC    & BALACC & AUROC  & NLL   & F1 \\ \hline
    RDST  & \textbf{0.876 (1)}  & \textbf{0.856 (1)}  & 0.907 (4) & 4.457 (4) & \textbf{0.872 (1)}\\
    MrSQM & 0.863 (3)  & 0.841 (3)  & \textbf{0.959 (1)}  & \textbf{0.446 (1)} & 0.857 (3)\\
    STC   & 0.864 (2)  & 0.842 (2)  & 0.958 (2)  & 0.485 (2) & 0.859 (2)\\
    RSF   & 0.801 (4)  & 0.774 (4)  & 0.927 (3)  & 0.724 (3) & 0.792 (4)\\
    \hline
    \end{tabular}
    \label{tab:ShapeletBased}
\end{table}

\subsection{Dictionary Based}
\label{sec:dictionary}

\begin{figure}[hb]
    \begin{centering}
    \includegraphics[width=1\linewidth]{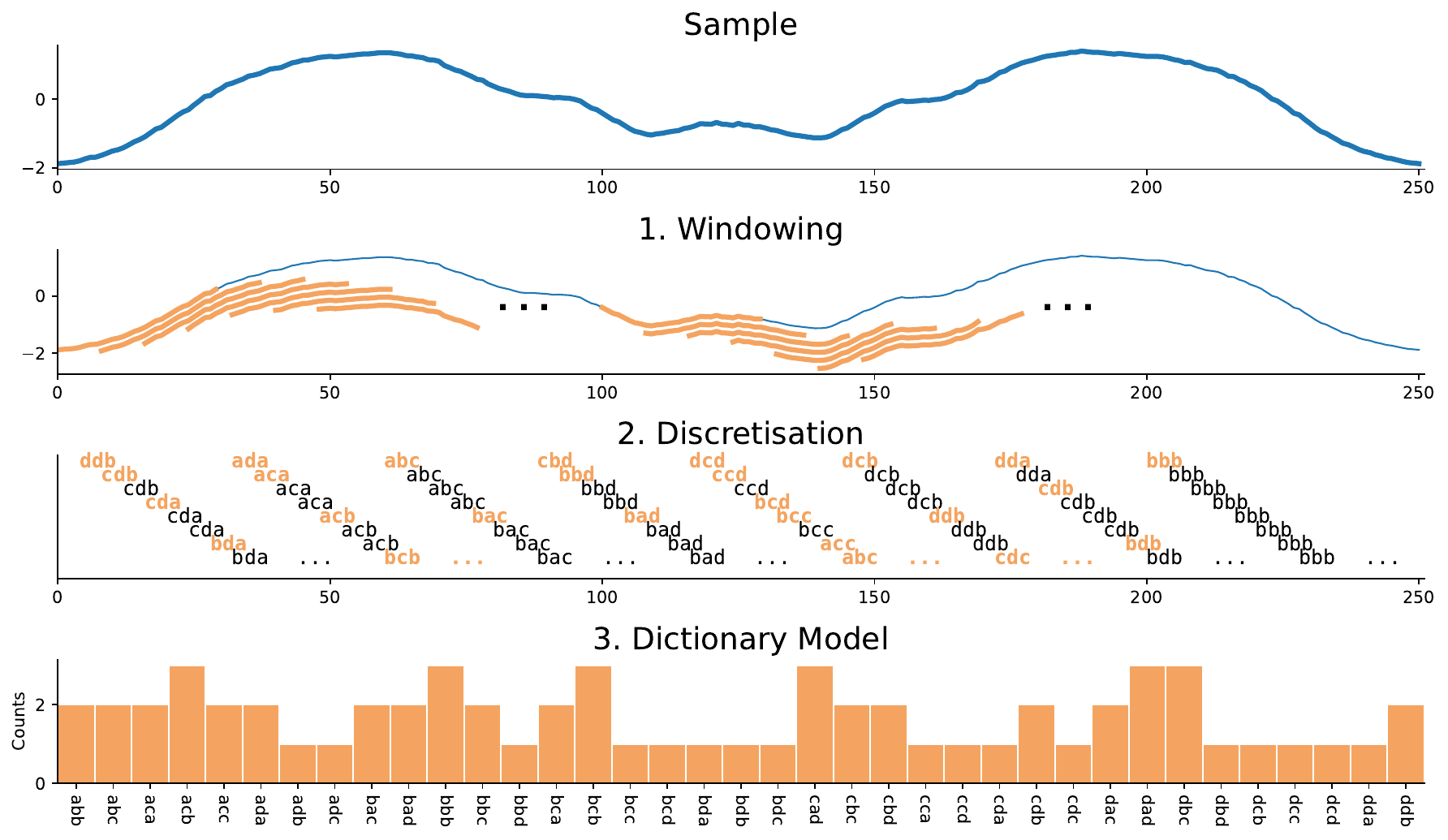}
    \par\end{centering}
    \caption{Transformation of a TS into the dictionary-based model (following ~\citep{schaefer17weasel}) using overlapping windows (second to top), discretisation of windows to words (second from bottom), and word counts (bottom).\label{fig:transformation}}
\end{figure}

Similar to shapelet based algorithms, \emph{dictionary approaches} extract phase-independent subseries. However, instead of measuring the distance to a subseries, each window is converted into a short sequence of discrete symbols, commonly known as a word. Dictionary methods differentiate based on word frequency and are often referred to as bag-of-words approaches.  Figure~\ref{fig:transformation} illustrates the process that algorithms following the dictionary model take to create a classifier. This process can be summarized as:

\begin{enumerate}
    \item Extracting subseries, or windows, from a time series;
    \item Transforming each window of real values into a discrete-valued \emph{word} (a sequence of symbols over a fixed alphabet);
    \item Building a sparse feature vector of histograms of word counts, and
    \item Finally, using a classification method from the machine learning repertoire on these feature vectors.
\end{enumerate}

Dictionary-based methods differ in the way they transform a window of real-valued measurements into discrete words. For example, the basis of the BOSS model~\citep{schaefer15boss} is a representation called Symbolic Fourier Approximation (SFA)~\citep{schaefer12sfa}. SFA works as follows:

\begin{enumerate}
    \item Values in each window of length $w$ are normalized to have standard deviation of $1$ to obtain amplitude invariance.
    \item Each normalized window of length $w$ is subjected to dimensionality reduction by the use of the truncated Fourier transform, keeping only the first $l<w$ coefficients for further analysis. This step acts as a low pass filter, as higher order Fourier coefficients typically represent rapid changes like dropouts or noise.
    \item Discretisation bins are derived through Multiple Coefficient Binning (MCB). It separately records the $l$ distributions of the real and imaginary values of the Fourier transform. These distributions are then subjected to either equi-depth or equi-width binning. The resulting output consists of $l$ sets of bins, corresponding to the target word length of $l$.
    \item Each coefficient is discretized to a symbol of an alphabet of fixed size $\alpha$ to achieve further robustness against noise.
\end{enumerate}

\begin{figure}
    \begin{centering}
    \includegraphics[width=1\linewidth]{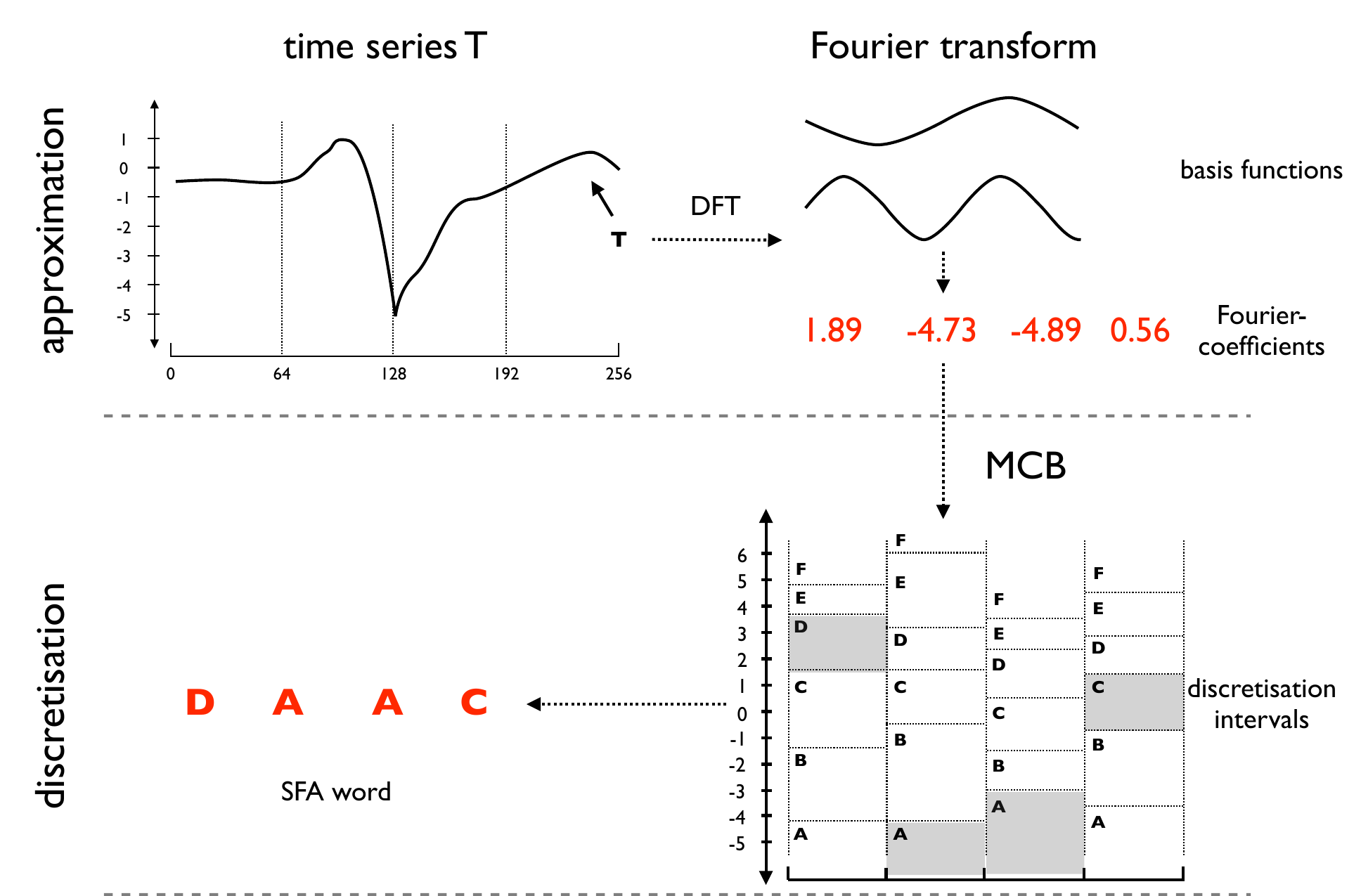}
    \par\end{centering}
    \caption{The Symbolic Fourier Approximation (SFA) (from ~\citep{schaefer17weasel}): A time series~(top left) is approximated using the  Fourier transform~(top right) and discretised to the word \emph{DAAC}~(bottom left) using data adaptive bins (bottom right).\label{fig:SFATransform}}
\end{figure}

\begin{figure}[tb]
    \centering
    \includegraphics[width=.8\linewidth,trim={2cm 3cm 2cm 2cm},clip]{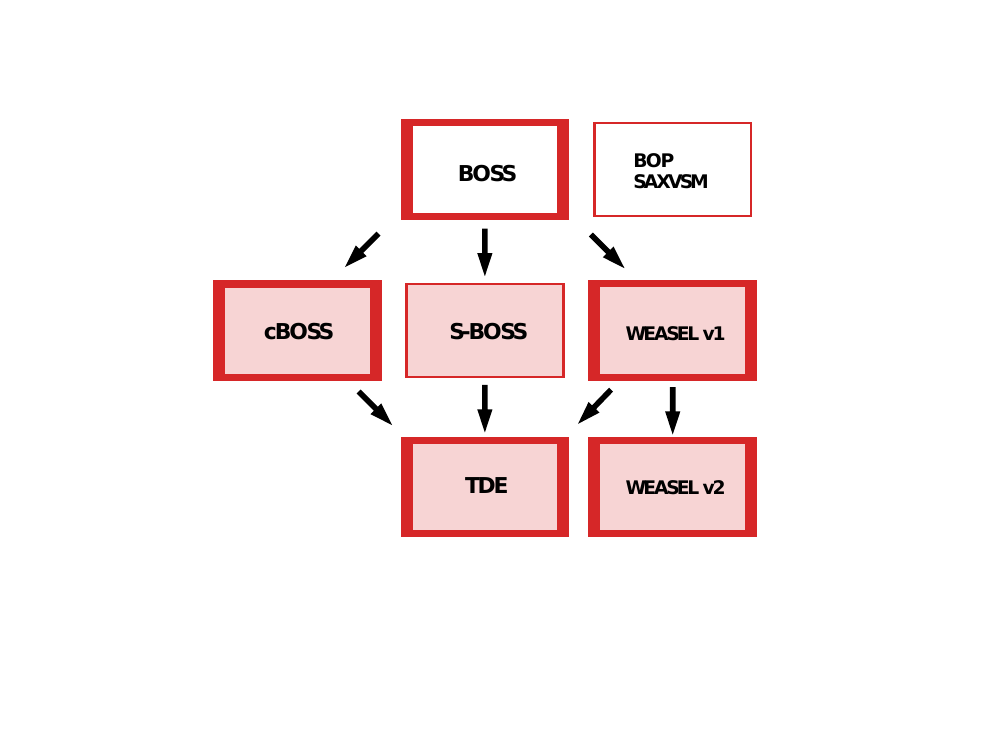}
    \caption{An overview of dictionary based classifiers and the relationship between them. Filled algorithms were released after the 2017 bake off~\citep{bagnall17bakeoff} and algorithms with a thin border are not included in our experiments.}
    \label{fig:dictionary_flow}
\end{figure}

Figure~\ref{fig:SFATransform} exemplifies this process from a window of length $128$ to its DFT representation, and finally the word \emph{DAAC}. The relation flowchart for dictionary based algorithms is shown in Figure~\ref{fig:dictionary_flow}.

\subsubsection{Bag-of-SFA-Symbols (BOSS)}

Bag-of-SFA-Symbols (BOSS)~\citep{schaefer15boss} was among the top-performing algorithms in the initial bake-off study and led to significant further investigation into dictionary-based classifiers.  An individual BOSS classifier undergoes the same process described earlier, whereby each sliding window is transformed into a word using SFA. Subsequently, a feature vector is generated by counting the occurrences of each word over all windows. A non-symmetric distance function is then employed with a 1-NN classifier to categorize new instances. Experiments have shown that when presented with a query and a sample time series, disregarding words that exist solely in the sample time series using, the non-symmetric distance function leads to improved performance compared to using the Euclidean distance metric~\citep{schaefer15boss}.

The complete BOSS classifier is an ensemble of individual BOSS classifiers. This ensemble is created by exploring a range of parameters, assessing each base classifier through cross-validation, and keeping all base classifiers with an estimated accuracy within $92\%$ of the best classifier. For new instances, the final prediction is obtained through a majority vote of the base classifiers.

\subsubsection{Word Extraction for Time Series Classification (WEASEL v1.0)}

Word Extraction for Time Series Classification (WEASEL v1.0)~\citep{schaefer17weasel} is a pipeline classifier that revolves around identifying words whose frequency count distinguishes between classes and discarding words that lack discriminatory power. The classifier generates histograms of word counts over a broad spectrum of window sizes and word lengths parameters, including bigram words produced from non-overlapping windows. A Chi-squared test is then applied to determine the discriminatory power of each word, and those that fall below a particular threshold are discarded through feature selection. Finally, a linear Ridge classifier is trained on the remaining feature space. WEASEL utilizes a supervised variation of SFA to create discriminative words, and it leverages an information-gain based methodology for identifying breakpoints that separate the classes.

\subsubsection{WEASEL v2.0 (with dilation)}

The dictionary-based WEASEL v2.0~\citep{schaefer23weasle2} is a complete overhaul of the WEASEL v1.0 classifier~\citep{schaefer17weasel}. It addresses the problem of the extensive memory footprint of WEASEL by controlling the search space using randomly parameterized SFA transformations. It also significantly improves accuracy. Notably, the most prominent modification is the inclusion of dilation to the sliding window approach. Table~\ref{tab:dict_family} presents a comprehensive summary of its alterations.

To extract subseries with non-consecutive values from a time series, a dilated sliding window approach is employed, where the dilation parameter maintains a fixed gap between each value. These dilated subseries undergo a Fourier transform, and a word is generated by discretising them using SFA. The unsupervised learning of bins is achieved using equi-depth and equi-width with an alphabet size of $2$. To improve performance, a feature selection strategy based on variance is introduced, which retains only the real and imaginary Fourier values with the highest variance.

Each of the $50$ to $150$ SFA transformations is randomly initialized subject to:
\begin{enumerate}
    \item \textbf{Window length} $w$: Randomly chosen from interval $[w\_{min}, \dots, w\_{max}]$.
    \item \textbf{Dilation} $d$: Randomly chosen from interval $[1,\dots,  2^{\log(\frac{n-1}{w-1})}]$. The formula is inherited from the convolution-based ROCKET group of classifiers~\cite{dempster20rocket,dempster21minirocket,tan22multirocket}.
    \item \textbf{Word length} $l$: Randomly chosen from $\{7,8\}$.
    \item \textbf{Binning strategy}: Randomly chosen from \{"equi-depth", "equi-width"\}.
    \item \textbf{First order differences}: To extract words from both, the raw time series, and its first order difference, effectively doubling the feature space.
\end{enumerate}

When using an alphabet size of $2$ and a length of $8$, each SFA transformation creates a dictionary containing only $256$ unique words of a fixed size. These dictionaries are then combined to produce a feature vector containing approximately $30k$ to $70k$ features. No feature selection is implemented by default. The resulting features serve as input for training a linear Ridge classifier.

\subsubsection{Contractable BOSS (cBOSS)}

The size of the parameter grid searched by BOSS is data dependent, and BOSS uses a method of retaining ensemble members using a threshold of accuracy estimated from the train data. This makes its time and memory complexity unpredictable. BOSS was one of the slower algorithms tested in the bake off and could not be evaluated on the larger datasets in reasonable time.
Contractable BOSS (cBOSS)~\citep{middlehurst19scalable} revises the ensemble structure of BOSS to solve these scalability issues, using the same base transformations as the BOSS ensemble. cBOSS randomly selects $k$ parameter sets of hyper-parameters ($w$, $l$ and $\alpha$) for BOSS base classifiers. It retains the best $s$ classifiers (based on a cross validation estimate of accuracy) are retained for the final ensemble. cBOSS allows the $k$ parameter to be replaced by a train time limit $t$ through contraction, allowing the user to better control the training time of the classifier. A subsample of the train data is randomly selected without replacement for each ensemble member and an exponential weighting scheme used in the CAWPE~\citep{large19cawpe} ensemble is introduced. The cBOSS alterations to the BOSS ensemble structure showed an order of magnitude improvement in train times with no reduction in accuracy.

\subsubsection{SpatialBOSS}

\begin{figure}
    \begin{centering}
    \includegraphics[width=1\linewidth]{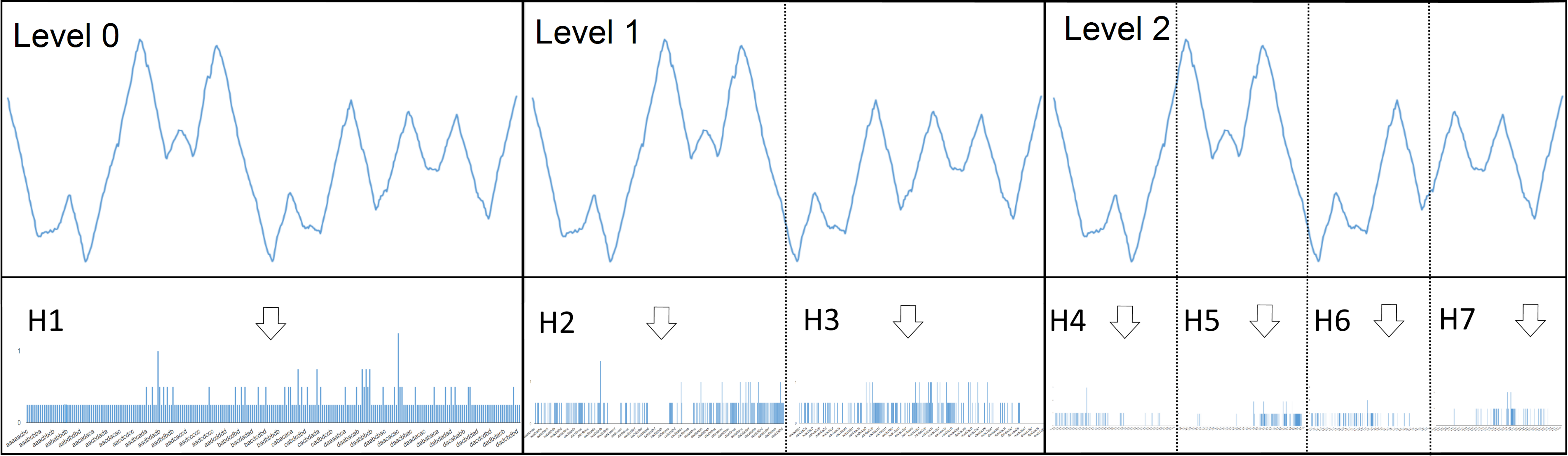}
    \par\end{centering}
    \caption{An example of using a spatial pyramid to form 7 distinct word count histograms.}
    \label{fig:sboss}
\end{figure}

BOSS intentionally ignores the locations of words in series, classifying based on the frequency of patterns rather than their location. Spatial Boss~\citep{large19dictionary} introduced location information into the design of a BOSS classifier. Spatial pyramids~\citep{lazebnik06pyramid} are a technique used in computer vision to retain some temporal information back into the bag-of-words paradigm. The core idea, illustrated in Figure~\ref{fig:sboss} is to split the series into different resolutions, segmenting the series based on depth and position, then building independent histograms on the splits. The histograms for each level are concatenated into a single feature vector which is used with a 1-NN classifier. While more accurate than BOSS, the increase in parameter search space and bag size makes it very difficult to run in practice.

\subsubsection{Temporal Dictionary Ensemble (TDE)}

The Temporal Dictionary Ensemble (TDE)~\citep{middlehurst20temporal} combines the best improvements introduced in WEASEL, SpatialBOSS and cBOSS and also includes several novel features. TDE is an ensemble of 1-NN classifiers which transforms each series into a histogram of word counts using SFA~\citep{schaefer12sfa}. From WEASEL, TDE takes the method for finding supervised breakpoints for discretisation, and captures frequencies of bigrams found from non-overlapping windows. The locality information derived from the spatial pyramids used in SpatialBOSS are incorporated. Word counts are found for each spatial subseries independently, with the resulting histograms being concatenated. Bigrams are only found for the full series.
The cBOSS ensemble structure is applied with a modified parameter space sampling algorithm. It first randomly samples a small number of parameter sets, then constructs a Gaussian processes regressor on the historic accuracy for unseen parameter sets. The regressor is used to estimate the parameter set for the next candidate, and the model is then updated before the process is repeated. TDE has two additional parameters for its candidate models: the number of levels for the spatial pyramid and the method of generating breakpoints.

\subsubsection{Comparison of Dictionary Based Approaches}

\begin{figure}[htb]
    \centering
    \includegraphics[width=1\linewidth]{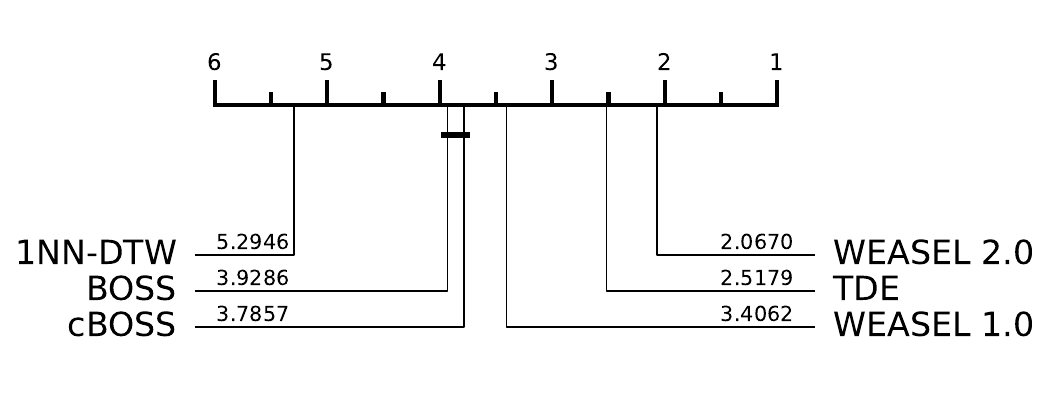}
    \caption{Ranked test accuracy of five dictionary based classifiers with the benchmark 1-NN DTW on 112 UCR UTSC problems. Accuracies are averaged over $30$ resamples of train and test splits.}
    \label{fig:dictionary}
\end{figure}

\begin{figure}[htb]
    \centering
            \begin{tabular}{c c}
    \includegraphics[width=0.5\linewidth]{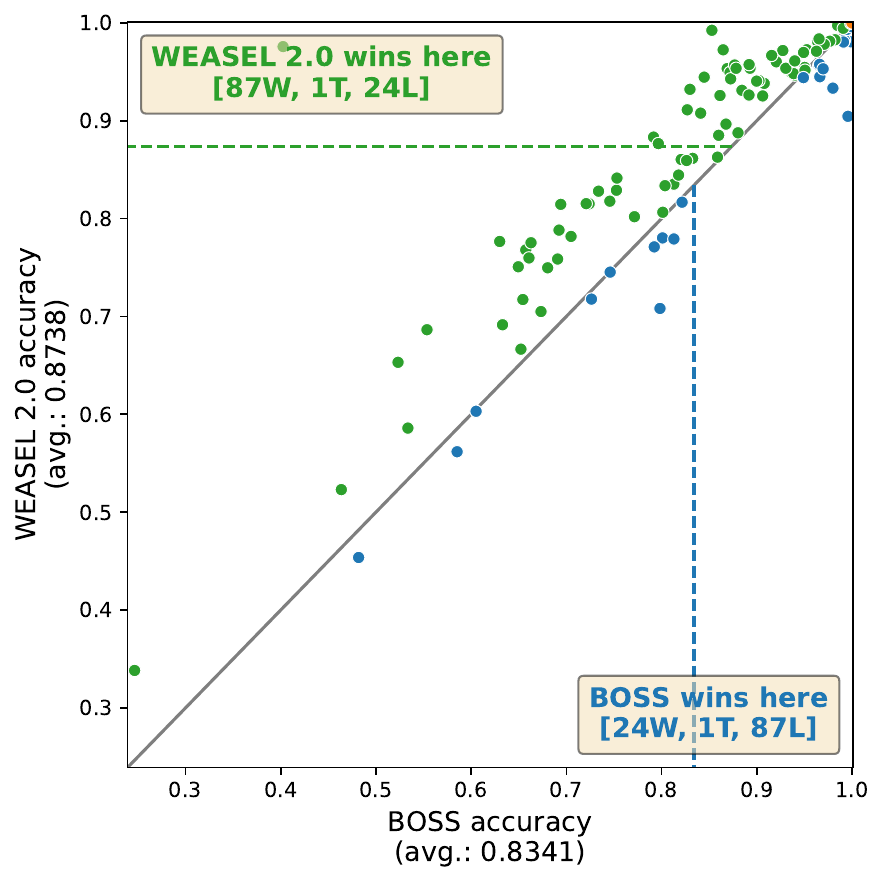} &
    \includegraphics[width=0.5\linewidth]{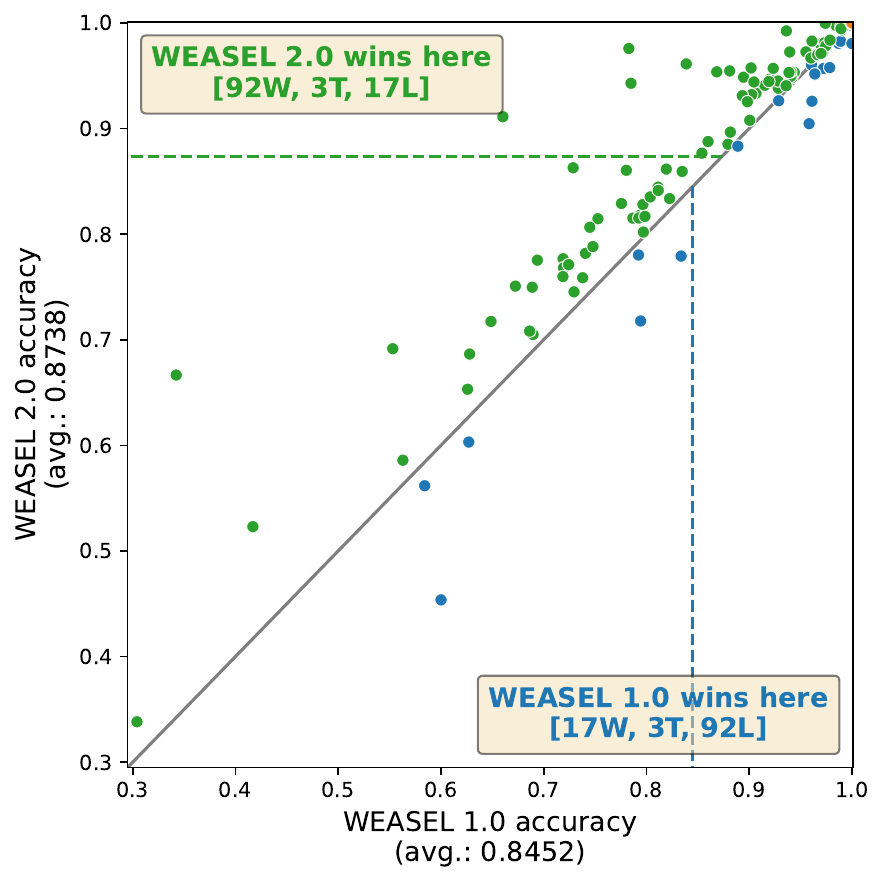}\\
    (a) BOSS vs WEASEL 2.0 & (b) WEASEL 1.0 vs WEASEL 2.0
    \end{tabular}
    \caption{Scatter plot of test accuracies of BOSS vs WEASEL 2.0.}
    \label{fig:dict_scatter}
\end{figure}

Table~\ref{tab:dict_family} shows the key design differences between the dictionary based approaches.
\begin{table}
    \small
    \caption{Key differences in dictionary based TSC algorithms}
    \label{tab:dict_family}
    \begin{centering}
    \begin{tabular}{|p{2.1cm}|>{\centering}p{2.3cm}|>{\centering}p{2.3cm}|>{\centering}p{2.2cm}|>{\centering}p{2.2cm}|}
    \hline
     & \textbf{BOSS} & \textbf{TDE} & \textbf{WEASEL v1.0} & \textbf{WEASEL v2.0}\tabularnewline
    \hline
    \hline
    word length & \{8,10,12,14,16\} & \{8,10,12,14,16\} & \{4, 6\} & \{7, 8\}\tabularnewline
    \hline
    alphabet-size & 4 & 4 & 4 & 2\tabularnewline
    \hline
    FFT Features & First & First & Anova-F-Test & Variance\tabularnewline
    \hline
    Binning & equi-depth & equi-depth, IG & IG & equi-depth, equi-width\tabularnewline
    \hline
    Bigrams & No & Yes & Yes & No\tabularnewline
    \hline
    Pyramids & No & Yes & No & No\tabularnewline
    \hline
    Feature Selection & None & None & Chi-squared & optional (default: None)\tabularnewline
    \hline
    Window Sizes & variable (Ensemble) & variable (Ensemble) & variable (concatenate) & variable (concatenate)\tabularnewline
    \hline
    1st order dif.& no & no & no & yes\tabularnewline
    \hline
    Dilation & no & no & no & yes\tabularnewline
    \hline
    Classifier & 1-NN & 1-NN & Ridge Regression & Ridge Regression CV\tabularnewline
    \hline
    Feature Vector Size & data dependent & data dependent & data dependent & 30 to 70k\tabularnewline
    \hline
    \end{tabular}
    \par\end{centering}
\end{table}
Figure~\ref{fig:dictionary} shows the ranked test accuracy of five dictionary classifiers we have described, with 1-NN DTW as a benchmark. SpatialBOSS is not included due to its significant runtime and memory requirements which would require the exclusion of multiple datasets. We believe that comparing more recent advances on the full archive is more valuable than its inclusion, and suggest those interested in SpatialBOSS view the results presented in~\cite{middlehurst20temporal} which show it is comparable to WEASEL 1.0 in performance. WEASEL 1.0 and TDE are significantly more accurate than BOSS, but WEASEL 2.0 is the most accurate overall. Figure~\ref{fig:dict_scatter}(a) illustrates the improvement of WEASEL 2.0 over BOSS, and Figure~\ref{fig:dict_scatter}(b) shows the improvement dilation provides over WEASEL. Table~\ref{tab:dictionary} summarises the performance of the four new dictionary algorithms. WEASEL 2.0 is on average $4\%$ more accurate than BOSS and improves balanced accuracy by almost the same amount.

\begin{table}[htb]
    \centering
    \caption{Averaged performance statistics for dictionary based classifiers on $30$ resamples of 112 UTSC problems. Best in bold.}
    \begin{tabular}{l|ccccc}
                & ACC & BALACC  & AUROC & NLL  & F1 \\ \hline
    WEASEL v2.0 & \textbf{0.874 (1)} & \textbf{0.853 (1)} & 0.945 (2) & 4.547 (5) & \textbf{0.869 (1)}\\
    TDE         & 0.861 (2) & 0.836 (2) & \textbf{0.954 (1)} & \textbf{0.646 (1)} & 0.854 (2)\\
    WEASEL v1.0 & 0.845 (3) & 0.824 (3) & 0.905 (5) & 0.891 (3) & 0.841 (3)\\
    cBOSS       & 0.833 (5) & 0.806 (5) & 0.945 (2) & 0.732 (2) & 0.824 (5)\\
    BOSS        & 0.834 (4) & 0.813 (4) & 0.938 (4) & 1.385 (4) & 0.828 (4)\\
    \hline
    \end{tabular}
        \label{tab:dictionary}
\end{table}

\subsection{Convolution Based}\label{sec:conv}

\textit{Kernel/Convolution} classifiers use convolutions with kernels, which can be seen as subseries used to derive discriminatory features. Each kernel is convolved with a time series through a sliding dot product creating an activation map. Technically, each convolution creates a series to series transform from time series to the activation map (see Definition~\ref{def:convolution}). Activation maps are used to create summary features. Convolutions and shapelets share a close methodological relationship. Shapelets can be realised through a convolution operation, followed by a min-pooling operation on the array of windowed Euclidean distances. This was first observed by~\citep{grabocka14learning-shapelets}. However, despite this methodological connection, there is significant difference in the results obtained by convolution based and shapelet based approach, as illustrated in the Appendix~\ref{app:results}, Figure~\ref{fig:conv_shapelet_comparison}. For example, the ROCKET results are negatively correlated with shapelet based approaches such as STC or RDST. The main difference between convolutions and shapelets is that shapelets are subseries from the training data whereas convolutions are found from the entire space of possible real-values.

\begin{figure}
    \centering
    \includegraphics[width=1\linewidth]{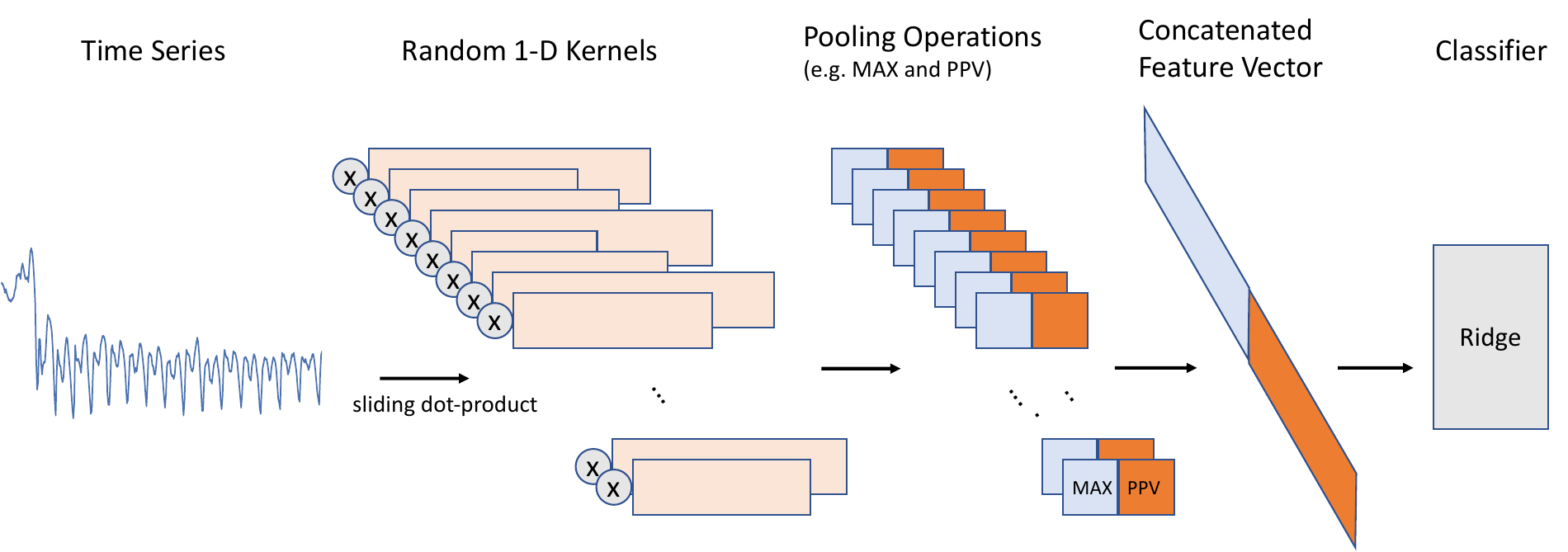}
    \caption{Pipeline of convolution based approaches such as ROCKET, MiniROCKET or  MultiROCKET.}
    \label{fig:rocket_pipeline}
\end{figure}
\begin{figure}[tb]
    \centering
    \includegraphics[width=.8\linewidth,trim={4cm 1cm 8cm 1cm},clip]{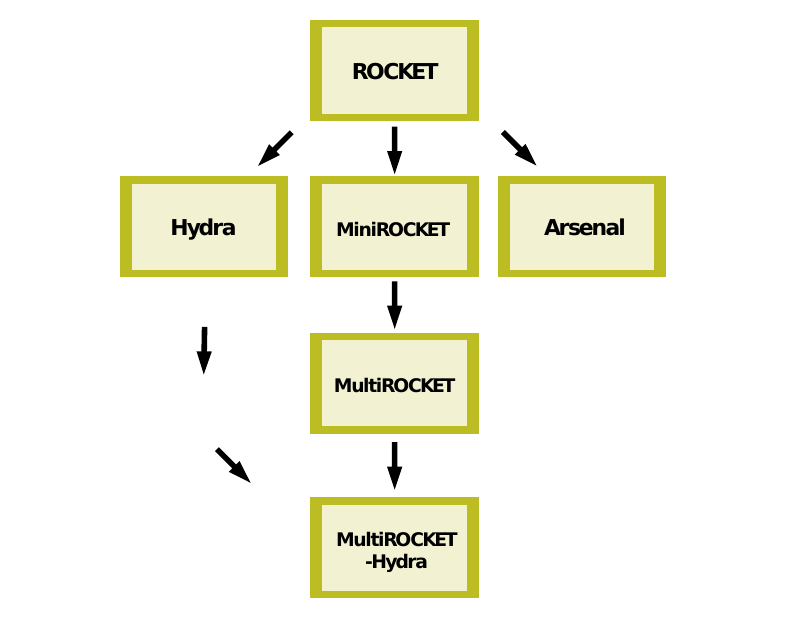}
    \caption{An overview of convolution based classifiers and the relationship between them.}
    \label{fig:convolution_flow}
\end{figure}

Convolution based TSC algorithms follow a standard pipeline pattern depicted in Figure~\ref{fig:rocket_pipeline}. The activation map is formed for each convolution, followed by pooling operations to extract one relevant feature for each operation. The resulting features are then concatenated to form a single feature vector. Finally, a Ridge classifier is trained on the output to classify the data. The relation flowchart for convolution based algorithms is shown in Figure~\ref{fig:convolution_flow}.

\subsubsection{Random Convolutional Kernel Transform (ROCKET)}

The most well known convolutional approach is the Random Convolutional Kernel Transform (ROCKET)~\citep{dempster20rocket}. ROCKET is a pipeline classifier. It generates a large number of randomly parameterised convolutional kernels  (typically in the range of thousands to tens of thousands), then uses these to transform the data through two pooling operations: the max value and the proportion of positive values (PPV). These two features are concatenated into a feature vector for all kernels. For $k$ kernels, the transformed data has $2k$ features.

In ROCKET, each kernel is randomly initialised with respect to the following parameters:
\begin{enumerate}
    \item the \emph{kernel length $l$}, randomly selected from $\{7,9,11\}$;
    \item the \emph{kernel weights $w$}, randomly initialised from a normal distribution;
    \item a \emph{bias term $b$} added to the result of the convolution operation;
    \item the \emph{dilation $d$} to define the spread of the kernel weights over the input instance, which allows for detecting patterns at different frequencies and scales. The dilation is randomly drawn from an exponential function; and
    \item padding \emph{$p$} the input series at the start and the end (typically with zeros), such that the activation map has the same length as the input;
\end{enumerate}

The result of applying a kernel $\omega$ with dilation $d$ to a time series $T$ at offset $i$ is defined by:
    $$ T_{i : (i+l)} * \omega = \sum_{j=0}^{l-1} T_{i-(\lfloor m / 2 \rfloor) \times d)+(j \times d)}  \times w_j $$

 The feature vectors are then used to train a Ridge classifier using cross-validation to train the $L_2$-regularisation parameter $\alpha$. A Logistic Regression classifier is suggested as a replacement for larger datasets. The combination of ROCKET with Logistic (RIDGE) Regression  is conceptually the same as a single-layer Convolutional Neural Network with randomly initialised kernels and softmax loss.

\subsubsection{Mini-ROCKET and Multi-ROCKET}

ROCKET has two extensions. The first extension is MiniROCKET~\citep{dempster21minirocket}, which speeds up ROCKET by over an order of magnitude with no significant difference in accuracy. MiniROCKET removes many of the random components of ROCKET, making the classifier almost deterministic. The kernel length is fixed to 9, only two weight values are used, and the bias value is drawn from the convolution output. Only the PPV is extracted, discarding the max. These changes alongside general optimisations taking advantage of the new fixed values provide a considerable speed-up to the algorithm. MiniROCKET generates a total of 10k features from 10k kernels and PPV pooling.

MultiROCKET~\citep{tan22multirocket} further extends the MiniROCKET improvements, extracting features from first order differences and adding three new pooling operations extracted from each kernel: mean of positive values (MPV), mean of indices of positive values (MIPV) and longest stretch of positive values (LSPV). MultiROCKET generates a total of 50k features from 10k kernels and 5 pooling operations.

\subsubsection{Hydra and MultiROCKET-Hydra}

\textbf{HYbrid Dictionary–ROCKET Architecture (Hydra)}~\citep{dempster22hydra} is a model that combines dictionary-based and convolution-based models. It begins by utilizing random convolutional kernels to calculate the activation of time series. These kernels, unlike ROCKET, are arranged into $g$ groups of $k$ kernels each. In each group of $k$ kernels, the activation of a kernel with the input time series is calculated, and we record how frequently this kernel is the best match (counts the highest activation). This results in a $k$-dimensional count vector for each of the $g$ groups, resulting in a total of $g \times k$ features.

To implement Hydra, the time series is convolved with the kernels, and the resulting activation maps are organized into $g$ groups. Next, an (arg)max operation is performed to count the number of best matches, and the counts for each group's dictionary are increased. The main hyperparameters to consider are the number of groups and the number of kernels per group, with default values of $g = 64$ and $k = 8$. Hydra is applied to both the time series and its first-order differences. The best results in ~\citep{dempster22hydra} come from concatenating features from Hydra with features from MultiROCKET to form its pipeline. We call this classifier \textbf{MultiROCKET-Hydra}.

\subsubsection{Comparison of Convolution Based Approaches}

Table~\ref{tab:rocket_family} highlights the key differences between the convolution based approaches.

\begin{figure}[htb]
    \centering
    \includegraphics[width=1\linewidth]{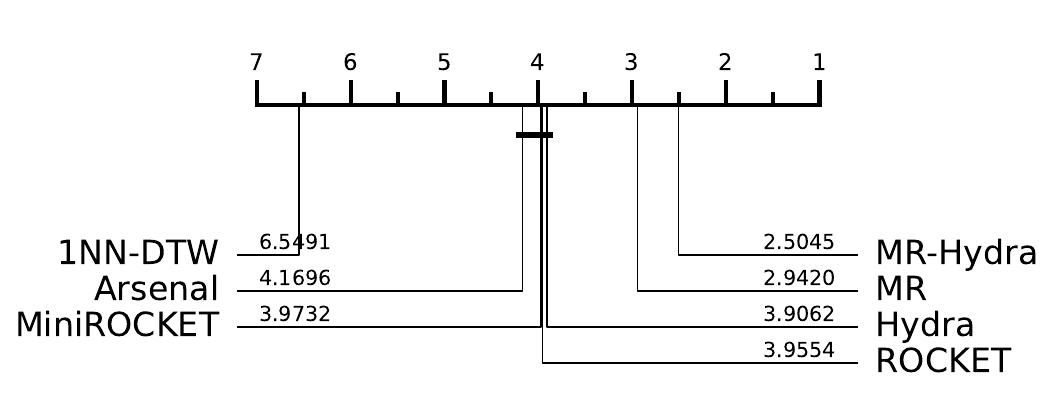}
    \caption{Ranked test accuracy of six convolution based classifiers and the benchmark 1NN-DTW on 112 UCR UTSC problems. Accuracies are averaged over $30$ resamples of train and test splits.}
    \label{fig:convo}
\end{figure}

\begin{figure}[htb]
    \centering
    \begin{tabular}{c c}
    \includegraphics[width=0.5\linewidth]{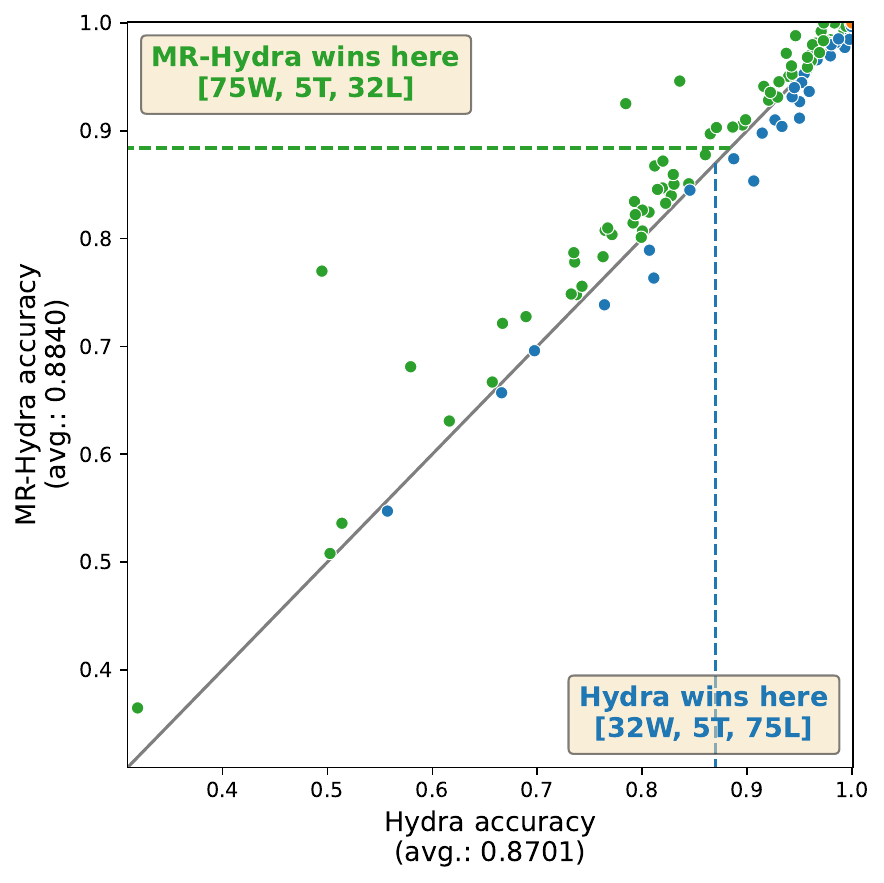} &
    \includegraphics[width=0.5\linewidth]{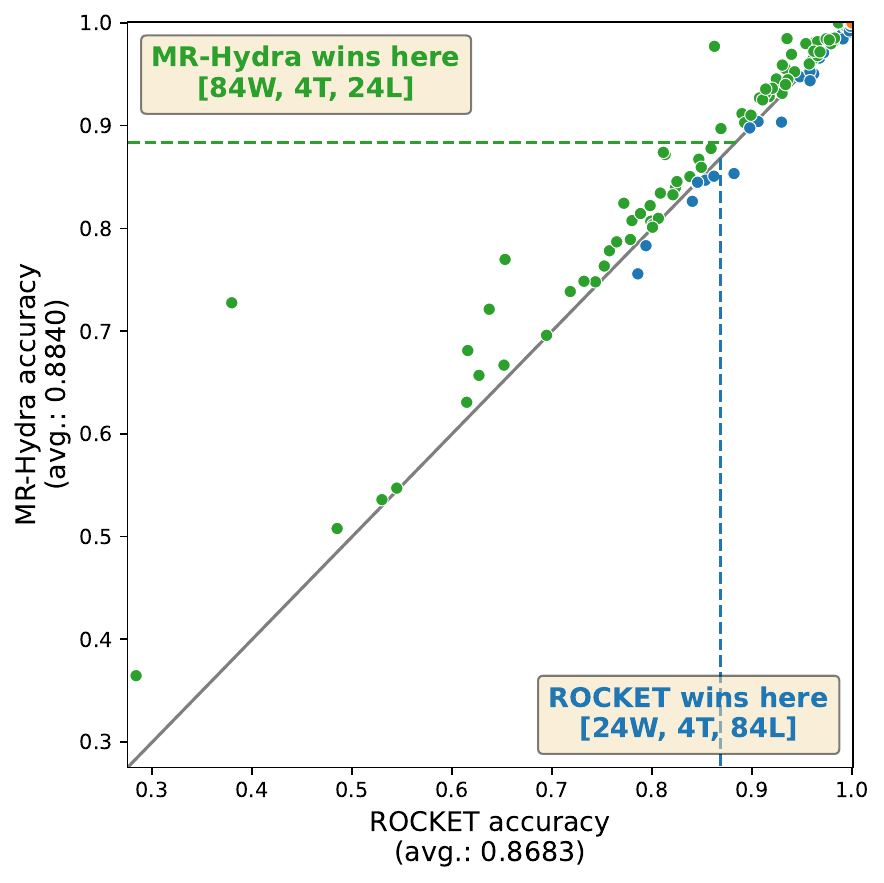}\\
    (a)  & (b)
    \end{tabular}
    \caption{Scatter plot of test accuracies of convolution based classifiers.}
    \label{fig:convo_scatter}
\end{figure}

\begin{table}
    \small
    \caption{Key Differences in approaches from ROCKET to MiniROCKET to MultiROCKET.\label{tab:rocket_family}}
    \begin{centering}
    \begin{tabular}{|p{1.8cm}|>{\centering}p{2.0cm}|>{\centering}p{2.0cm}|>{\centering}p{2.0cm}|>{\centering}p{2.0cm}|}
    \hline
     & \textbf{ROCKET} & \textbf{MiniR} & \textbf{MultiR} & \textbf{Hydra}\tabularnewline
    \hline
    \hline
    kernel length & \{7, 9, 11\} & $9$ & $9$ & $9$\tabularnewline
    \hline
    kernel weights & $N(0, 1)$ & ${-1, 2}$ & ${-1, 2}$ & $N(0, 1)$\tabularnewline
    \hline
    bias & $U(-1,1)$ & from output & from output & none\tabularnewline
    \hline
    dilation & random & fixed (relative to input) & fixed (relative to input) & random \tabularnewline
    \hline
    padding & random & fixed & fixed & always\tabularnewline
    \hline
    pooling operators & MAX, PPV & PPV & PPV, MPV, MIPV, LSPV & Response per Kernel/Group\tabularnewline
    \hline
    1st order difference & no & no & yes & yes\tabularnewline
    \hline
    feature vector size & 20k & 10k & 50k & relative to input\tabularnewline
    \hline
    \end{tabular}
    \par\end{centering}
\end{table}
Figure~\ref{fig:convo} shows the average ranks of the convolution based classifiers.
MultiROCKET-Hydra is the top performer, and is significantly better ranked than the next best, Multi-ROCKET. Table~\ref{tab:ConvolutionBased} and Figure~\ref{fig:convo_scatter}(a) show that the actual difference between the algorithms is small. Progress in the field is demonstrated by Figure~\ref{fig:convo_scatter}(b). MR-Hydra is nearly 2\% better on average than ROCKET, which itself was considered state of the art as recently as 2020~\citep{bagnall20hivecote1}.

\begin{table}[htb]
    \centering
    \caption{Summary performance measures for convolution based classifiers on $30$ resamples of 112 UTSC problems. Best in bold.}
    \begin{tabular}{l|ccccc}
                 & ACC    & BALACC & AUROC  & NLL   & F1 \\ \hline
    MR-Hydra     & \textbf{0.884 (1)}  & \textbf{0.866 (1)}  & 0.913 (2) & 4.181 (2) & \textbf{0.880 (1)} \\
    Multi-Rocket & 0.881 (2)  & 0.863 (2)  & 0.911 (3)  & 4.273 (3)  & 0.878 (2) \\
    Mini-Rocket  & 0.874 (3)  & 0.856 (3)  & 0.906 (4)  & 4.526 (4)  & 0.871 (3) \\
    Hydra        & 0.870 (4)  & 0.850 (4)  & 0.903 (5)  & 4.681 (5)  & 0.866 (4) \\
    ROCKET       & 0.868 (5)  & 0.850 (4)  & 0.903 (5)  & 4.748 (6)  & 0.864 (5) \\
    Arsenal      & 0.866 (6)  & 0.846 (6)  & \textbf{0.925 (1)}  & \textbf{3.317 (1)}  & 0.861 (6) \\
    \hline
    \end{tabular}
    \label{tab:ConvolutionBased}
\end{table}

\subsection{Deep Learning}
\label{sec:deep}

\begin{figure}[tb]
    \centering
    \includegraphics[width=.8\linewidth,trim={4cm 1cm 8cm 1cm},clip]{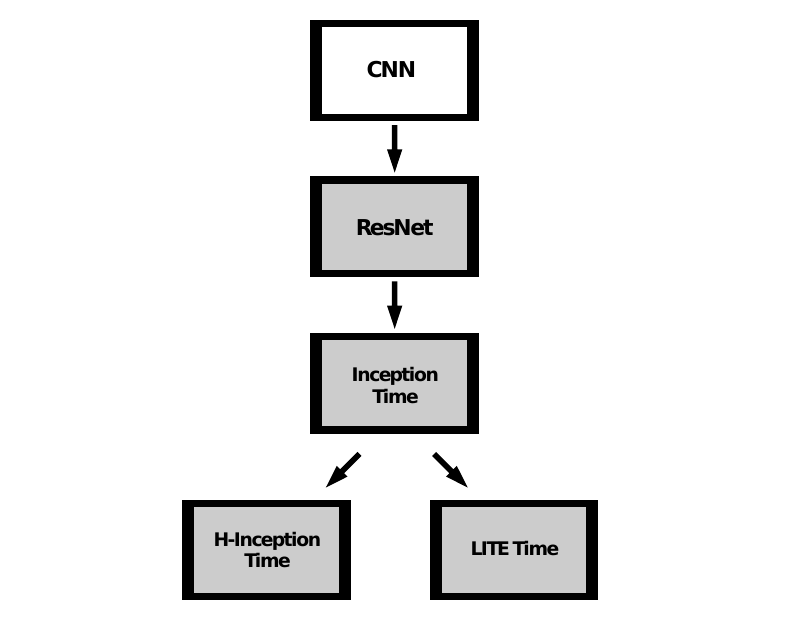}
    \caption{An overview of deep learning classifiers and the relationship between them. Filled algorithms were released after the 2017 bake off~\citep{bagnall17bakeoff}.}
    \label{fig:dl_flow}
\end{figure}

\textit{Deep learning} has been the most active area of TSC research since the bake off in terms of the number of publications. It was thought by many that the impact deep learning had on fields such as vision and speech would be replicated in TSC research. In a paper with \emph{"Finding AlexNet for time series classification"} in the title,~\cite{fawaz20inception} discuss the impact AlexNet had on computer vision and observe that this lesson indicates that \emph{"given the similarities in the data, it is easy to suggest that there is much potential improvement for deep learning in TSC."}. A highly cited survey paper~\citep{fawaz19deep} found that up to that point, ResNet~\citep{wang17fcn} was the most accurate TSC deep learner. Subsequently, the same group proposed InceptionTime~\citep{fawaz20inception}, which was not significantly different to top perfming hybrid algorithms in terms of accuracy~\citep{bagnall20hivecote1}. Since InceptionTime there have been a huge number of deep learning papers proposing TSC algorithms: a recent survey~\citep{foumani23deeplearning} references 246 papers, most of which have been published in the last three years. Table~\ref{tab:deep} summarises some recently proposed deep learning classification algorithms. Without giving specific examples, there are several concerning trends in the deep learning TSC research thread. Most seriously, there is a tendency to perform model selection on test data, i.e. maximize the test accuracy over multiple epochs. This is obviously biased, yet seems to happen even with publications in highly selective venues. Secondly, many papers do not make their source code available. Given all these algorithms are based on standard tools like TensorFlow and PyTorch, this seem inexcusable. Thirdly, they often evaluate on subsets of the archive without any clear rationale as to why. Most are evaluated only on the multivariate archive. Whilst cherry-picking data is questionable, using just MTSC data is not, since deep learning classifiers are usually proposed specifically for MTSC. However, it puts them beyond the scope of this paper. Fourthly, they frequently only compare against other deep learning classifiers, often set up as weak straw men. Finally, they often do not seem to offer any advance on previous research. We have not seen any algorithm that can realistically claim to outperform InceptionTime~\citep{fawaz20inception}, nor its successor H-InceptionTime~\citep{ismail22hybrid}. Because of this, we restrict our attention to five deep learning algorithms. We include a standard Convolutional Neural Network (CNN) implementation as a baseline. We use the same CNN structure as used in the deep learning bake off~\citep{fawaz19deep}. We evaluate ResNet since it was best performing in ~\citep{fawaz19deep}. InceptionTime~\citep{fawaz20inception} is included since it is, to the best our knowledge, best in category for deep learning. We also evaluate two recent extensions of InceptionTime: H-InceptionTime~\citep{ismail22hybrid} and LiteTime~\citep{ismail23lite}. The relation flowchart for deep learning algorithms is shown in Figure~\ref{fig:dl_flow}.

\begin{table}
    \caption{Overview of recently proposed deep learning classifiers.}
    \label{tab:deep}
    \begin{tabular}{|l|c|c|c|c|}
    \hline
    Name & Year & Code & Uni/Mul & Benchmark\tabularnewline
    \hline
    \hline
    Disjoint-CNN & 2021 & y & M & MTCS-26\tabularnewline
    \hline
    Inception-FCN & 2021 & y & U & UTCS-85\tabularnewline
    \hline
    KDCTime & 2022 & n & U & UTCS-113\tabularnewline
    \hline
    Multi-Stage-Att & 2020 & n & M & own\tabularnewline
    \hline
    CT\_CAM & 2020 & n & M & 15 MTCS\tabularnewline
    \hline
    CA-SFCN & 2020 & y & M & 14\tabularnewline
    \hline
    RTFN & 2021 & n & U/M & UTCS-85, MTCS-30\tabularnewline
    \hline
    LAXCAT & 2021 & n & M & 4\tabularnewline
    \hline
    MACNN & 2021 & y & U & UTCS-85\tabularnewline
    \hline
    T2 & 2021 & y & M & own\tabularnewline
    \hline
    GTN & 2021 & y & M & MTCS-13\tabularnewline
    \hline
    TRANS & 2021 & n & M & own\tabularnewline
    \hline
    FMLA & 2022 & n & U & UTCS-85\tabularnewline
    \hline
    AutoTransformer & 2022 & n & U & UTCS-85\tabularnewline
    \hline
    BENDER & 2021 & y & M & 5 EEG\tabularnewline
    \hline
    TST & 2021 & y & M & MTCS-11\tabularnewline
    \hline
    TARNET & 2022 & y & M & MTCS/UCI-34\tabularnewline
    \hline
    \end{tabular}
\end{table}

\subsubsection{Convolution Neural Networks (CNN)}

Convolution Neural Networks (CNN), were first introduced in~\citep{fukushima80neocognitron}, and have gained widespread use in image recognition. Their popularity has increased significantly since AlexNet won the ImageNet competition in 2012. CNNs comprise three types of layers: convolutional, pooling, and fully connected. The convolutional layer slides a filter over a time series, extracting features that are unique to the input. Convolving a one-dimensional filter with the input produces an activation or feature map.  

The result of applying one filter $\omega$ to a time series $T$ at offset $t$ is defined by:
    $$ C_t = f\left( \omega *  T_{t : (t+l)} + b \right) \forall t \in [1 \dots n-l+1]$$

Where the filter $\omega$ is of length $l$, the bias parameter is $b$, and $f$ is a non-linear activation function such as ReLu applied to the result of the convolution. One significant advantage of CNNs is that the filter weights are shared across each convolution, reducing the number of weights that must be learned when compared to fully connected neural networks. But instead of manually setting filter weights, these are learned by the CNN directly from the training data.

As multiple learned filters are applied to the input, each resulting in one activation map of roughly the same size as the input, a pooling layer is used in-between every two convolution layers. A pooling layer, such as Max or Min-pooling, reduces the number of features in each map to i.e. the maximum value, thus providing phase-invariance. After several blocks of convolutional and pooling layers, one or more fully connected layers follow. Finally, a softmax layer with one output neuron per class is used in the final layer.

\subsubsection{Residual Network (ResNet)}

\begin{figure}[htb]
    \centering
    \includegraphics[width=\linewidth, trim={0cm 0cm 0cm 0cm},clip]{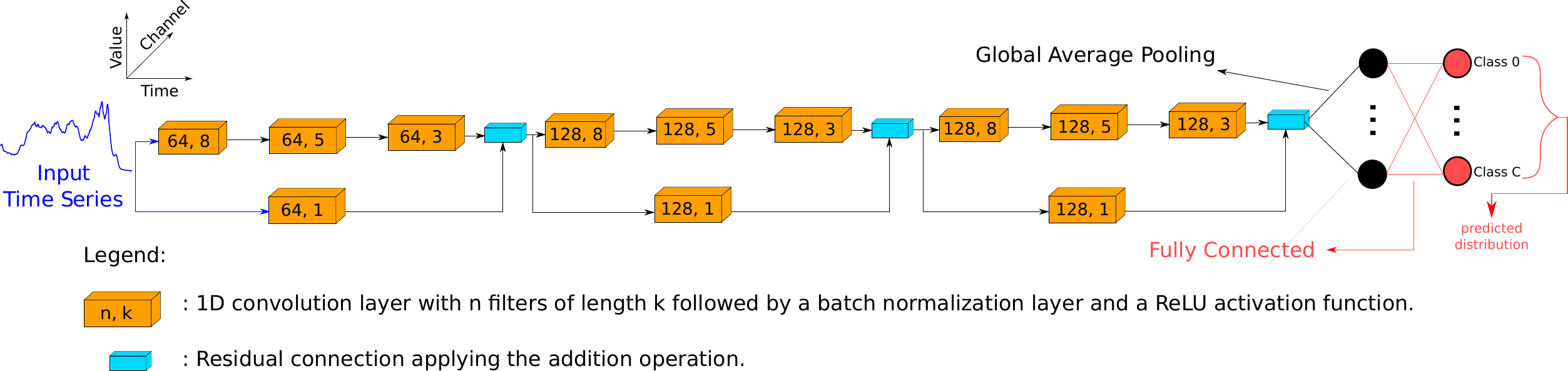}
    \caption{Overview of the ResNet structure, image taken from~\citep{fawaz22hccf} with permission.}
    \label{fig:resnet}
\end{figure}

The Residual Network (ResNet)~\citep{wang17fcn}, is a deep learning architecture that has been successfully adapted for time series analysis. ResNet is composed of three residual blocks, each comprising two main components: (a) three convolutional layers that extract features from the input data followed by batch normalization and a ReLu non-linear activation function, and (b) a shortcut connection that allows the direct propagation of information from earlier layers to later ones. Figure~\ref{fig:resnet} illustrates the structure.

The shortcut connection is designed to mitigate the vanishing gradients problem for deep neural networks, and the convolutional layers extract features from time series data. At the end of the model, the features are passed through one Global Average Pooling (GAP) and one fully-connected softmax layer is used with the number of neurons equal to the number of classes.

\subsubsection{InceptionTime}

InceptionTime is a deep learning model proposed by~\cite{fawaz20inception}. It is an ensemble of five deep learning classifiers, each with the same architecture built on cascading Inception modules~\citep{szegedy15inception}. Diversity is achieved through randomising initial weight values in each of the five models.

\begin{figure}[htb]
    \centering
    \includegraphics[width=\linewidth, trim={0cm 0cm 0cm 0cm},clip]{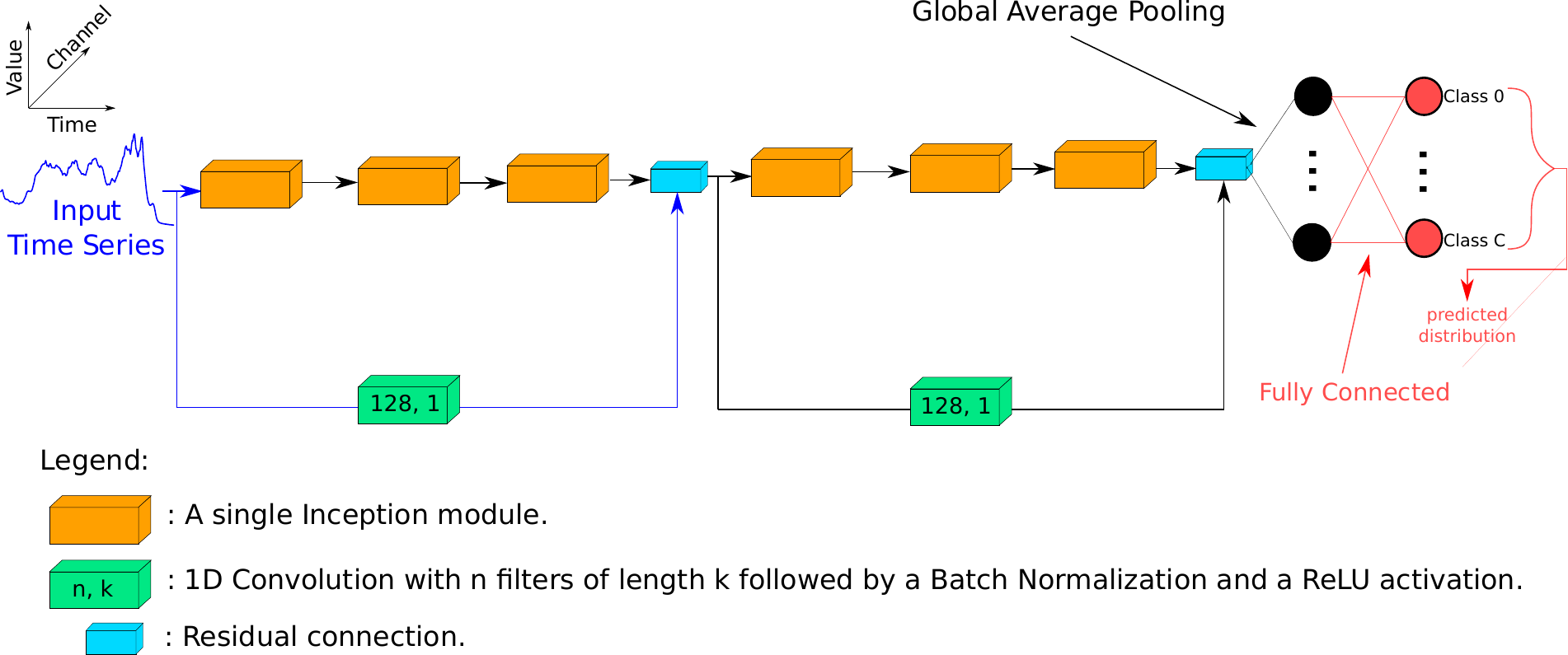}
    \caption{Overview of the InceptionTime structure, image taken from~\citep{fawaz22hccf} with permission}
    \label{fig:it}
\end{figure}

The network, illustrated in Figure~\ref{fig:it}, is composed of two consecutive residual blocks. Where each residual block is composed of three inception modules. The input of the residual block is connected via a shortcut connection to the block's output, to address the vanishing gradient problem. A Global Average Pooling (GAP) layer follows the two residual blocks. Finally, a fully-connected softmax output layer is used with the number of neurons equal to the number of classes. An inception module first applies a \emph{bottleneck layer}, to transform an input multivariate TS to a lower dimensional TS. It then applies multiple convolutional filters of varying kernel sizes, termed multiplexing convolution, to capture temporal features at different scales.

Key design differences to ResNet are ensembling of models, the use of bottleneck layers, multiplexing convolution using varying kernel sizes, and the use of only two residual blocks, as opposed to three in ResNet.

\subsubsection{H-InceptionTime}  

\cite{ismail22hybrid} proposed an extension of InceptionTime that included hand-craft one-dimensional convolution filters to detect very specific patterns in a time series: increasing trends, decreasing trends and peaks. Hybrid Inception (H-Inception) uses the hand-crafted filters in parallel with the first module of the Inception network. Like InceptionTime, H-InceptionTime is an ensemble of five base models. To avoid the need to find the best length of hand-crafted filters, H-InceptionTime chooses different lengths for each hand-crafted filter and used all of them. They found H-InceptionTime provided a small, but significant, improvement over InceptionTime on the 112 UCR datasets.

\subsubsection{LITETime}  

Both ResNet and Inception have approximately 500k trainable parameters and are computationally intensive. In 2023, \cite{ismail23lite} proposed a smaller model for InceptionTime, called Light Inception with boosTing tEchniques (LITE). LITETime uses the DepthWise Separable Convolutions in order to significantly reduce the number of parameters while using boosting techniques to balance the trade off between complexity and performance. These boosting techniques are multiplexing convolution, dilated convolution and hand-crafted-filters. Multiplexing convolution is the approach of applying multiple convolution layers in parallel of different kernel size, motivated from Inception. The usage of dilated convolution is motivated from the fact of it boosting many TSC models in the literature, such as ROCKET. The LITETime ensemble of five base models is not significantly worse than full InceptionTime, but much faster.

\subsubsection{Comparison of Deep Learning Based Approaches}
\begin{figure}[htb]
    \centering
    \includegraphics[width=1\linewidth]{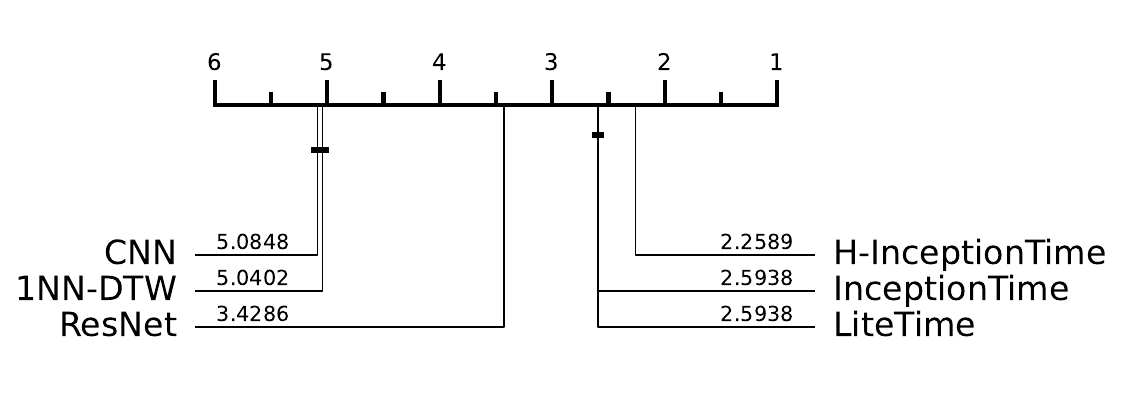}
    \caption{Ranked test accuracy of five deep learning based classifiers and the benchmark 1NN-DTW on 112 UCR UTSC problems. Accuracies are averaged over $30$ resamples of train and test splits.}
    \label{fig:deep}
\end{figure}

Figure~\ref{fig:deep} shows the relative performance of the five deep learning algorithms and 1NN-DTW. Averaged statistics are shown in Table~\ref{tab:DeepLearning}. The results confirm our prior belief: CNN is no better than 1NN-DTW, ResNet is significantly better than CNN, and InceptionTime is significantly better than ResNet. They also confirm recent findings in ~\cite{ismail22hybrid} and~\cite{ismail23lite}: H-InceptionTime gives a small, but significant improvement over InceptionTime and LiteTime is as accurate as InceptionTime but takes a fraction of the time. We choose H-InceptionTime as our best in class.

\begin{table}[htb]
    \centering
    \caption{Summary performance measures for deep learning classifiers on $30$ resamples of 112 UTSC problems. Best in bold.}
    \begin{tabular}{l|ccccc}
     & ACC & BALACC & AUROC & NLL & F1\\ \hline
H-InceptionTime  &	\textbf{0.876 (1)}	&	\textbf{0.861 (1)} &	\textbf{0.959 (1)} & 0.526 (3) &	\textbf{0.873 (1)} \\
InceptionTime    &  0.874 (2)  &  0.859 (2) &	\textbf{0.959 (1)} & 0.515 (2) & 0.872 (2)\\
LiteTime         &  0.869 (3)  &  0.854 (3) &	0.958 (3) &	\textbf{0.476 (1)} & 0.866 (3)\\
ResNet	         &  0.833 (4)  &  0.818 (4) &	0.940 (4) &	1.112 (4) &	0.827 (4)\\
CNN	             &  0.727 (5)  &  0.704 (5) &	0.857 (5) &	2.129 (5) &	0.717 (5)\\
    \hline
    \end{tabular}
    \label{tab:DeepLearning}
\end{table}

\subsection{Hybrid}
\label{sec:hybrid}

The nature of the data and the problem dictate which category of algorithm is most appropriate. The most accurate algorithms on average, with no apriori knowledge of the best approach, combine multiple transformation types in a \textit{hybrid} algorithm. We define a hybrid algorithm as one which by design encompasses or ensembles multiple of the discriminatory representations we have previously described. Some algorithms will naturally include multiple transformation characteristics, but are not classified as hybrid approaches. For example, many interval approaches extract unsupervised summary statistics from the intervals they select, but as the focus of the algorithm is on generating features from intervals we would not consider it a hybrid.

The overall best performing approach in the bake off by a significant margin was the \textbf{Collective of Transformation Ensembles (COTE)~\cite{bagnall15cote}}, which at the time was the only algorithm that explicitly ensembles over different representations. It has been subsequently renamed Flat-COTE due to its structure: it is an ensemble of $35$ time series classifiers built in the time, auto-correlation, power spectrum and shapelet domains. The components of the ST-HESCA~\cite{hills14shapelet} and EE~\cite{lines15elastic} ensembles are pooled with classifiers built on autocorrelation (ACF) and power spectrum (PS) representation. All together, this includes the eight classifiers built on the shapelet transform from ST-HESCA, the 11 elastic distance $1$-NN classifiers from EE and the eight HESCA classifiers built on ACF and PS transformed series. A weighed vote is used to label new cases, with each classifier being weighted using its train set cross-validation accuracy.

\begin{figure}[tb]
    \centering
    \includegraphics[width=1\linewidth,trim={3cm 1.5cm 0.5cm 1cm},clip]{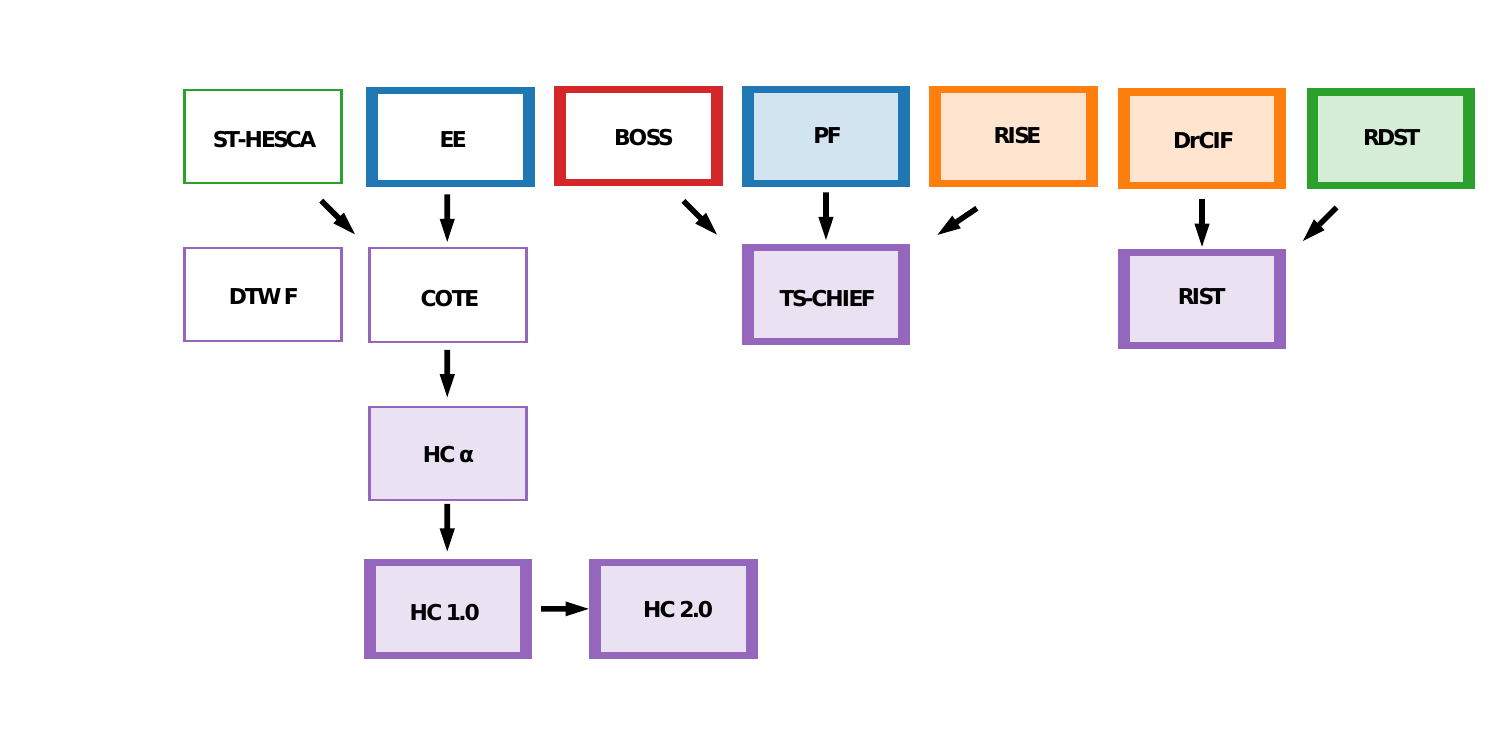}
    \caption{An overview of feature based classifiers and the relationship between them. Filled algorithms were released after the 2017 bake off~\citep{bagnall17bakeoff} and algorithms with a thin border are not included in our experiments.}
    \label{fig:hybrid_flow}
\end{figure}

The COTE family of classifiers has evolved since Flat-COTE, and new hybrid algorithms have been produced following the success shown by ensembling multiple representations. The relation flowchart for hybrid based algorithms is shown in Figure~\ref{fig:hybrid_flow}

\subsubsection{HIVE-COTE (HC$\alpha$)}

\textbf{The Hierarchical Vote Collective of Transformation Ensembles (HIVE-COTE)}~\cite{lines18hive} was proposed to overcome some of the problems with Flat-COTE. This first version of HIVE-COTE, subsequently called HIVE-COTE$_\alpha$ (HC$_\alpha$), is a heterogeneous ensemble containing five modules each from a different representation: EE from the distance based representation; TSF from interval based methods; BOSS from dictionary based approaches and ST-HESCA from shapelet based techniques and the spectral based RISE. The five modules are ensembled using the Cross-validation Accuracy Weighted Probabilistic Ensemble (CAWPE, known at the time as HESCA,~\cite{large19cawpe}). CAWPE employs a tilted probability distribution using exponential weighing of probabilities estimated for each module found through cross-validation on the train data. The weighted probabilities from each module are summed and standardised to produce the HIVE-COTE probability prediction.

\subsubsection{HIVE-COTE version 1 (HC1)}

Whilst state-of-the-art in terms of accuracy, HC$_\alpha$ scales poorly. A range of improvements to make HIVE-COTE more usable were introduced in HIVE-COTE v1.0 (HC1)~\citep{bagnall20hivecote1}. HC1 has four modules instead of the five used in HIVE-COTE$_\alpha$: it drops the computationally intensive EE algorithm without loss of accuracy. BOSS is replaced by the more configurable cBOSS~\citep{middlehurst19scalable}. The improved randomised version of STC~\citep{bostrom17binary} is included with a default limit on the search and the Rotation Forest classifier. TSF and RISE had usability improvements. HC1 is designed to be contractable, in that you can specify a maximum train time.

\subsubsection{HIVE-COTE version 2 (HC2)}

\begin{figure}[htb]
    \centering
    \includegraphics[width=0.9\linewidth]{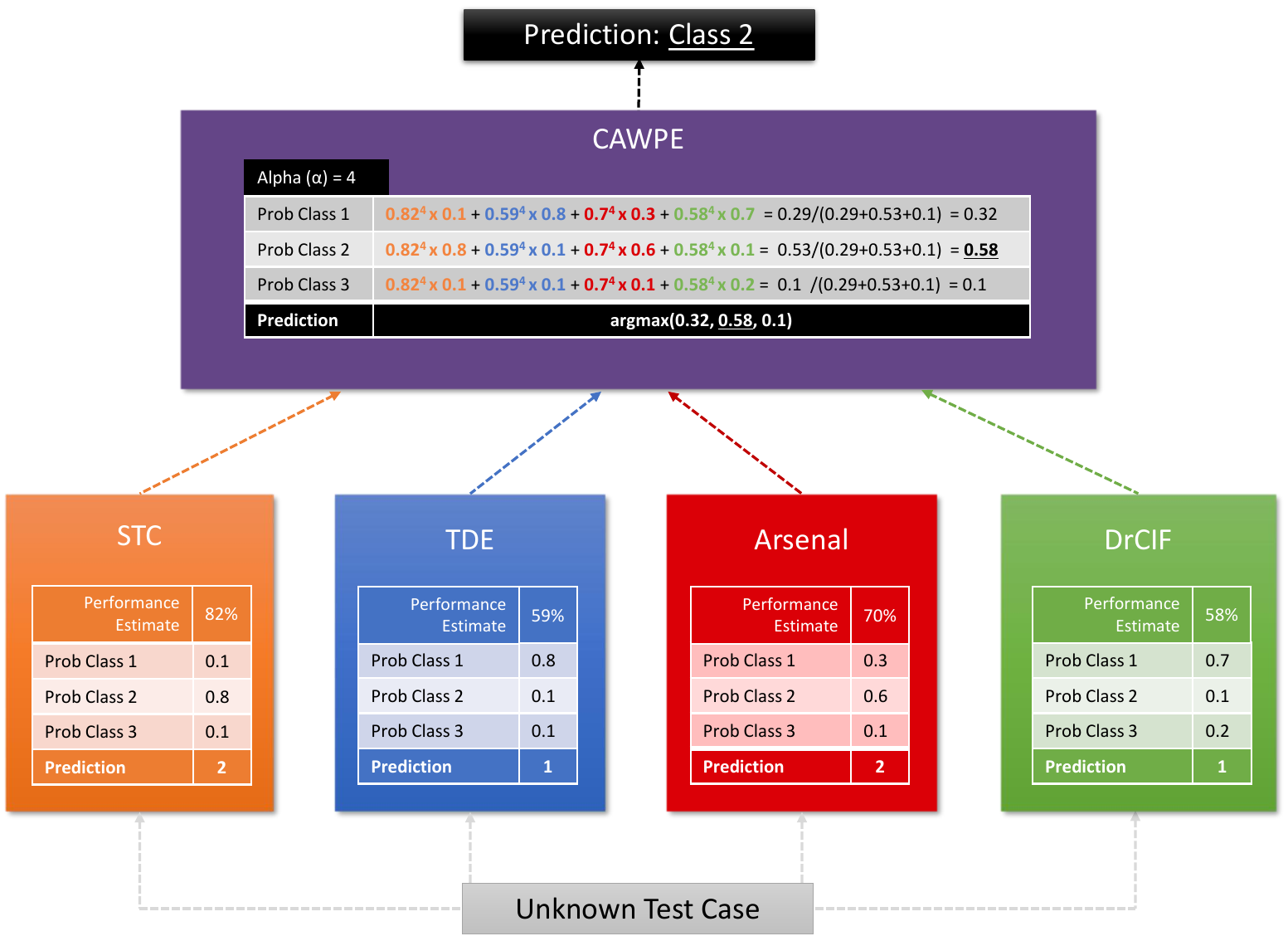}
    \caption{Overview of the HIVE-COTE version 2 ensemble structure.}
    \label{fig:HC2}
\end{figure}

\begin{figure}[tb]
    \centering
    \includegraphics[width=1\linewidth,trim={1.5cm 2cm 3.5cm 1cm},clip]{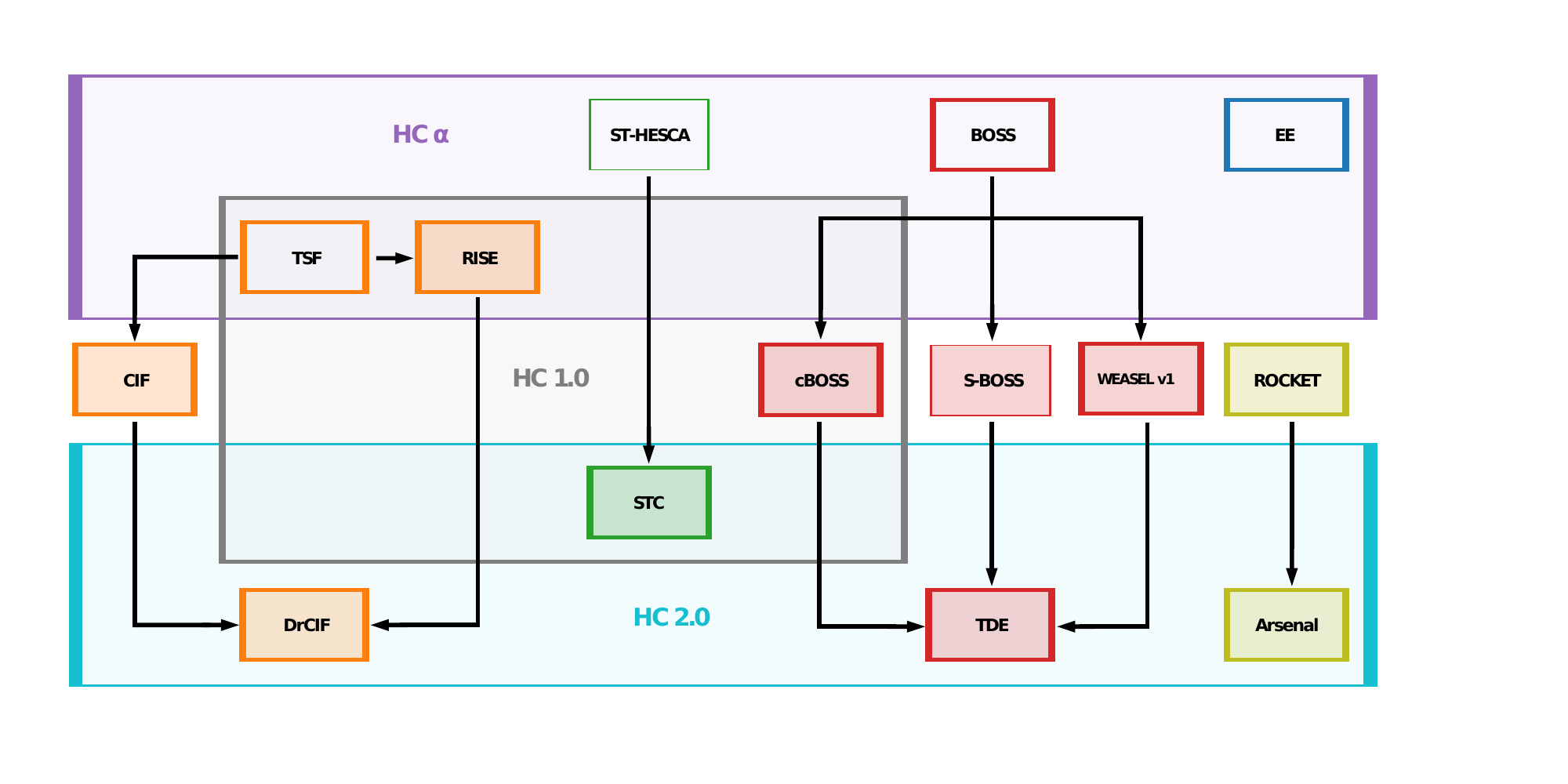}
    \caption{A flowchart displaying the progression of the HIVE-COTE meta ensemble.}
    \label{fig:hc_flow}
\end{figure}

In 2021, HIVE-COTE was again updated to further address scalability issues and reflect recent innovations to individual TSC representations and HIVE-COTE v2.0 (HC2)~\citep{middlehurst21hc2} was proposed. In HC2, RISE, TSF and cBOSS are replaced, with only STC retained. TDE~\citep{middlehurst20temporal} replaces cBOSS as the dictionary classifier. DrCIF replaces both TSF and RISE for the interval and frequency representations. An ensemble of ROCKET classifiers called the Arsenal is introduced as a new convolutional based approach. Estimation of test accuracy via cross-validation is replaced by an adapted form of out-of-bag error, although the final model is still built using all training data. Unlike previous versions, HC2 is capable of classifying multivariate time series. Figure~\ref{fig:HC2} illustrates the structure of HC2, while Figure~\ref{fig:hc_flow} visualises the ensemble members of HIVE-COTE over its evolution.

\subsubsection{TS-CHIEF}

The Time Series Combination of Heterogeneous and Integrated Embedding Forest (TS-CHIEF)~\citep{shifaz20ts-chief} is a homogeneous ensemble where hybrid features are embedded in tree nodes rather than modularised through separate classifiers. The TS-CHIEF comprises an ensemble of trees that embed distance, dictionary, and spectral base features. At each node, a number of splitting criteria from each of these representations are considered. These splits use randomly initialised parameters to help maintain diversity in the ensemble. The dictionary based splits are based on BOSS, distance splits based on EE and interval splits based on RISE. The goal of TS-CHIEF was to obtain the benefits of multiple representations without the massive processing requirement of the original HIVE-COTE.

\subsubsection{Randomised Interval-Shapelet Transformation (RIST)}

The Randomised Interval-Shapelet Transformation (RIST) pipeline is a simpler approach than the previously described hybrids. Rather than constructing an ensemble, RIST concatenates the output of multiple transformations to form a pipeline classifier. RIST uses the transformation portions from the interval based DrCIF~\citep{middlehurst21hc2} and the shapelet based RDST~\citep{guillaume22rdst} algorithms. For both of these transformations, features are extracted from both the base series and multiple series representations. These representations are the first order differences, the periodogram of the series and the series autoregression coefficients. After concatenating the output, these features are then used to build an Extra Trees~\citep{geurts06extremely} classifier. The aim of RIST is to provide a simple and relatively efficient hybrid algorithm which can be applied to both classification and extrinsic regression tasks.

\subsubsection{Comparison of Hybrid Approaches}

\begin{figure}[htb]
    \centering
    \includegraphics[width=1\linewidth]{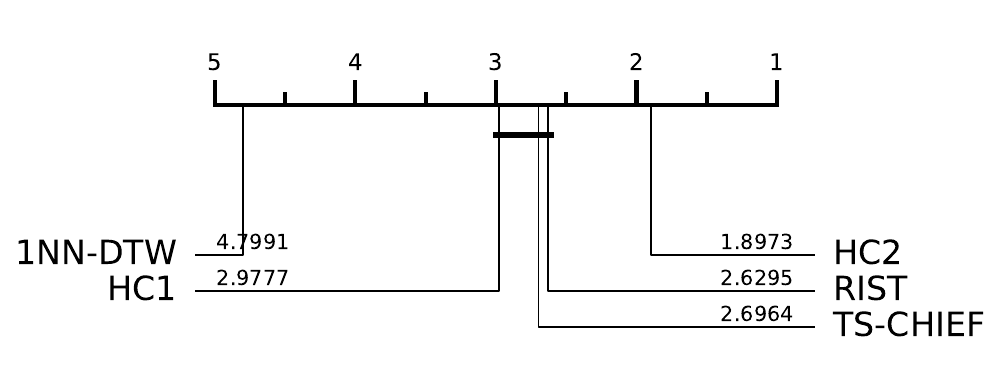}
    \caption{Ranked test accuracy of hybrid classifiers and the benchmark 1NN-DTW on 112 UCR UTSC problems. Accuracies are averaged over $30$ resamples of train and test splits.}
    \label{fig:hybrid}
\end{figure}

\begin{figure}[htb]
    \centering
    \begin{tabular}{c c}
    \includegraphics[width=0.5\linewidth]{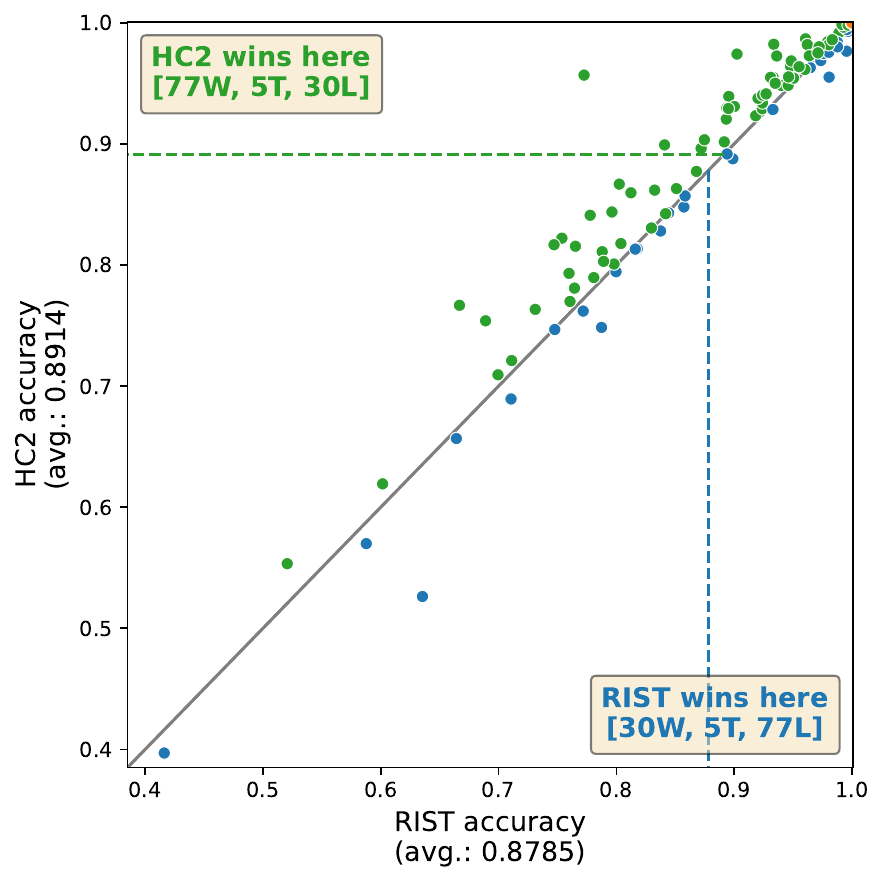} &
    \includegraphics[width=0.5\linewidth]{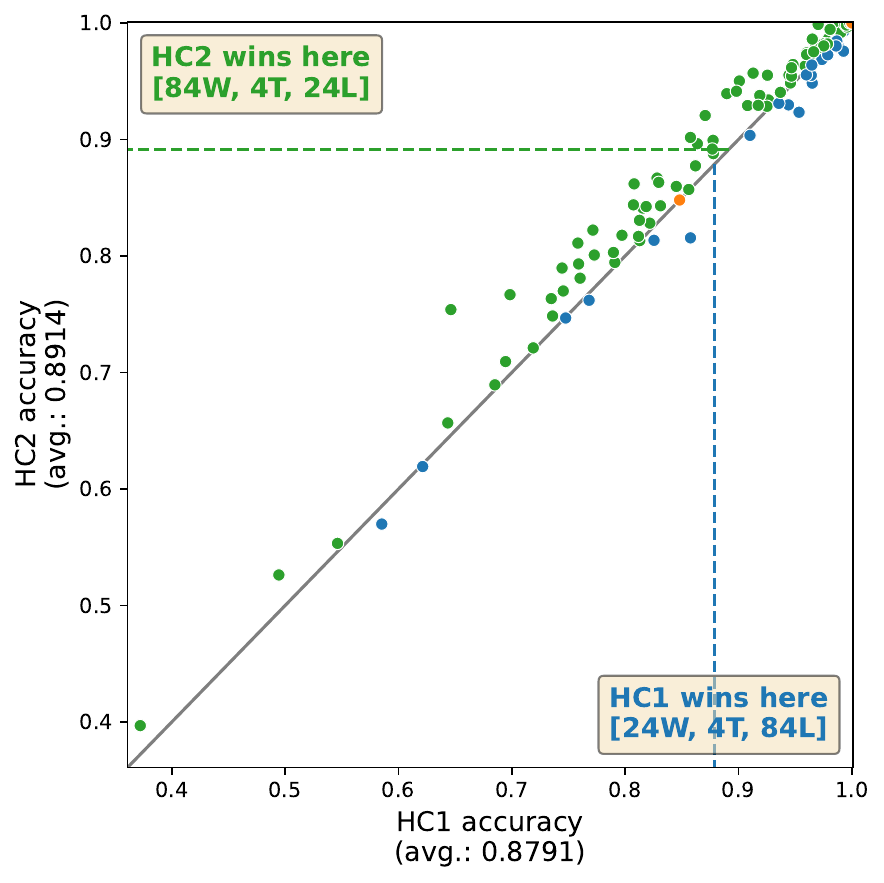}
    \end{tabular}
    \caption{Scatter plot of test accuracies of hybrid classifiers.}
    \label{fig:hybrid_scatter}
\end{figure}

Figure~\ref{fig:hybrid} shows that HC2 is significantly more accurate than the other three classifiers in terms of accuracy. This is true of all five metrics shown in Table~\ref{tab:Hybrid}. The scatter plots in Figure~\ref{fig:hybrid_scatter} shows it is consistently better than HC1 and RIST.

\begin{table}[htb]
    \centering
    \caption{Summary performance measures for hybrid classifiers on $30$ resamples of 112 UTSC problems. Best in bold.}
    \begin{tabular}{l|ccccc}
     & ACC & BALACC & AUROC & NLL   & F1\\ \hline
    HC2      & \textbf{0.891 (1)}  & \textbf{0.871 (1)}  & \textbf{0.968 (1)}  & \textbf{0.365 (1)} & \textbf{0.886 (1)}\\
    RIST     & 0.878 (3) & 0.854 (3)  & 0.966 (2)  & 0.482 (4) & 0.872 (4)\\
    TS-CHIEF & 0.878 (3) & 0.857 (2)  & 0.960 (4)  & 0.439 (2) & 0.873 (2)\\
    HC1      & 0.879 (2) & 0.854 (3)  & 0.964 (3)  & 0.471 (3) & 0.873 (2)\\
    \hline
    \end{tabular}
    \label{tab:Hybrid}
\end{table}

\section{Results}
\label{sec:results}

\begin{figure}[b]
	\centering
    \begin{tabular}{cc}
       \includegraphics[width=0.5\linewidth]{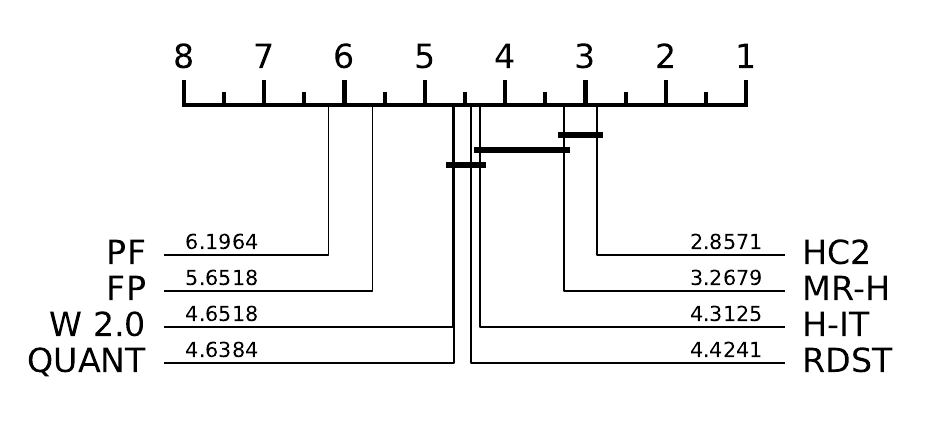} &
       \includegraphics[width=0.5\linewidth]{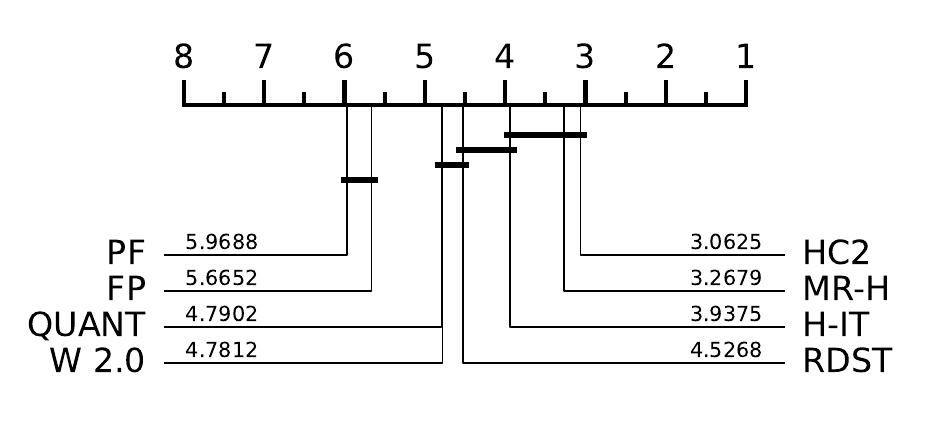}  \\
       (a) Accuracy &  (b) Balanced Accuracy \\
       \includegraphics[width=0.5\linewidth]{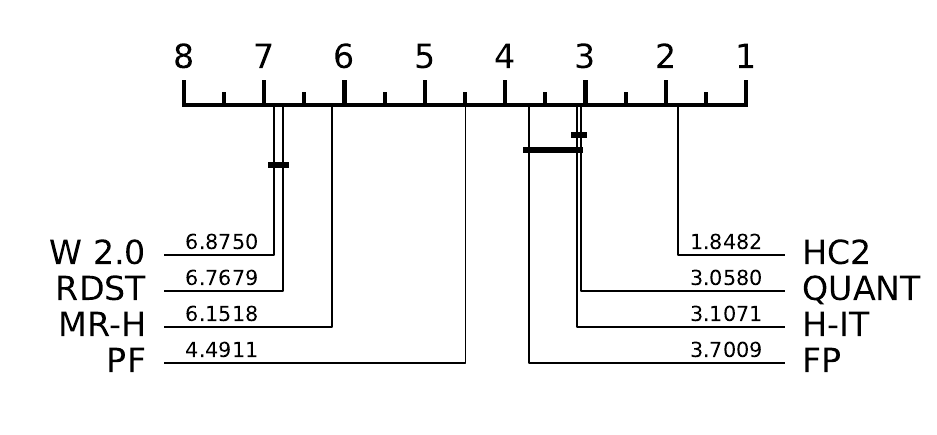} &
       \includegraphics[width=0.5\linewidth]{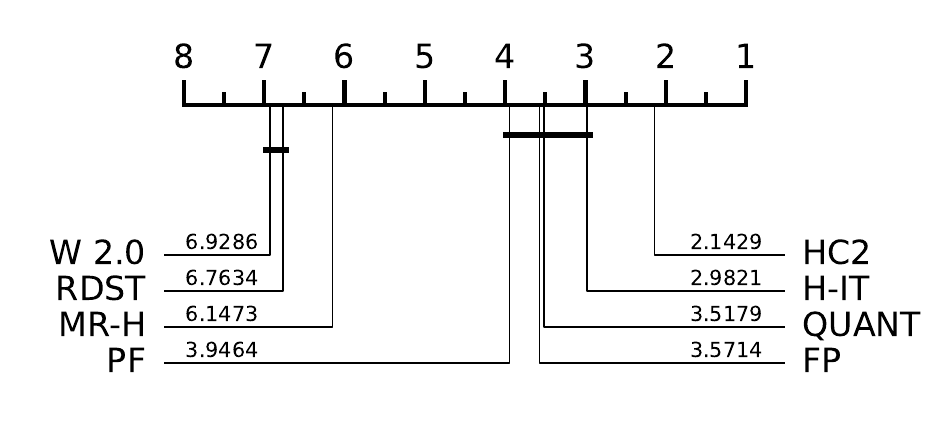} \\
        (c) AUROC & (d) Negative Log Likelihood \\
       \end{tabular}
       \caption{Averaged ranked performance statistics for eight best of category algorithms on 112 UCR UTSC problems. Statistics are averaged over $30$ resamples of train and test splits. Names shortened for clarity: FP is FreshPrince, W 2.0 is Weasel 2.0, MR-H is MR-Hydra and H-IT is H-InceptionTime.}
       \label{fig:sota}
\end{figure}

To keep the analysis tractable, we restrict further analysis of performance to the best classifier in each of the eight categories. Further results tables and figures are available in Appendix~\ref{app:results}, with all results files available on the accompanying website and can be accessed directly accessible in code withg aeon. Figure~\ref{fig:sota} shows the ranking of these classifiers on the 112 UCR data for $30$ resamples of train/tests splits. HC2 and MR-Hydra are the top performing algorithms in terms of accuracy. There is no significant difference between QUANT, H-IT, RDST and WEASEL 2.0 in terms of test accuracy. HC2 is best performing with AUROC and NLL measures. H-IT performs better with balanced accuracy and NLL.

The AUROC and NLL need to be interpreted in context: Weasel 2.0, RDST and MR-Hydra use classifiers that only produce 0/1 predictions. This means they will inevitably perform poorly on AUROC and NLL.

\begin{figure}[htb]
    \centering
            \begin{tabular}{c c}
    \includegraphics[width=0.5\linewidth]{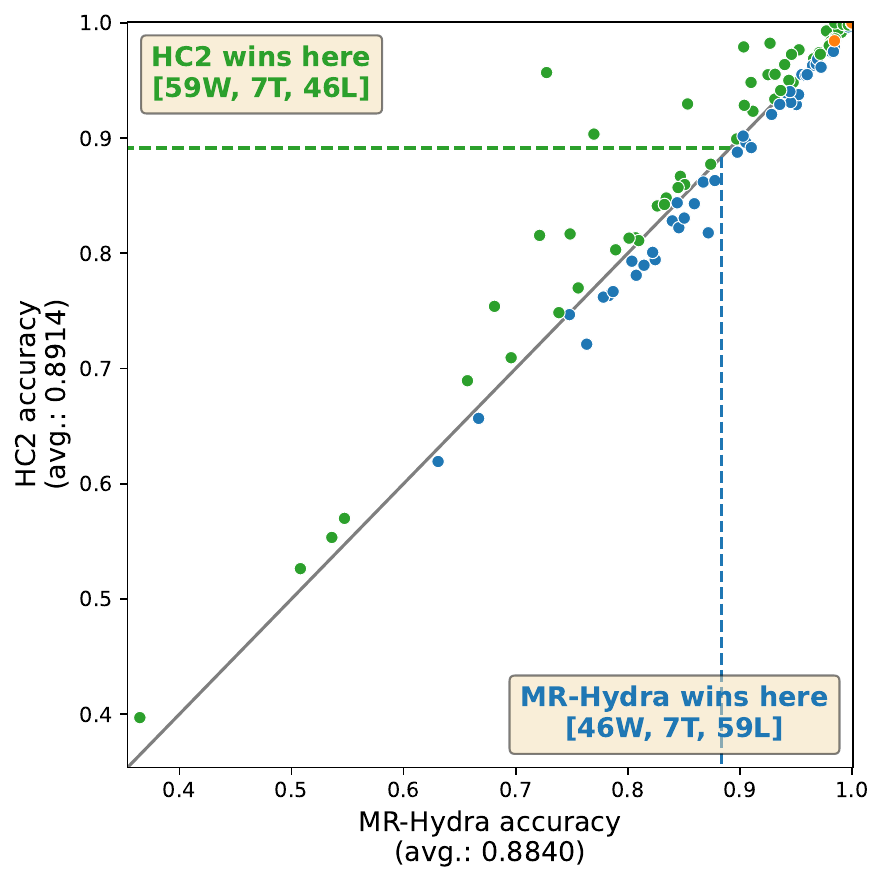} &
    \includegraphics[width=0.5\linewidth]{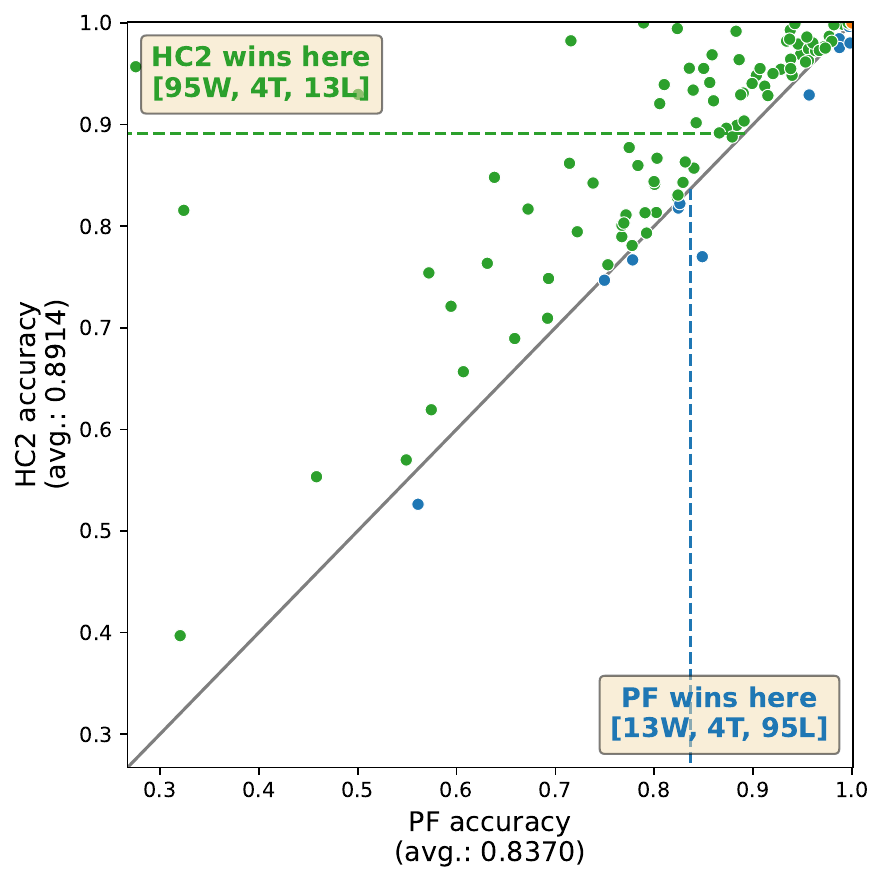}
    \end{tabular}
    \caption{Scatter plot of test accuracies of state of the art classifiers.}
    \label{fig:sota_scatter}
\end{figure}

For context, Figure~\ref{fig:sota_scatter} shows the scatter plot of HC2 against the next best (MR-Hydra) and the worst performing (PF). It is worth reiterating that PF is significantly better than both EE and 1-NN DTW, both of which were considered state of the art until recently.

For convenience, Table~\ref{tab:sota} summarises the summary statistics presented in Section~\ref{sec:algos}. HC2 is on average about 0.5\% more accurate than MR-Hydra, over 6\% more accurate than PF and over 12\% more accurate than 1NN-DTW.

\begin{table}[htb]
    \centering
    \caption{Summary performance measures for best in category classifiers on $30$ resamples of 112 UTSC problems. Best in bold.}
    \begin{tabular}{l|ccccc}
                & ACC   & BALACC & AUROC & NLL & F1\\ \hline
HC2             & \textbf{0.891 (1)} & \textbf{0.871 (1)} & \textbf{0.968 (1)} & \textbf{0.365 (1)} & \textbf{0.886 (1)} \\
MR-Hydra (MR-H)        & 0.884 (2) & 0.866 (2) & 0.913 (6) & 4.181 (6) & 0.880 (2)\\
RDST                   & 0.876 (3) & 0.856 (4) & 0.907 (7) & 4.457 (7) & 0.872 (3)\\
H-InceptionTime (H-IT) & 0.874 (4) & 0.859 (3) & 0.959 (3) & 0.526 (4) & 0.872 (3)\\
WEASEL 2.0 (W 2.0)     & 0.874 (4) & 0.853 (5) & 0.905 (8) & 4.547 (8) & 0.869 (5)\\
QUANT                  & 0.867 (6) & 0.845 (6) & 0.962 (2) & 0.497 (2) & 0.862 (6)\\
FreshPRINCE (FP)       & 0.855 (7) & 0.834 (7) & 0.958 (4) & 0.501 (3) & 0.850 (7)\\
Proximity Forest (PF)  & 0.837 (8) & 0.819 (8) & 0.942 (5) & 0.692 (5) & 0.833 (8)\\
1NN-DTW                & 0.756 (9) & 0.739 (9) & 0.820 (9)  & 8.812 (9) & 0.752 (9)\\
    \hline     \end{tabular}
    \label{tab:sota}
\end{table}

\subsection{Performance on New TSC Datasets}

\begin{figure}[!ht]
	\centering
    \begin{tabular}{cc}
       \includegraphics[width=0.5\linewidth]{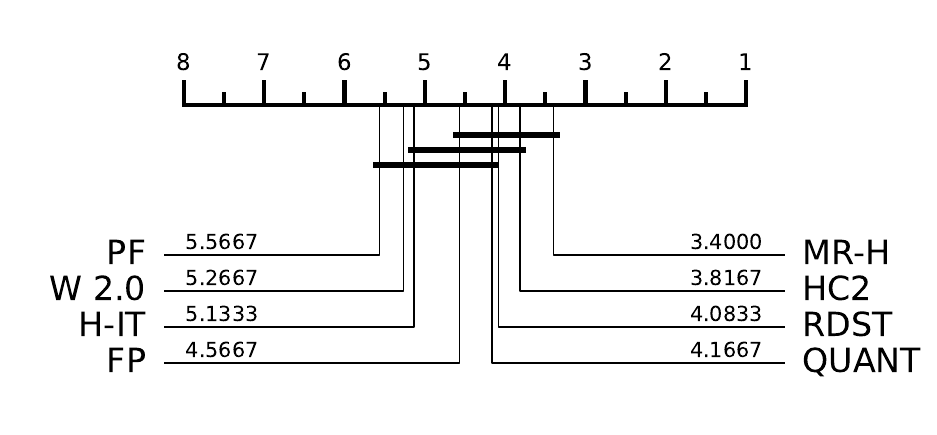}  &
       \includegraphics[width=0.5\linewidth]{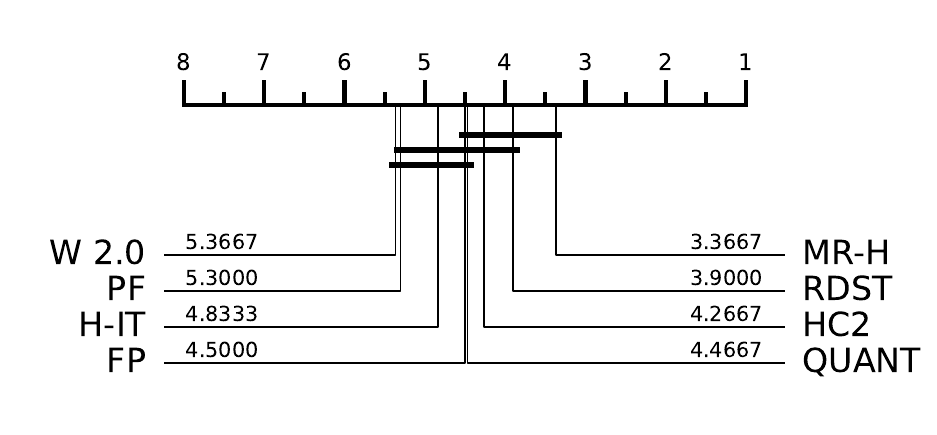}  \\
       (a) Accuracy &  (b) Balanced Accuracy \\
       \end{tabular}
       \caption{Averaged ranked performance statistics for eight best of category algorithms on 30 new UTSC problems. Statistics are averaged over $30$ resamples of train and test splits.}
       \label{fig:new_subset}
\end{figure}

\begin{figure}[!ht]
	\centering
    \begin{tabular}{cc}
       \includegraphics[width=0.5\linewidth]{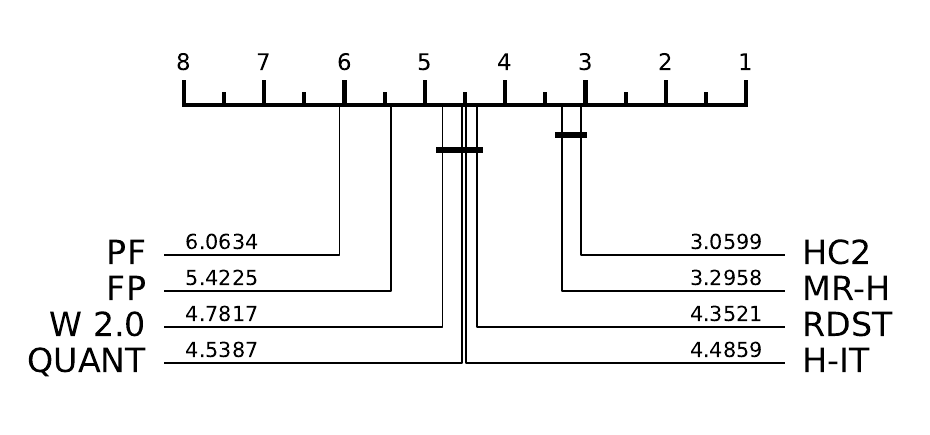} &
       \includegraphics[width=0.5\linewidth]{img/BiC_142_accuracy_critical_difference.pdf}  \\
       (a) Accuracy &  (b) Balanced Accuracy \\
       \end{tabular}
       \caption{Averaged ranked performance statistics for eight best of category algorithms on 142 TSC problems. Statistics are averaged over $30$ resamples of train and test splits.}
       \label{fig:new}
\end{figure}

\begin{figure}[!ht]
	\centering
    \includegraphics[height=0.60\linewidth, angle=90, trim={0cm 0cm 0cm 0cm},clip]{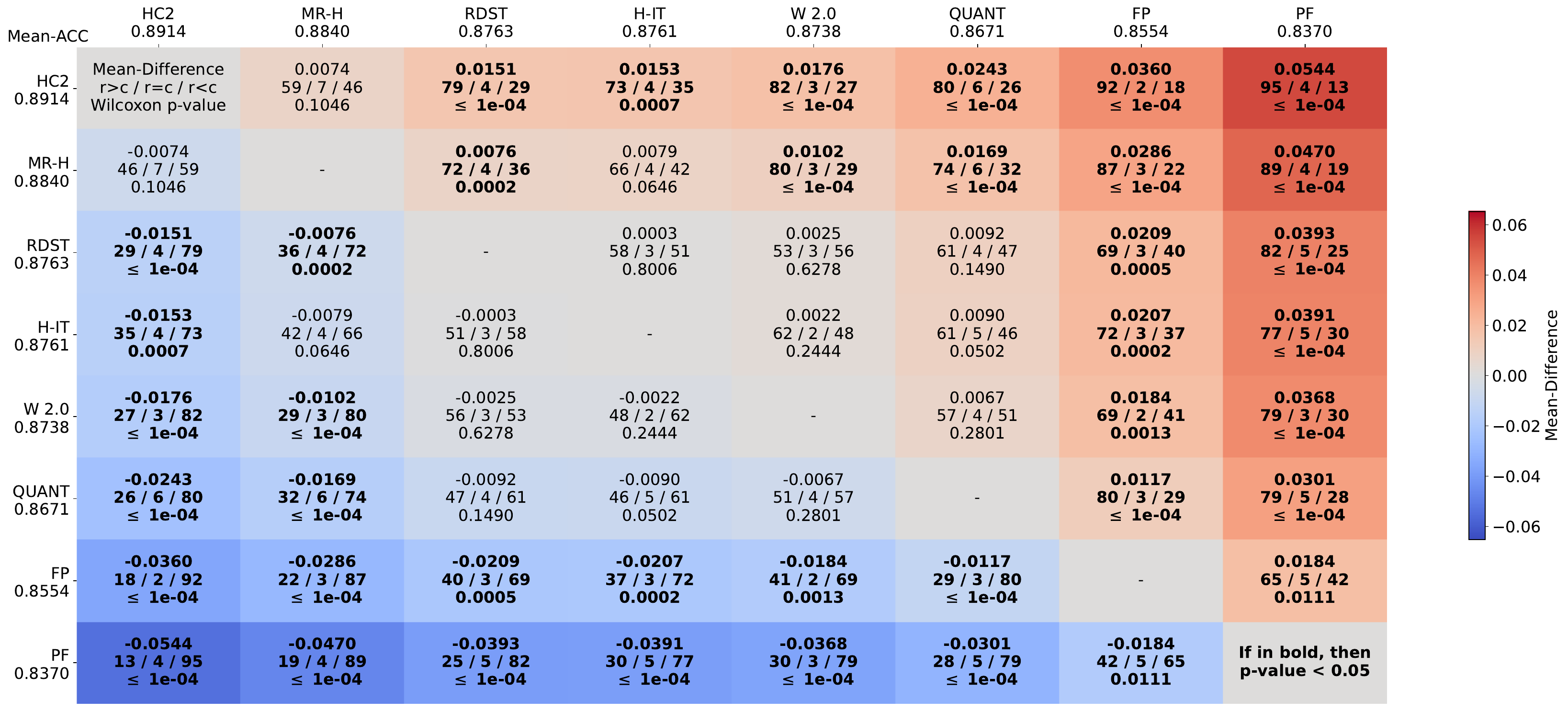}
    \caption{Summary performance statistics for eight classifiers on 112 datasets, generated using the multiple comparison matrix (MCM). The MCM shows pairwise comparisons. Each cell shows the mean difference in accuracy, wins/draws/losses, and Wilcoxon p-value for two comparates.}
    \label{fig:heatmap112}
\end{figure}

HC2 performs the best on the 112 datasets. However, HIVE-COTE has been in development for over five years, and all advances were judged by evaluation on these datasets. As acknowledged in~\cite{middlehurst21hc2}, there is always the risk of the introduction of subconscious bias in the design decisions that lead to the new algorithms. To counter this, we have assembled $30$ new datasets, as described in Section~\ref{sec:new_data}. Figure~\ref{fig:new_subset} shows the ranks for the eight best of category on these data sets. The top clique for accuracy contains MR-Hydra, HC2, RDST, QUANT and FreshPRINCE. The results for balanced accuracy are similar.  Figure~\ref{fig:new} shows the results for the combined 142 datasets. The performance of HC2 and MR-Hydra is very similar, and they are in a clique that is significantly better than the other six classifiers.

~\cite{ismail23multiple} noted that critical difference diagrams (CD) can be deceptive and lack stability, with the relative ordering being highly sensitive to the selection of comparates included in the comparisons. This sensitivity renders them susceptible to inadvertent manipulation. To circumvent this problem, they propose a bespoke pairwise comparison tool, called multiple comparative matrix (MCM)\footnote{\url{https://github.com/MSD-IRIMAS/Multi_Comparison_Matrix}}. It shows pairwise comparisons between all comparates, and includes difference in average scores, wins/draws/losses, and Wilcoxon p-values. Colors of the heat map represent mean differences in scores. Red indicates that the comparate in the row wins by more on average than the comparate in the column. Bold text indicates that the difference in significant. Figure~\ref{fig:heatmap112} summarises the performance of the eight classifiers on 112 datasets using the MCM, with comparisons to the 30 new datasets and 142 datasets available in Appendix~\ref{app:results}. A notable observation arises when comparing the CD on accuracy in Figure~\ref{fig:sota} to this MCM on the 112 UCR UTSC. The rankings of WEASEL 2.0, QUANT, H-IT and RDST are deceptive. Despite WEASEL 2.0 demonstrating more pairwise wins compared to RDST or QUANT in the MCM, its ranking appears higher (worse) than both in the CD when all 8 comparates are taken into account. In addition, H-IT has less pairwise wins than RDST in the MCM, yet shows the lower (better) rank in the CD.

\section{Analysis}
\label{Sec:analysis}

Relative performance on test suites is important when evaluating classifiers, but it does not necessarily generalise to new problems. There will be problem domains and specific applications where different classifiers will be the most effective. Furthermore, characteristics such as the variability in performance and the run time complexity of algorithms are also of great interest to the practitioner.

We model the approach used in~\cite{bagnall17bakeoff} by comparing performance by data characteristics using all 142 datasets.  Tables~\ref{tab:length}, ~\ref{tab:train} and~\ref{tab:classes} break performance down by series length, train set size and number of classes. HC2 and MR-Hydra are first or second on average in each category. HC2 seems to do better with longer series. MR-Hydra performs better with larger train set sizes. Table~\ref{tab:problem} breaks down performance by problem type. HC2 and MR-Hydra are the top two ranked in all categories except MOTION. MR-Hydra does particularly well on image outlines, whereas HC2 excels at electric devices and spectrograms.

\begin{table}[htb]
    \centering
    \caption{Average accuracy rank of classifiers  on $30$ resamples of  142 TSC problems split by series length.}
    \begin{tabular}{l|cccc}
     & 1-199 (44) & 200-499 (44) & 500-999 (27) & 1000+ (27) \\ \hline
HC2          & \textbf{3.28 (1)} &          3.08 (2) & \textbf{3.09 (1)} & \textbf{2.63 (1)} \\
MR-Hydra (MR-H)  &          3.43 (2) & \textbf{2.89 (1)} &          3.48 (3) &          3.56 (2) \\
RDST         &          4.72 (5) &          3.66 (3) &          5.02 (6) &          4.22 (4) \\
QUANT        &          4.15 (3) &          4.94 (6) &          4.85 (4) &          4.20 (3) \\
H-InceptionTime &          4.86 (6) &          4.72 (5) &          3.19 (2) &          4.80 (5) \\
WEASEL 2.0 (W 2.0)  &          4.59 (4) &          4.51 (4) &          4.98 (5) &          5.33 (7) \\
FreshPrince (FP)  &          4.90 (7) &          6.19 (8) &          5.26 (7) &          5.19 (6) \\
Proximity Forest (PF) &          6.07 (8) &          6.01 (7) &          6.13 (8) &          6.07 (8) \\    \hline
    \end{tabular}
    \label{tab:length}
\end{table}

\begin{table}[htb]
    \centering
    \caption{Average accuracy rank of classifiers  on $30$ resamples of  142 TSC problems split by train set size.}
    \begin{tabular}{l|cccc}
     & 1-99 (42) & 100-299 (32) & 300-699 (47) & 700+ (21) \\ \hline

HC2          & \textbf{2.70 (1)} & \textbf{2.81 (1)} &          3.26 (2) &          3.71 (2) \\
MR-Hydra (MR-H)    &          3.68 (2) &          3.39 (2) & \textbf{3.17 (1)} & \textbf{2.67 (1)} \\
RDST         &          4.33 (3) &          3.45 (3) &          4.79 (5) &          4.79 (6) \\
QUANT        &          4.58 (5) &          4.80 (5) &          4.50 (4) &          4.14 (4) \\
H-InceptionTime &          5.20 (6) &          4.20 (4) &          4.32 (3) &          3.86 (3) \\
WEASEL 2.0 (W 2.0)   &          4.35 (4) &          4.81 (6) &          4.86 (6) &          5.43 (7) \\
FreshPrince (FP) &          5.48 (7) &          6.28 (8) &          5.15 (7) &          4.62 (5) \\
Proximity Forest (PF) &          5.68 (8) &          6.25 (7) &          5.96 (8) &          6.79 (8) \\
    \hline
    \end{tabular}
    \label{tab:train}
\end{table}

\begin{table}[htb]
    \centering
    \caption{Average accuracy rank of classifiers  on $30$ resamples of  142 TSC problems split by number of classes.}
    \begin{tabular}{l|cccc}
     & 2 (50) & 3-5 (38) & 6-10 (30) & 11+ (24) \\ \hline

HC2             &          3.38 (2) & \textbf{2.79 (1)} & \textbf{2.85 (1)} &          3.08 (2) \\
MR-Hydra (MR-H)       & \textbf{3.20 (1)} &          3.62 (2) &          3.68 (2) & \textbf{2.50 (1)} \\
RDST            &          4.81 (6) &          4.34 (3) &          4.08 (3) &          3.75 (4) \\
QUANT           &          4.11 (3) &          4.59 (4) &          4.52 (4) &          5.38 (6) \\
H-InceptionTime    &          4.57 (4) &          4.78 (5) &          4.97 (6) &          3.25 (3) \\
WEASEL 2.0 (W 2.0)     &          4.75 (5) &          4.79 (6) &          4.85 (5) &          4.75 (5) \\
FreshPrince (FP) &          4.95 (7) &          5.33 (7) &          5.40 (7) &          6.58 (7) \\
Proximity Forest (PF) &          6.23 (8) &          5.76 (8) &          5.65 (8) &          6.71 (8) \\
    \hline
    \end{tabular}
    \label{tab:classes}
\end{table}

\begin{table}[htb]
    \caption{Average accuracy rank of classifiers  on $30$ resamples of  142 TSC problems split by problem type.}
\begin{tabular}{l|cccc}
\toprule
 Type &            DEVICE &               ECG &               HAR &             IMAGE \\
\midrule
HC2           & \textbf{2.42 (1)} & \textbf{2.14 (1)} &          2.90 (2) &          3.52 (2) \\
MR-Hydra (MR-H)     &          2.83 (2) &          2.43 (2) & \textbf{2.40 (1)} & \textbf{3.09 (1)} \\
RDST          &          4.17 (5) &          4.29 (3) &          3.26 (3) &          3.98 (4) \\
QUANT         &          3.88 (3) &          5.00 (6) &          5.93 (7) &          4.74 (6) \\
H-InceptionTime  &          3.96 (4) &          4.57 (5) &          4.19 (4) &          4.33 (5) \\
WEASEL 2.0 (W 2.0)   &          6.00 (7) &          4.43 (4) &          4.81 (5) &          3.86 (3) \\
FreshPrince (FP) &          5.42 (6) &          5.71 (7) &          6.74 (8) &          5.97 (7) \\
Proximity Forest (PF) &          7.33 (8) &          7.43 (8) &          5.76 (6) &          6.50 (8) \\
    \hline
    \end{tabular}

    \begin{tabular}{l|cccc}
\toprule
 Type &            MOTION &            SENSOR &         SIMULATED &           SPECTRO \\
\midrule
HC2           &          3.22 (2) &          3.60 (2) & \textbf{2.61 (1)} & \textbf{2.38 (1)} \\
MR-Hydra (MR-H)    &          4.33 (3) & \textbf{3.54 (1)} &          3.50 (2) &          3.79 (2) \\
RDST          &          4.67 (5) &          4.69 (4) &          4.56 (5) &          5.29 (6) \\
QUANT         & \textbf{2.33 (1)} &          5.27 (7) &          3.78 (3) &          6.17 (8) \\
H-InceptionTime  &          6.11 (8) &          5.38 (8) &          4.44 (4) &          5.62 (7) \\
WEASEL 2.0 (W 2.0)   &          4.44 (4) &          3.83 (3) &          5.00 (6) &          4.00 (4) \\
FreshPrince   &          5.11 (6) &          4.90 (6) &          5.44 (7) &          4.83 (5) \\
Proximity Forest (PF) &          5.78 (7) &          4.79 (5) &          6.67 (8) &          3.92 (3) \\

    \bottomrule
    \hline
    \end{tabular}
    \label{tab:problem}
\end{table}

\begin{figure}[!ht]
	\centering
    \includegraphics[width=\linewidth,trim={2cm 4.5cm 2cm 4.2cm},clip]{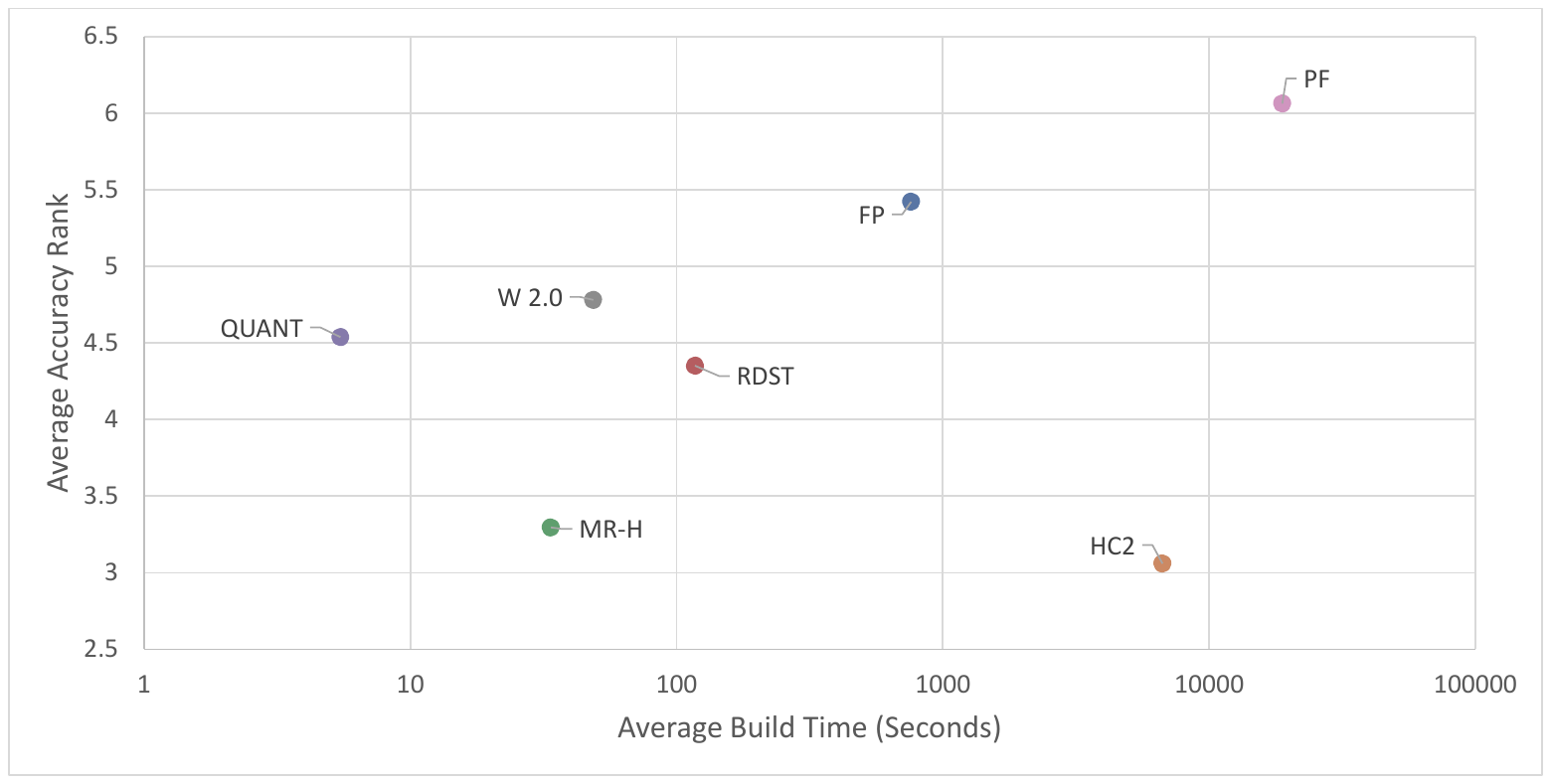}
    \caption{Averaged rank on accuracy against train time on 142 UTSC problems.}
    \label{fig:time}
\end{figure}

Run time is clearly an important consideration. The speed of QUANT and the ROCKET family of classifiers is a significant feature. It was stated in the bake off that {\em``[a]n algorithm that is faster than [the current state of the art] but not significantly less accurate would be a genuine advance in
the field``}. ROCKET and the subsequent refinements fulfil this criteria and represent an important advance. Table~\ref{tab:time} shows the total train time for classifiers on the $142$ problems and Figure~\ref{fig:time} shows the plot of rank against train time (on a logarithmic scale). We do not include H-IT in these measurements because it was run on two different types of GPU, whereas the other algorithms were all trained on the same CPU (Intel Xeon Gold 5220R 2.2GHz). HC2 is clearly much slower than MR-Hydra. This is at least in parts the result of the configuration and implementation of HC2. For example, TDE, a component of HC2, is not optimised using numba\footnote{\url{http://numba.pydata.org/}}. Nevertheless, there is no doubt that MR-Hydra offers a good accuracy/train time trade off: it is on average as accurate as HC2 but orders of magnitude faster. If results are required very quickly or train set sizes are large, MR-Hydra would seem to be the better option. However, for smaller train set sizes (see Table~\ref{tab:train}), or if probabilities or orderings are required (see Table~\ref{tab:sota}), the results indicate that HC2 is the better option. A special mention must be given to QUANT. It achieves high accuracy remarkably fast: it is an order of magnitude faster. We would recommend QUANT for very large problems, assuming it scales accordingly.

\begin{table}[htb]
    \centering
    \caption{Train time statistics on 142 TSC problems. H-InceptionTime is omitted because the times are not comparable. Best in bold.}
    \begin{tabular}{l|cccc}
    & Total (Hours) & Median (Minutes) & Min (Seconds) & Max (Hours) \\ \hline
    QUANT                 & \textbf{0.22}    & \textbf{0.02}   & 0.17  & \textbf{0.02}\\
    MR-Hydra (MR-H)       & 1.33    & 0.13   & 0.8   & 0.13\\
    WEASEL 2.0 (W 2.0)    & 1.92    & 0.17   & 1.8   & 0.17\\
    RDST                  & 4.63    & 0.95   & 16.51 & 0.23\\
    FreshPRINCE (FP)      & 30.02   & 2.95   & 12.27 & 4.08\\
    HC2                   & 263.89  & 15.28  & 38.94 & 65.66\\
    Proximity Forest (PF) & 743.42  & 8.7    & \textbf{0.09}  & 260.43\\
    \hline
    \end{tabular}
    \label{tab:time}
\end{table}

\begin{figure}[!ht]
    \centering
    \includegraphics[width=1.0\linewidth]{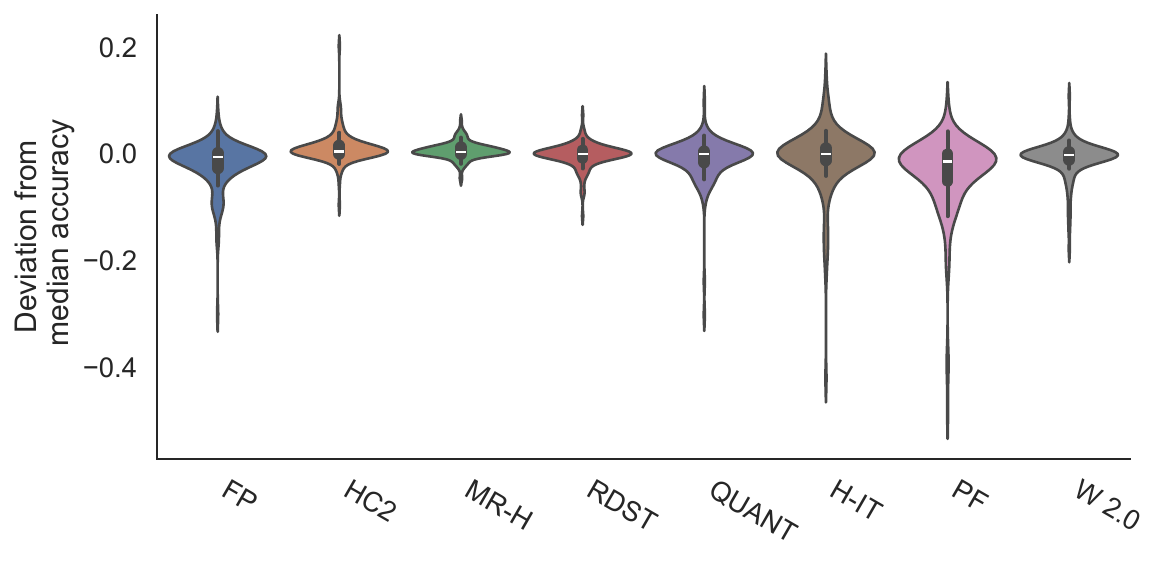}
    \caption{Distributions of the deviation of test accuracies from the median value of eight classifiers.}
    \label{fig:boxplot}
\end{figure}

 Figure~\ref{fig:boxplot} shows the violin plot of the deviation of each classifier from the median performing algorithm. It shows that both HC2 and MR-Hydra have tightly grouped distributions, with HC2 having a wider spread of positive values. H-InceptionTime (H-IT) has a very wide spread, both positive and negative, reflecting the wide variation in performance we have previously observed. FreshPrince (FP), QUANT and Proximity Forest (PF) also have wide distributions. PF performs relatively poorly more often than it does well.

We explored the effect on performance of the design decisions described in Table~\ref{tab:algorithm_characteristics}. If we group average accuracy ranks by each design factor of use of dilation, differences, ensemble, frequency domain and discretisation and perform a one factor ANOVA on each factor, we find a significant difference in rank distribution between those using dilation and those that do not, and those that use differences and those that do not. There was no significant difference in distribution when grouped by frequency, ensemble or discretisation. Care must be taken when interpreting these results since the assumptions behind the tests are not satisfied. However, there is at least some support for the utility of using dilation and differenced series.
 Finally, we have included a comprehensive correlation matrix on average accuracy ranks (Figure~\ref{fig:correlation_full} in the appendix). These demonstrate the diversity in performance of these classifiers and show the difference in performance between shapelet based and convolution based algorithms (Figure~\ref{fig:conv_shapelet_comparison}).

\section{Conclusions}
\label{sec:conc}

Research into algorithms for TSC has seen genuine progress in the last ten years, and the volume of research has dramatically increased. We have provided a particular view of this research landscape by grouping algorithms into eight categories defined by the core representation/transformation. We have compared the best in each category on 112 TSC problem and introduced 30 new datasets to counter any possible bias from over fitting. We evidence progress by benchmarking against algorithms previously considered the best performing, and show that two classifiers, MR-Hydra and HC2, generally perform the best.
HC2 performs significantly better on the current UCR archive, but there is less observable difference when we compare them on 30 new problems we have introduced. This could be due to the smaller sample size, the nature of the data sets or reflect some embedded bias in algorithm design. We note that HC2 does worse than MR-Hydra on imbalanced data and with larger train set sizes, but is better with more class balance, smaller train set sizes and with long series.

We are not claiming that these results should be taken to mean practitioners should always use MR-Hydra and/or HC2. There are strengths and weakness to all the algorithms we have described. Indeed, there is a case to be made for using QUANT by default, at least for exploratory analysis, because it is so fast. Understanding when it is appropriate to use which algorithm for a specific problem is an active research area. However, we suggest that, in the absence of any prior information these two algorithms make a sensible starting point for a new TSC problem. Despite significant research effort, there has not been an Alexnet for TSC, i.e. a deep learning approach that has dominated all others. It may be because the problems in the archives are relatively small compared to other archives used for deep learning evaluation: Table~\ref{tab:train} shows that H-Inception time improves relative to other algorithms as the number of training cases increases.

However, we think the core reason deep learning has not provided the gains many expected is that, unlike specific applications such as image classification or natural language processing, there is not one common underlying structure for the neural networks to exploit. Nevertheless, there is no doubt scope for improvements in deep learning algorithms for TSC. H-InceptionTime performs well overall, but Figure~\ref{fig:boxplot} demonstrates its limitations: it often performs terribly, and this makes its overall performance worse. If this tendency could be corrected, possibly by some automated structural optimisation, it seems likely that H-InceptionTime could match HC2 and MR-Hydra.

Since the original bake off a number of trends have developed in research, and some prior observations remain true. On average, hybrid algorithms still perform better than single domain approaches on the UCR archive. The ROCKET and HIVE-COTE family of classifiers work well because they combine convolution/shapelet approaches with dictionary based ones, i.e. they look for the presence of or the frequency of subseries. A key component of ROCKET based classifiers is dilation. We have shown that using dilation has significantly improved the single representation classifiers RDST and WEASEL 2.0. Incorporation of dilation could well benefit other algorithms, such as interval based classifiers. Ensemble algorithms are still effective and popular, but pipeline algorithms combining a transformation with a linear classifier such as ridge regression have shown to be just as competitive. The algorithms using linear classifiers such as ROCKET, WEASEL 2.0 and RDST have shown to be more scalable than ensembles generally, but cannot produce good probability estimates. More algorithms now incorporate transformed series such as first-order differences and periodograms into their feature extraction. This has been shown to increase accuracy in the majority of the algorithm types we have presented.

We believe there is great scope for improving time series specific classifiers: None except QUANT scale particularly well for large data, particularly in terms of memory: we are constructing a set of larger problems but none of the classifiers could be built in them in reasonable time and/or memory; there is a lack of principled work flows for using these classifiers to help understand the mechanisms for forming classifiers; multivariate TSC is less understood and many of the classifiers described have not been designed to be used in this way; and there has been little research into how best to handle unequal length series. We believe there are many unanswered questions in the field of TSC and predict it will remain as active and  productive for the next 10 years.

\backmatter

\section*{Acknowledgements}

This work is supported by the UK Engineering and Physical Sciences Research Council (EPSRC) grant number EP/W030756/1. The experiments were carried out on the High Performance Computing Cluster supported by the Research and Specialist Computing Support service at the University of East Anglia and the IRIDIS High Performance Computing Facility at the University of Southampton. We would like to thank all those responsible for helping maintain the time series classification archives and those contributing to open source implementations of the algorithms.

\section*{Statements and Declarations}

\noindent \textbf{Funding:} this work is supported by the UK Engineering and Physical Sciences Research Council (EPSRC) grant number EP/W030756/1.

\noindent \textbf{Conflicts of interest/Competing interests:} there are no conflicts of interest/competing interests.

\noindent \textbf{Ethics approval:} not applicable.

\noindent \textbf{Consent to participate:} all data used is freely available for scientific use.

\noindent \textbf{Consent for publication} The authors of this paper give our consent for the publication of identifiable details, which can include photograph(s) and/or videos and/or case history and/or details within the text

\noindent \textbf{Availability of data and material:} all data is available from \url{https://timeseriesclassification.com}.

\noindent \textbf{Code availability:} all code to reproduce experiments using open source software are available from the associated website \url{https://tsml-eval.readthedocs.io/en/latest/publications/2023/tsc_bakeoff/tsc_bakeoff_2023.html}.

\noindent \textbf{Authors' contributions:} Matthew Middlehurst, Patrick Sch\"afer and Anthony Bagnall were all involved in the implementation of algorithms used and the writing of the paper. Experiments were carried out on the High Performance Computing Cluster supported by the Research and Specialist Computing Support service at the University of East Anglia by Matthew Middlehurst and Anthony Bagnall.

\bibliography{TSCMaster,machineLearning,TSRegression,TSClustering, new_references}
\begin{appendices}

\section{Reproducibility}
\label{app:code}

The majority of the algorithms we have evaluated in this work are available in the aeon toolkit (see Footnote 2), and those that are not we plan to include over time. aeon is a Python based toolbox for time series analysis which contains a developed classification module. The aeon toolkit is compatible with scikit-learn\footnote{\url{https://scikit-learn.org/stable/}}, and aims to follow its interface and enable usage  of its tools where possible.

The version of aeon classifiers used in this work is aeon v0.6.0, with this version and later releases available on PyPi\footnote{\url{https://pypi.org/project/aeon/}}. In Listing~\ref{aeon_example} we show a usage example for a time series classifier in aeon. While we use ROCKET in the example as it is fast, the same interface applied to other aeon classifiers\footnote{\url{https://www.aeon-toolkit.org/en/latest/api_reference.html}}.

\lstset{style=python_jay}

\begin{lstlisting}[language=Python, caption={Loading data, building an estimator and making predictions using a time series classifier in aeon.}, label=aeon_example]
import numpy as np

from aeon.classification.convolution_based import RocketClassifier
from aeon.datasets import load_from_tsfile

if __name__ == "__main__":
    # 1a. Load data
    X_train, y_train = load_from_tsfile("data_TRAIN.ts")
    X_test, y_test = load_from_tsfile("data_TEST.ts")

    # 1b. Alternatively, format your data as a 3D numpy array and labels as a 1D array
    # shape == (n_instances, n_channels, series_length)
    X_train = 2 * np.random.uniform(size=(100, 1, 100))
    y_train = X_train[:, 0, 0].astype(int)
    X_test = 2 * np.random.uniform(size=(50, 1, 100))
    y_test = X_test[:, 0, 0].astype(int)

    # 2. Call fit() to build the classifier
    clf = RocketClassifier()
    clf.fit(X_train, y_train)

    # 3a. To predict the class label for new cases, use predict()
    predictions = clf.predict(X_test)

    # 3b. If probabilities are required, use predict_proba()
    probabilities = clf.predict_proba(X_test)

    # 3c. To just calculate accuracy, use score()
    accuracy = clf.score(X_test, y_test)
\end{lstlisting}

To run our experiments, we primarily use the tsml-eval\footnote{\url{https://github.com/time-series-machine-learning/tsml-eval}} package to produce results files for aeon and scikit-learn estimators. The version used for our experiments is tsml-eval v0.2.1, which contains additional estimators not currently available in aeon. Listing~\ref{tsmleval_example} gives an example for running an experiment using an aeon classifier loading from .ts files. Alternatively, you can input already loaded data as shown in Listing~\ref{tsmleval_example2}. For more guidance on producing our results using tsml-eval, visit our accompanying webpage hosted on the repository (see Footnote 9).

\begin{lstlisting}[language=Python, caption={Running a classification experiment using tsml-eval with data loaded from file.}, label=tsmleval_example]
from aeon.classification.convolution_based import RocketClassifier

from tsml_eval.experiments import load_and_run_classification_experiment
from tsml_eval.experiments import get_classifier_by_name

if __name__ == "__main__":
    # The directory where the data is stored. This directory should contain a directory with the name *dataset*, holding a *dataset*.TS file or *dataset*_TRAIN.TS and *dataset*_TEST.TS files
    data_dir = "../"
    # The directory to write the results file to
    results_dir = "../"
    # The name of the dataset to load
    dataset = "ItalyPowerDemand"
    # The resample id to use for random resampling, 0 uses the original train/test split if available
    resample_id = 0
    # The name of the classifier to use in the experiment
    classifier_name = "ROCKET"

    # The classifier to use
    classifier = RocketClassifier(random_state=resample_id)
    # Alternatively, use the set_classifier function to select a predefined classifier
    classifier = get_classifier_by_name(classifier_name)

    # Run the experiment
    load_and_run_classification_experiment(
        data_dir,
        results_dir,
        dataset,
        classifier,
        resample_id=resample_id,
        classifier_name=classifier_name,
    )
\end{lstlisting}

\begin{lstlisting}[language=Python, caption={Running a classification experiment using tsml-eval with pre-loaded data.}, label=tsmleval_example2]
import numpy as np
from aeon.classification.convolution_based import RocketClassifier

from tsml_eval.experiments import run_classification_experiment

if __name__ == "__main__":
    # shape == (n_cases, n_channels, n_timepoints)
    X_train = 2 * np.random.uniform(size=(100, 1, 100))
    y_train = X_train[:, 0, 0].astype(int)
    X_test = 2 * np.random.uniform(size=(50, 1, 100))
    y_test = X_test[:, 0, 0].astype(int)

    results_dir = "../"
    dataset = "ItalyPowerDemand"
    resample_id = 0
    classifier_name = "ROCKET"
    classifier = RocketClassifier(random_state=resample_id)

    run_classification_experiment(
        X_train,
        y_train,
        X_test,
        y_test,
        classifier,
        results_dir,
        classifier_name=classifier_name,
        dataset_name=dataset,
        resample_id=resample_id,
    )
\end{lstlisting}

The experiment functions will output a results file. The format for this file can be found in the \texttt{results\_format.ipynb} example Jupyter notebook\footnote{\url{https://github.com/time-series-machine-learning/tsml-eval/tree/main/examples}}. The utilities used to evaluate algorithms over multiple datasets and resamples can be found in the \texttt{evaluation.ipynb} notebook.

All results files from our experiments and a table of parameters can be found on the accompanying website and the publication directory\footnote{\url{https://github.com/time-series-machine-learning/tsml-eval/tree/main/tsml_eval/publications/y2023/tsc_bakeoff}}.

\section{Algorithms}
\label{app:algorithms}

Table~\ref{tab:pre-algorithms} shows the algorithms used in the original classification bake off~\cite{bagnall17bakeoff}. Table~\ref{tab:algorithms} shows the algorithms used in our experiments. Table~\ref{tab:algorithm_characteristics} shows the different characteristics for each algorithm.

\begin{table}
    \caption{Algorithms used in the 2017 bake off.}
    \label{tab:pre-algorithms}
    \footnotesize
    \begin{tabular}{|l|l|} \hline
    \multicolumn{2}{|c|}{Standard classifiers}\\ \hline
    Logistic & Logisitic regression \\
    C45     & Decision Tree \\
    NB      & Naive Bayes \\
    BN      & Bayesian Network \\
    SVML    & Linear kernel Support Vector Machine \\
    SVMQ    & Quadratic kernel Support Vector Machine \\
    MLP     & Multilayer Perceptron \\
    RandF   & Random Forest \\
    RotF    & Rotation Forest \\ \hline
    \multicolumn{2}{|c|}{Distance based}\\ \hline
    ED      & Euclidean Distance \\
    DTW     & Dynamic Time Warping \\
    WDTW    & Weighted DTW~\citep{jeong11weighted} \\
    TWE     & Time Warp Edit~\citep{marteau09stiffness} \\
    MSM     & Move-Split-Merge~\citep{stefan13msm}\\
    CID$_DTW$ & Complexity Invariant Distance with DTW~\citep{batista14cid} \\
    DD$_DTW$& Derivative DTW~\citep{gorecki13derivative} \\
    DTD$_C$ & Derivative Transform Distance~\citep{gorecki14nonisometric} \\
    EE      & Elastic Ensemble~\citep{lines15elastic} \\ \hline

    \multicolumn{2}{|c|}{Interval Based} \\ \hline
    TSF     & Time Series Forest~\citep{deng13forest} \\
    TSBF    & Time Series Bag of Features~\citep{baydogan13tsbf} \\
    LPS     & Learned Pattern Similarity~\citep{baydogan16lps} \\ \hline
    \multicolumn{2}{|c|}{Shapelet Based} \\ \hline
    FS   & Fast Shapelets~\citep{rakthanmanon13fastshapelets} \\
    ST   & Shapelet Transform~\citep{hills14shapelet} \\
    LS   & Learned Shapelets~\citep{grabocka14learning-shapelets} \\  \hline
    \multicolumn{2}{|c|}{Dictionary Based} \\ \hline
    BoP     & Bag of Patterns~\citep{lin12bagofpatterns} \\
    SAXVSM  & Symbolic Aggregate Approximation-vector \\
    & Space Model~\citep{senin13sax_vsm} \\
    BOSS    & Bag of Symbolic Fourier Approximation Symbols~\citep{schaefer15boss} \\ \hline
    \multicolumn{2}{|c|}{Hybrid} \\ \hline
    COTE/flat-COTE    & Collective of Transformation-based \\
    & Ensembles~\citep{bagnall15cote} \\
    DTW$_F$ & DTW Features~\citep{kate16features} \\ \hline

     \end{tabular}
\end{table}

\begin{table}
    \caption{Algorithms used in the Redux bake off.}
    \label{tab:algorithms}
    \footnotesize
    \begin{tabular}{|l|l|} \hline
    \multicolumn{2}{|c|}{Distance based}\\ \hline
    DTW     & Dynamic Time Warping \\
    ShapeDTW & Shape Based DTW~\citep{zhao18shape} \\
    EE  & Fast Elastic Ensemble~\citep{oastler19significantly} \\
    PF      & Proximity Forest~\citep{lucas19proximity} \\
    GRAIL   & Generic RepresentAtIon Learning~\citep{paparrizos19grail} \\ \hline
    \multicolumn{2}{|c|}{Feature Based} \\ \hline
    Catch22  &  Canonical Time Series Characteristics~\citep{lubba19catch22} \\
    Signatures &  Canonical Signature Pipeline~\cite{morrill20generalised} \\
    TSFresh & Time Series Feature Extraction Based on Scalable \\
    & Hypothesis Tests~\citep{christ18time} \\
    FreshPRINCE  &  Fresh Pipeline with Rotation Forest  \\
    & Classifier~\citep{middlehurst22freshprince} \\  \hline
    \multicolumn{2}{|c|}{Shapelet Based} \\ \hline
    RSF  & Random Shapelet Forest~\citep{karlsson16generalized} \\
    STC   & Binary Shapelet Transform Classifier~\citep{bostrom17binary} \\
    MrSQM  &  Multiple Representations Sequence Miner~\citep{nguyen22mrsqm} \\
    RDST  &  Random Dilated Shapelet Transform~\citep{guillaume22rdst} \\  \hline
    \multicolumn{2}{|c|}{Interval Based} \\ \hline
    TSF &  Time Series Forest~\citep{deng13forest} \\
    RISE & Random Interval Spectral Ensemble~\cite{flynn19contract} \\
    CIF & Canonical Interval Forest~\citep{middlehurst20canonical} \\
    DrCIF &  Diverse Representation CIF~\citep{middlehurst21hc2} \\
    STSF &  Supervised TSF~\citep{cabello20fast} \\
    R-STSF &  Randomised STSF~\citep{cabello21fast} \\
    QUANT & QUANTiles~\cite{dempster23quant} \\ \hline
    \multicolumn{2}{|c|}{Dictionary Based} \\ \hline
    BOSS  & Bag of Symbolic Fourier Approximation Symbols~\citep{schaefer15boss} \\
    cBOSS  & Contractable BOSS~\citep{middlehurst19scalable} \\
    WEASEL v1.0 &  Word Extraction for Time Series Classification~\citep{schaefer17weasel} \\
    TDE   &  Temporal Dictionary Ensemble~\citep{middlehurst20temporal} \\
    WEASEL v2.0  & WEASEL with Dilation~\citep{schaefer23weasle2} \\ \hline
    \multicolumn{2}{|c|}{Kernel/convolution Based} \\ \hline
    ROCKET & Random Convolutional Kernel Transform~\citep{dempster20rocket} \\
    Arsenal & The Arsenal~\citep{middlehurst21hc2} \\
    MultiROCKET & MultiROCKET~\citep{tan22multirocket} \\
    MiniROCKET & MiniROCKET~\citep{dempster21minirocket} \\
    Hydra   & Hybrid Dictionary–ROCKET Architecture~\citep{dempster22hydra}  \\
    MultiROCKET-Hydra & MultiROCKET + Hydra~\citep{dempster22hydra}\\ \hline
    \multicolumn{2}{|c|}{Deep Learning Based} \\ \hline
    CNN &  Convolution Neural Network~\citep{fukushima80neocognitron} \\
    ResNet &  Residual Network~\citep{wang17fcn} \\
    InceptionTime & Inception Time~\citep{fawaz20inception} \\   H-InceptionTime & Hybrid Inception Time~\citep{ismail22hybrid} \\ LiteTime    & Lite Inception Time~\citep{ismail23lite} \\ \hline
    \multicolumn{2}{|c|}{Hybrid} \\ \hline
    TS-CHIEF &  Time Series Combination of Heterogeneous and Integrated \\
    & Embedding Forest~\citep{shifaz20ts-chief} \\
    HC1    & Hierarchical Vote Collective of Transformation-based \\
    & Ensembles~\citep{bagnall20hivecote1} \\
    HC2    & HIVE-COTE version 2~\citep{middlehurst21hc2} \\
    RIST   & Randomised Interval-Shapelet Transformation ~\citep{middlehurst23rist} \\

    \hline
    \end{tabular}
\end{table}

\begin{table}
    \small
    \caption{Characteristics of the 40 approaches included in our experiments. Columns are dilation (dil), discretisation (disc), differences/derivatives (diff), frequency domain (freq), ensemble (ens) and linear classifier (lin).}
    \label{tab:algorithm_characteristics}
    \begin{centering}
    \begin{tabular}{|p{2.6cm}|>{\centering}p{1.1cm}|>{\centering}p{1.1cm}|>{\centering}p{1.1cm}|>{\centering}p{1.1cm}|>{\centering}p{1.1cm}|>{\centering}p{1.1cm}|}
    \hline
     & \textbf{dil} & \textbf{disc} & \textbf{diff} & \textbf{freq} & \textbf{ens} & \textbf{lin}\tabularnewline
    \hline
    \hline
    DTW     & & & & & &\tabularnewline
    \hline
    ShapeDTW & & & & & &\tabularnewline
    \hline
    EE  & & & X & & X &\tabularnewline
    \hline
    PF      & & & X & & X &\tabularnewline
    \hline
    GRAIL   & & & & & &\tabularnewline
    \hline
    Catch22  & & & & & &\tabularnewline
    \hline
    Signatures & & & & & &\tabularnewline
    \hline
    TSFresh & & & & & &\tabularnewline
    \hline
    FreshPRINCE  & & & & & &\tabularnewline
    \hline
    RSF  & & & & & X &\tabularnewline
    \hline
    STC   & & & & & &\tabularnewline
    \hline
    MrSQM  & & X & & X & & X \tabularnewline
    \hline
    RDST  & X & & & & & X \tabularnewline
    \hline
    TSF & & & & & X &\tabularnewline
    \hline
    RISE & & & & X & X &\tabularnewline
    \hline
    CIF & & & & & X &\tabularnewline
    \hline
    DrCIF & & & X & X & X &\tabularnewline
    \hline
    STSF & & & X & X & X &\tabularnewline
    \hline
    R-STSF & & & X & X & &\tabularnewline
    \hline
    QUANT & & & X & X & &\tabularnewline
    \hline
    BOSS  & & X & & X & X &\tabularnewline
    \hline
    cBOSS  & & X & & X & X &\tabularnewline
    \hline
    WEASEL v1.0 & & X & & X & & X \tabularnewline
    \hline
    TDE   & & X & & X & X &\tabularnewline
    \hline
    WEASEL v2.0  & X & X & X & X & & X \tabularnewline
    \hline
    ROCKET & X & & & & & X \tabularnewline
    \hline
    Arsenal & X & & & & X & X \tabularnewline
    \hline
    MultiROCKET & X & & X & & & X \tabularnewline
    \hline
    MiniROCKET & X & & & & & X \tabularnewline
    \hline
    Hydra   & X & & X & & & X \tabularnewline
    \hline
    MR-Hydra & X &  & X & & & X \tabularnewline
    \hline
    CNN & & & & & &\tabularnewline
    \hline
    ResNet & & & & & &\tabularnewline
    \hline
    InceptionTime & & & & & X &\tabularnewline
    \hline
    H-InceptionTime & & & & & X &\tabularnewline
    \hline
    LiteTime    & & & & & X &\tabularnewline
    \hline
    TS-CHIEF & & & X & X & X &\tabularnewline
    \hline
    HC1    & & X & & X & X & \tabularnewline
    \hline
    HC2   & X & X & X & X & X & X \tabularnewline
    \hline
    RIST   & X & & X & X & &\tabularnewline
    \hline
    \end{tabular}
    \par\end{centering}
\end{table}

\section{Results}
\label{app:results}

\begin{figure}[t]
    \centering
    \includegraphics[width=1.3\linewidth,angle=90,trim={4cm 0cm 3cm 0cm},clip]{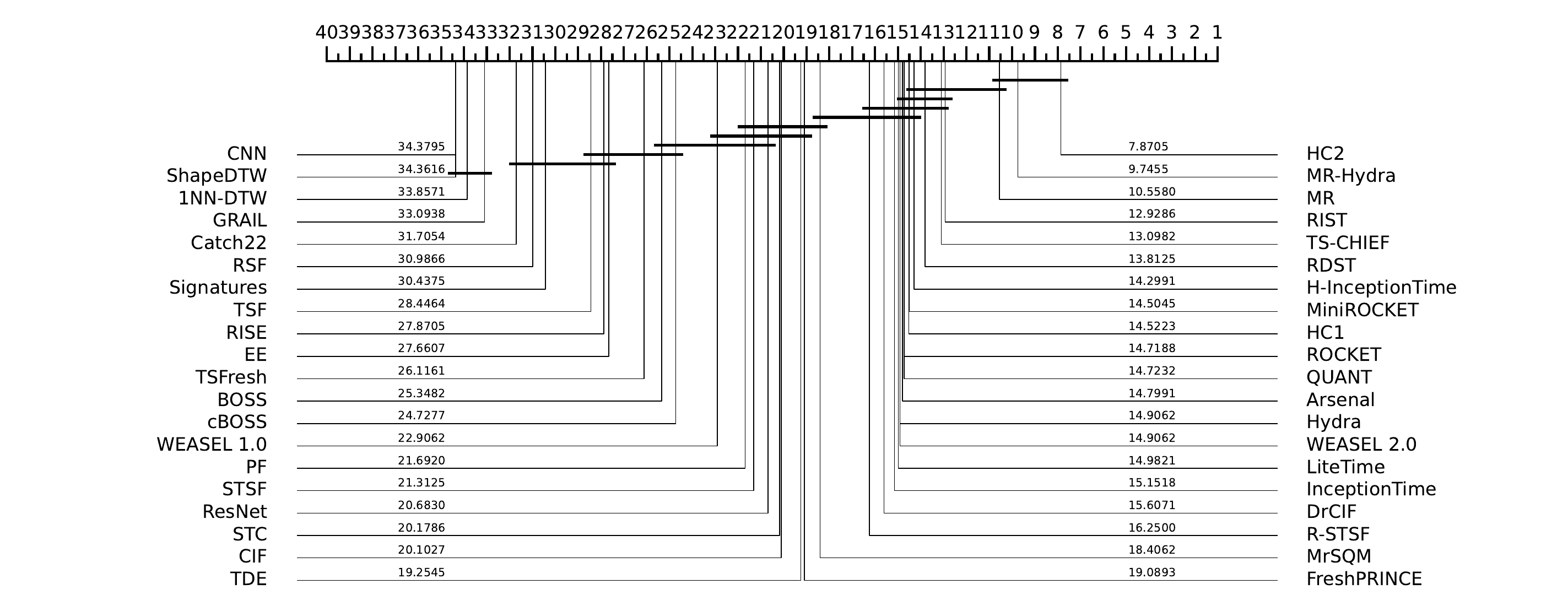}
    \caption{Ranked test accuracy of all 40 classifiers in the bake-off on 112 UCR UTSC problems. Accuracies are averaged over $30$ resamples of train and test splits.}
    \label{fig:all-cd}
\end{figure}

\begin{figure}[t]
	\centering
    \includegraphics[height=0.60\linewidth, angle=90, trim={0cm 0cm 0cm 0cm},clip]{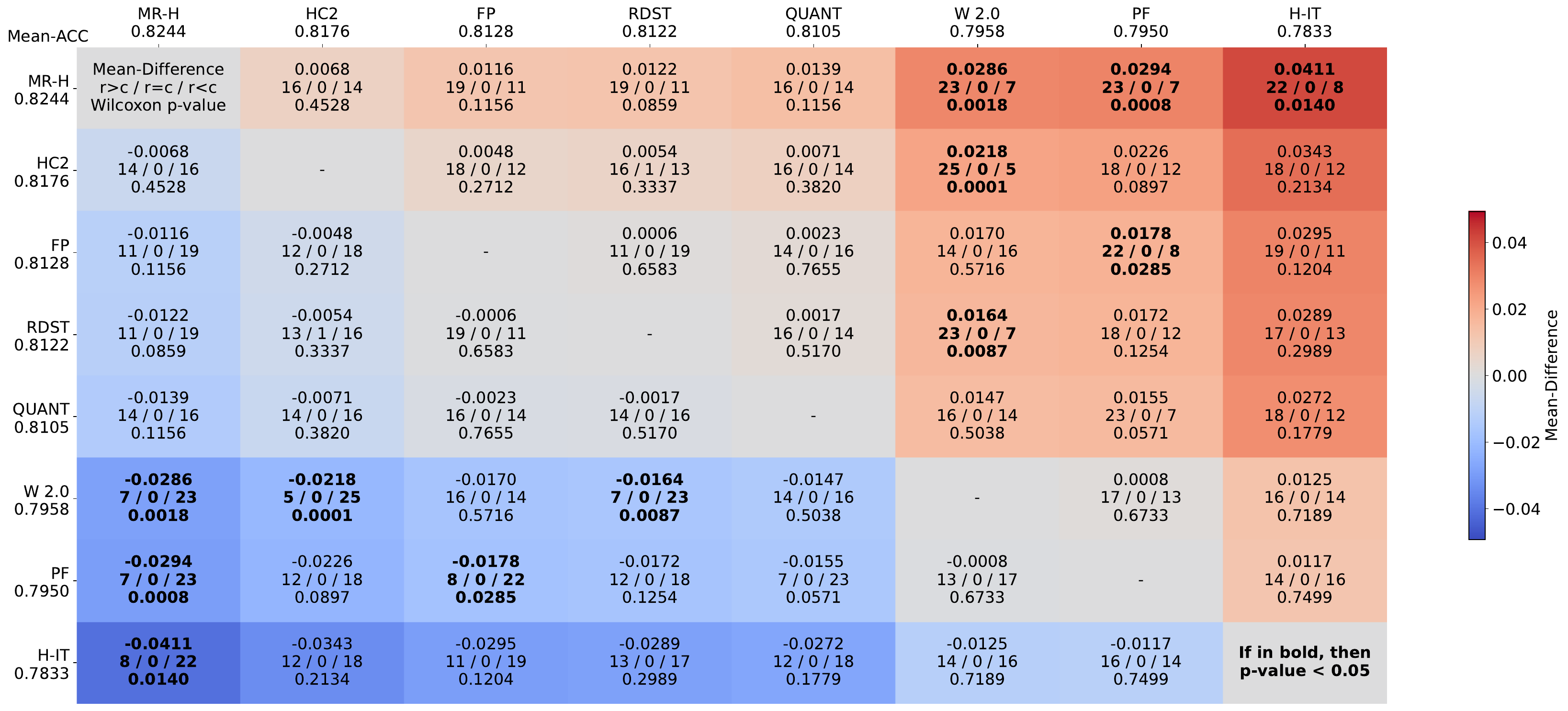}
    \caption{Summary performance statistics for eight classifiers on 30 datasets, generated using the multiple comparison matrix (MCM).}
    \label{fig:heatmap30}
\end{figure}

\begin{figure}[t]
	\centering
    \includegraphics[height=0.60\linewidth, angle=90, trim={0cm 0cm 0cm 0cm},clip]{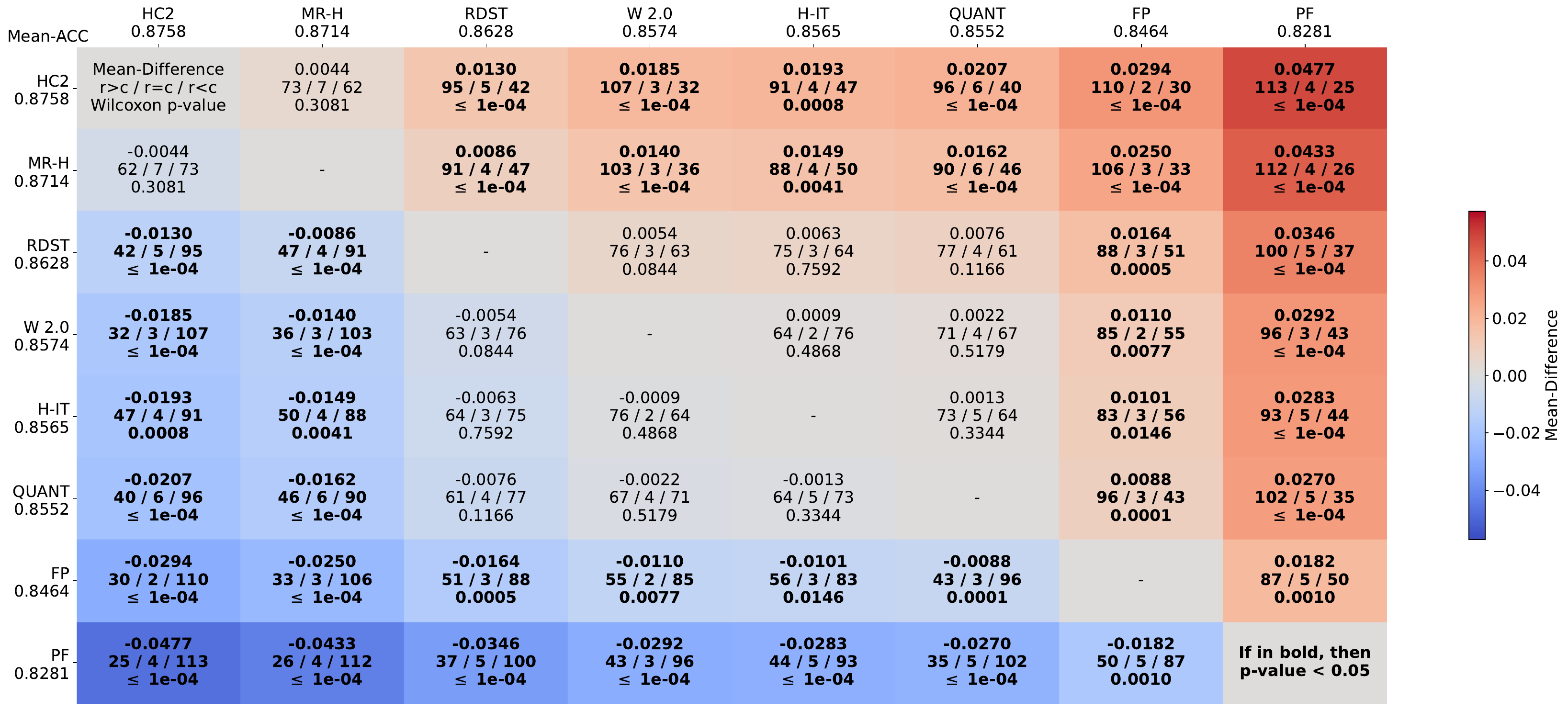}
    \caption{Summary performance statistics for eight classifiers on 142 datasets, generated using the multiple comparison matrix (MCM).}
    \label{fig:heatmap142}
\end{figure}

\begin{figure}[t]
	\centering
    \includegraphics[width=0.9\linewidth]{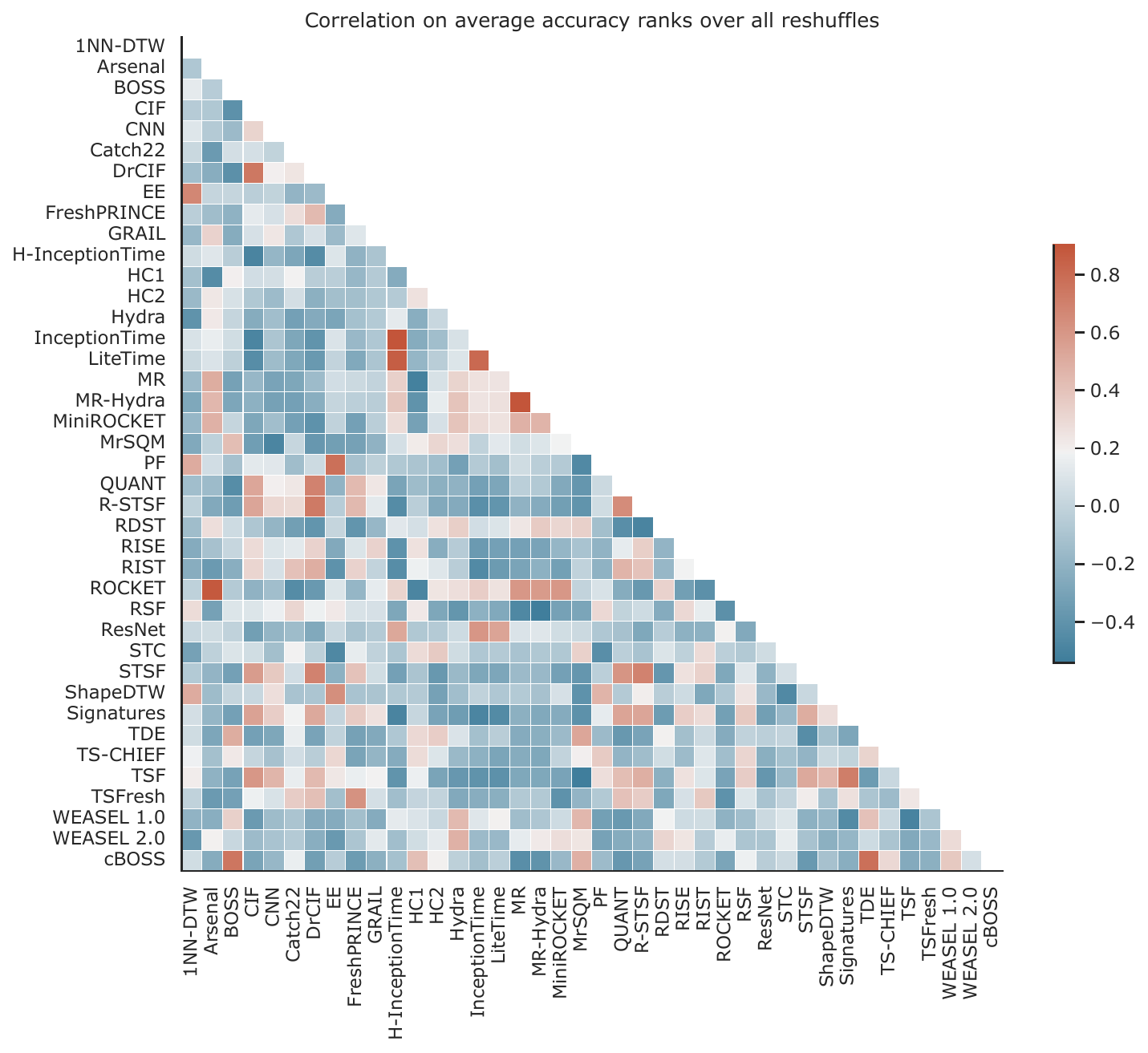}
    \caption{Correlation on Average Accuracy Ranks for all competitors.}
    \label{fig:correlation_full}
\end{figure}

\begin{figure}[t]
	\centering
    \includegraphics[width=0.6\linewidth]{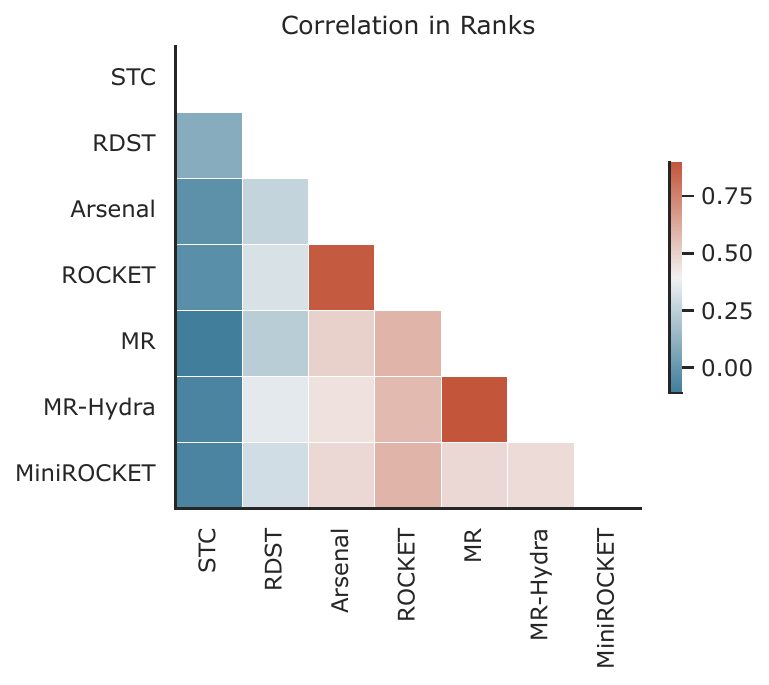}
    \caption{Correlation on Average Accuracy Ranks between Shapelet-based and Convolution-based approaches.}
    \label{fig:conv_shapelet_comparison}
\end{figure}

Table~\ref{tab:results30} shows the accuracy results of the best performing algorithms on the 30 new classification datasets. Figure~\ref{fig:all-cd} shows a critical difference diagram for all algorithms included in our experiments on the 112 UCR datasets. Figures~\ref{fig:heatmap30} and ~\ref{fig:heatmap142} contain additional MCM diagrams for the 30 new datasets and original 142 datasets respectively on the best in class algorithms. Figures~\ref{fig:correlation_full} and ~\ref{fig:conv_shapelet_comparison} show correlation diagrams for the algorithms.

\begin{table}[!ht]
    \caption{Accuracy of classifiers on the 30 new datasets (averaged over 30 resamples).}
    \label{tab:results30}
    \tiny
    \centering
    \begin{tabular}{|l|l|l|l|l|l|l|l|l|l|}
    \hline
         & \rot[60]{MR-Hydra} & \rot[60]{HC2} & \rot[60]{RDST} & \rot[60]{QUANT} & \rot[60]{FreshPRINCE} & \rot[60]{H-IT} & \rot[60]{WEASEL v2.0} & \rot[60]{PF} & \rot[60]{1NN-DTW} \\ \hline
        AconityMINIPrinterLarge\_eq   & 95.70\% & 95.46\%	 & 95.82\% & 94.39\% & 94.61\% & 96.49\% & 95.34\% & 91.04\% & 85.85\% \\ \hline
        AconityMINIPrinterSmall\_eq   & 97.75\% & 97.85\%	 & 97.56\% & 97.58\% & 97.19\% & 97.74\% & 97.63\% & 96.40\% & 95.00\% \\ \hline
        AllGestureWiimoteX\_eq       & 76.53\% & 73.79\%	 & 71.03\% & 67.63\% & 68.82\% & 81.44\% & 70.08\% & 76.68\% & 68.67\% \\ \hline
        AllGestureWiimoteY\_eq        & 80.63\% & 77.60\%	 & 73.28\% & 70.51\% & 71.43\% & 83.83\% & 73.16\% & 75.34\% & 67.65\% \\ \hline
        AllGestureWiimoteZ\_eq        & 73.07\% & 72.89\%	 & 67.76\% & 66.18\% & 67.49\% & 78.24\% & 69.17\% & 74.19\% & 68.22\% \\ \hline
        AsphaltObstaclesUni\_eq       & 88.52\% & 88.90\%	 & 88.00\% & 84.28\% & 85.72\% & 91.19\% & 88.37\% & 88.87\% & 80.49\% \\ \hline
        AsphaltPavementTypeUni\_eq    & 91.76\% & 89.08\%	 & 89.24\% & 93.59\% & 92.93\% & 93.84\% & 79.96\% & 89.26\% & 59.97\% \\ \hline
        AsphaltRegularityUni\_eq      & 98.69\% & 95.81\%	 & 97.69\% & 98.87\% & 98.66\% & 98.56\% & 93.27\% & 98.30\% & 69.24\% \\ \hline
        Colposcopy                  & 37.59\% & 39.74\%	 & 39.60\% & 42.41\% & 41.19\% & 36.11\% & 38.28\% & 32.90\% & 28.65\% \\ \hline
        Covid3Month\_disc          & 60.87\% & 65.52\%	 & 65.30\% & 61.48\% & 61.91\% & 46.94\% & 60.98\% & 60.33\% & 55.85\% \\ \hline
        DodgerLoopDay\_nmv            & 54.81\% & 60.78\%	 & 61.08\% & 61.17\% & 57.45\% & 52.64\% & 60.39\% & 58.83\% & 41.95\% \\ \hline
        DodgerLoopGame\_nmv           & 86.54\% & 84.49\%	 & 80.84\% & 83.31\% & 84.88\% & 78.24\% & 84.25\% & 88.06\% & 86.77\% \\ \hline
        DodgerLoopWeekend\_nmv        & 98.10\% & 98.36\%	 & 98.36\% & 98.49\% & 97.88\% & 97.01\% & 97.94\% & 98.47\% & 95.53\% \\ \hline
        ElectricDeviceDetection     & 90.18\% & 89.15\%	 & 90.02\% & 90.20\% & 89.68\% & 75.14\% & 89.31\% & 88.99\% & 85.84\% \\ \hline
        FloodModeling1\_disc       & 93.53\% & 87.23\%	 & 86.95\% & 94.49\% & 94.95\% & 92.06\% & 79.92\% & 93.32\% & 88.48\% \\ \hline
        FloodModeling2\_disc       & 97.65\% & 93.42\%	 & 93.23\% & 97.81\% & 97.20\% & 96.50\% & 93.17\% & 96.57\% & 96.05\% \\ \hline
        FloodModeling3\_disc      & 90.94\% & 80.02\%	 & 82.08\% & 92.70\% & 94.18\% & 88.51\% & 71.72\% & 90.60\% & 86.39\% \\ \hline
        GestureMidAirD1\_eq           & 77.36\% & 78.03\%	 & 74.26\% & 71.08\% & 70.67\% & 75.36\% & 73.33\% & 64.46\% & 45.62\% \\ \hline
        GestureMidAirD2\_eq           & 63.67\% & 64.82\%	 & 62.21\% & 61.77\% & 61.36\% & 59.95\% & 63.33\% & 53.95\% & 31.79\% \\ \hline
        GestureMidAirD3\_eq           & 48.46\% & 53.18\%	 & 51.92\% & 47.03\% & 47.46\% & 39.62\% & 44.72\% & 34.54\% & 18.41\% \\ \hline
        GesturePebbleZ1\_eq           & 95.72\% & 95.95\%	 & 97.13\% & 94.52\% & 94.40\% & 97.09\% & 96.10\% & 89.96\% & 71.10\% \\ \hline
        GesturePebbleZ2\_eq           & 96.79\% & 96.75\%	 & 97.49\% & 94.98\% & 94.41\% & 96.12\% & 96.92\% & 90.78\% & 73.35\% \\ \hline
        KeplerLightCurves           & 92.46\% & 96.66\%	 & 92.84\% & 94.80\% & 96.57\% & 75.07\% & 92.39\% & 89.47\% & 80.14\% \\ \hline
        MelbournePedestrian\_nmv      & 96.31\% & 94.75\%	 & 95.95\% & 97.06\% & 96.44\% & 96.78\% & 92.09\% & 95.28\% & 88.55\% \\ \hline
        PhoneHeartbeatSound         & 65.07\% & 89.35\%	 & 66.87\% & 91.61\% & 65.26\% & 47.59\% & 64.08\% & 63.81\% & 54.27\% \\ \hline
        PickupGestureWiimoteZ\_eq     & 84.67\% & 64.58\%	 & 82.20\% & 66.23\% & 79.80\% & 63.66\% & 82.07\% & 80.93\% & 70.00\% \\ \hline
        PLAID\_eq                     & 93.92\% & 78.40\%	 & 91.78\% & 81.73\% & 88.87\% & 77.40\% & 89.85\% & 87.99\% & 84.97\% \\ \hline
        ShakeGestureWiimoteZ\_eq      & 92.87\% & 93.13\%	 & 93.73\% & 85.53\% & 90.47\% & 85.93\% & 92.73\% & 89.73\% & 85.47\% \\ \hline
        SharePriceIncrease          & 66.48\% & 68.62\%	 & 66.02\% & 69.01\% & 69.44\% & 65.26\% & 68.62\% & 69.13\% & 62.05\% \\ \hline
        Tools                       & 86.52\% & 88.43\%	 & 86.27\% & 81.00\% & 86.99\% & 85.52\% & 88.26\% & 76.74\% & 69.33\% \\ \hline
    \end{tabular}
\end{table}

\appendix{Reviews and Responses}

We include the reviews for future reference and to help give insights into the sometimes painful process of publishing. This work started in 2021 for a book chapter that has not yet appeared, experimentation was done in 2022 and the paper was finalised in 2023. It was finally published on line first on 19th April 2024.

\section{Machine Learning Journal}

We first submitted this work to Machine Learning Journal in April 2023. Reviewers were not assigned until April and the decision email, received on 8th December 2023, was reject with encouragement to resubmit. The reviewers, though helpful in many ways, did not seem to buy in to the point of the paper, which was submitted as a survey paper. Based on what the reviewers were asking and the time taken to review (eight months), we decided to withdraw the paper, update it with some more recent algorithms then submit elsewhere.

The action editors report was as follows:

The reports from the reviewers of your manuscript, "Bake off redux: a review and experimental evaluation of recent time series classification algorithms", which you submitted to Machine Learning, have now been received.

Based on the advice received, your manuscript requires major revisions in order to be acceptable for publication.
It is the policy of Machine Learning not to accept manuscripts requiring major revisions.
Instead, authors choosing to revise their manuscript according to the reviewers' comments must resubmit a revised manuscript, which usually will be reassessed by the same action editor.

I want to stress that the reviewers appreciate the survey and see a viable path forwards. An increased focus on the target audience (or audiences as it is potentially mixed between novices and experts) with clear take-away messages and covering a wide selection of techniques and metrics (a.o. scalability) makes this a valuable contribution.

\subsection{Reviewer 1}

The paper extends a prior large-scale study in the area of time-series classification from 2017. Since then, new datasets and new methods have appeared that alter the findings of the prior study. In particular, additional univariate and multivariate are now included, as well as several new methods, especially from the deep learning category. The study elaborates on the methods as well as on the new results.

Strengths:

- Extensive study in the area of time-series classification, which helps understand progress in the area
- More methods and datasets are now included

Weaknesses:

- Same subset of authors may create biases in which approaches are selected.
- Several solutions and new findings are ignored
- The "better" methods are essentially ensemble methods - not really surprising
- No scalability/runtime experiments

Comments:

- First, let me state that I TRULY appreciate the effort here. It's a very time-consuming process to design and experiment with so many methods and datasets! Huge applause for the effort.

- That being said, I hope you understand that novelty is limited. We have an overlap in methods, datasets, and even authors, with the previous study, which creates all types of problems. It's not clear what is the goal. The prior (subset) authors have created an entire website timeseriesclassification.com where the new methods and results can be reported. It's somewhat strange every 5 years to repeat the same study with a few more datasets and a few more methods, especially when the findings do not contradict earlier results or other studies in the area. I would consider this differently if a new set of authors would fully replicate/reproduce all previous experiments + added new methods/datasets. However, now, several of the earlier biases still exist, as I explain below.

- It's unclear how much more benefit we get vs. for example the study from Forestier's group in the same area back from 2019. Many newer DNN methods were reported in 2019, which are also included now (in part) and more or less the findings remain the same. A few more hybrid solutions are now on top, but that's very much expected. However, from truly new/novel methods (not ensemble), nothing changes if I parsed the results correctly vs previous study or Forestier's study.

- No runtime/scalability results are presented.

- For several of the older methods it's unclear if new experiments were performed or the prior results were used

- The experimental settings are identical. Again, this might have been an opportunity for an improvement/change to validate earlier findings under new/stronger settings

- Several datasets are being ignored from UCR because they contain missing data or unequal lengths. However, many other studies focus on 128 datasets by filling data/interpolation and resampling data.

- Several outcomes from the previous study, propagate to the new study, while findings from other groups are being ignored. For example, the study insists of using 1-NN-DTW as the baseline classifier when we have evidence that there are stronger distances available [a]. Similarly, sliding measures (equally good to DTW but faster) [a], embedding approaches [a], kernel measures plugged with SVM [b], representation learning methods from the non-deep learning literature [b], all are being ignored. Yet, variants of other methods that trade-off runtime performance for higher accuracy are being included. Similarly, there is a focus on hybrid/ensemble methods. combining dozens!! of methods. Yes, including even more methods in the ensembles will definitely improve performance.. but where will this reach? every time there is a new method, we will create a new ensemble including this new method?

[a] "Debunking four long-standing misconceptions of time-series distance measures." Proceedings of the 2020 ACM SIGMOD international conference on management of data. 2020.
[b] "Grail: efficient time-series representation learning." Proceedings of the VLDB Endowment 12.11 (2019): 1762-1777.

- The study separated convolution-based methods from DNN methods. It's unclear why this is happening.

\subsection{Reviewer 2}

The paper presents a comprehensive and structured comparison of recent techniques and advances in Time Series Classification (TSC) algorithms. Building on the work of Bagnall et al. from 2017, it aims to update and extend their findings. Specifically, the paper introduces new algorithms and taxonomies, and evaluates new datasets to determine how each category has improved since the original publication. Emphasising reproducibility and methodological transparency, the paper provides researchers with a detailed framework to benchmark current and emerging methods in TSC.

I had a generally positive impression of the methodology adopted in the paper, and I think it is useful, as the field advances rapidly, to keep track of the comparative performance of different algorithms. However, the target audience for the publication is unclear. Although it goes into great depth on subjects like CNNs and basic ideas (like giving a formal definition for time series), it occasionally skips over other terms that novices in the area would not be familiar with. This inconsistent approach to explaining technical jargon might cause confusion or repetition for experts who already have a solid understanding of the subject, in addition to posing difficulties for readers who are unfamiliar with these topics. The paper's effectiveness for both novice and expert audiences may be unintentionally hampered by the varying amount of explanation. I suggest being clearer from the beginning and addressing who the target audience is for the paper.

The paper's extensive length, coupled with its frequent use of acronyms, presents a readability challenge. I frequently found myself needing to refer back to previous sections to recall the meanings of acronyms introduced pages earlier. Although the appendix of the paper contains an acronym glossary and summary table (Table 22), its placement somewhat reduces its usefulness. It's possible that readers won't notice it right away or will find it annoying to have to continually go to the end of the document to find references. To improve readability and accessibility, it would be beneficial to clearly highlight the glossary's availability. This could be achieved by mentioning it in the introduction or situating it in a more convenient place within the paper. Additionally, the table could be expanded to not only help the reader navigate the acronyms but also to include brief descriptions or contextual notes regarding each acronym. This expansion would aid in providing a clearer understanding of the methods and concepts mentioned in the paper, thereby simplifying the reading experience.

To enhance the paper's readability, I recommend the inclusion of additional illustrations or diagrams. These visual aids could be particularly helpful in clarifying sections where the explanations may be less clear. Incorporating such images would not only aid in better understanding complex concepts but also provide a more engaging and accessible reading experience.

Given that the paper is titled as a 'review', it should ideally provide a clear and targeted overview of the state-of-the-art in time series classification. It should be structured to be easily comprehensible to readers with at least a basic background in time series analysis.

The paper would benefit from clearer, more concise English and shorter sentences, as well clearer structure in order to improve readability.

While the paper provides a solid analysis of algorithm performance across various datasets, an additional layer of depth could be achieved by exploring why certain algorithms perform better on some datasets than others. This further insight would be a valuable enhancement, enriching our understanding of the algorithms' nuanced behaviours.

Moving forward, I will provide specific comments on a page-by-page basis, accompanied by annotations for clarity.

\begin{itemize}
\item       The ordering of the series does not have to be in time: we could transform audio into the frequency domain or map one dimensional image outlines onto a one dimensional series” (p.
2)
    * The mention of transforming audio into the frequency domain or mapping image outlines might need further elaboration or a reference, as it could be complex for readers unfamiliar with these concepts.
\item       “bake off redux” (p. 3) o It is the first time that the term is used in the paper and should be clear to people familiar with the concepts, but a small definition could help someone who has just approached the paper without reading the original one.
\item       “tools we use” (p. 3)
    * It would be beneficial to expand this section to clearly delineate the tools used in your study. Providing a detailed description of the tools will help readers easily visualize the process and understand how to replicate it in their own work while reading.
\item       Definitions and Terminology (p. 3)
    * Maybe this section should have some definitions regarding the metrics you’ve used.
\item       Definition of MTS (p. 3)
 \item   * There are two typos in the formulas $A = (a_1, \ldots a_n, )$ and $k^{th}$ channel
       techniques” (p. 4)
    * Typo, should be technique.
\item       Section “Experimental Procedure” (p. 5)
    * Consider breaking down the section into subsections for ease of reading something like "Datasets", "Resampling Strategy", "Performance Measures", and "Statistical Analysis”. Also Adding a flowchart or a visual representation of the experiments could also be useful.
\item       “Because of this, we restrict our attention to univariate classification only” (p. 5)
    * This is not well justified. It might be useful to briefly touch upon why the focus is solely on univariate classification and not on multivariate time series classification.
\item       “3.1 New Datasets” (p. 6)
    * Consider introducing a brief rationale for the choice of each dataset, why is it relevant to the overarching goals of the paper?
\item       “We are happy to accept new datasets through the repository associated with the archive”
    * Probably too informal for a scientific paper, it should something like "Submissions of new datasets to the associated repository are welcomed"
\item       “sensitive” p. 14)
    * What defines "sensitive to mean and variance". Please better define the term sensitivity here
    * The notation is not clear and n,m,d, … should be explicitly defined
\item       “Experiments have shown that when presented with a query and a sample time series, disregarding words that exist solely in the sample time series using the non-symmetric distance function leads to improved performance compared to using the Euclidean distance metric” (p. 26)
    * This needs to be further explained and accompanied with references.
\item       “ROCKET-family.” (p. 27) o Never defined before in the text, a definition or reference should be given here.
\item       “different resolutions” (p. 27)
    * The term “resolution” should be better defined in the text. The image helps clarifying.
\item      “spits” (p. 27)  typo: splits
\item       “Randomly drawn from an exponential function” (p. 31)
    * Should be rephrased, something like "The dilation is randomly drawn from an exponential function" for clarity.
\item       “The combination of ROCKET with logistic (RIDGE) regression is conceptually the same as a single-layer CNN with randomly initialised kernels and softmax loss” (p. 32)
    * Significant claim and needs to be substantiated, perhaps with references or further elaboration.
\item       “Without naming and shaming” (p. 35)
    * Appears too colloquial for an academic paper.
\item       “Finally, they often do not seem to offer any advance on previous research” (p. 35)
    * This assertion appears to be a strong claim, particularly regarding the comparison with InceptionTime. I recommend a more detailed discussion or elaboration on this point to substantiate the claim. Specifically, it would be beneficial to include an analysis of how and why newer algorithms do not significantly surpass the performance of InceptionTime.
\item       In the “CNN” section (p. 35)
    * the term 'CNN' has been used multiple times prior to this section without a formal introduction or definition. For clarity and to aid readers who may not be familiar with the acronym, it would be beneficial to introduce and define 'CNN' at its first occurrence in the paper.
\item       In the “ResNet” section (p. 36)
    * The transition from the foundational aspects of CNNs to the more complex architecture of ResNet might be smooth for readers with a moderate level of expertise in deep learning. However, for those who are new or less familiar with these concepts, the jump in complexity could be somewhat challenging. I suggest that the authors consider adding a bridging paragraph at the beginning of the ResNet section.
\item       “HC1 has four modules: it drops the computationally intensive EE algorithm without loss of accuracy …” (p. 39)
    * The current sentence structure might be confusing for some readers. A clearer way to present this information could be through a bullet-point list or a structured format that separately highlights each of the four modules of HC1.
\item       “The TS-CHIEF is made up of an ensemble of trees which embed distance, dictionary and spectral base features.” (p. 40)
    * For clarity and correctness, would be better to say "The TS-CHIEF comprises an ensemble of trees that embed distance, dictionary, and spectral base features"
\item       “The picture for AUROC and NLL is more confused lower down the order” (p. 41)
    * Please define confusion in this context perhaps by specifying what aspects of the AUROC and NLL results contribute to this confusion.
\item       “and the algorithm could be better configured” (p. 46)

    * How would you better configure this algorithm? What are the parameters involved? It would be helpful if the authors could provide more specific details on what aspects of HC2's configuration contribute to its slower performance compared to Hydra-MR. Clarifying whether 'configuration' refers to algorithmic parameters, structural components of HC2, or other operational settings.
\item       “WE” (p. 50)
    * Typo: We
\end{itemize}
\section{Data Mining and Knowledge Discovery}

We withdrew from Machine Learning on 5/1/2024 and submitted to Data Mining and Knowledge discovery. This was a much easier process: we received reviews on 25/2/24, revised and resubmitted on 15/3/24 and had the final acceptance on 29/3/24. The action editor acted very efficiently. We had four reviewers for the first round of reviews.

\subsection{Round1}
Editor: Your manuscript, "Bake off redux: a review and experimental evaluation of recent time series classification algorithms", has now been assessed. We invite you to revise your paper. When your revision is ready, please submit the updated manuscript and a point-by-point response.

\subsubsection{Reviewer 1}
The paper is a revisit of the time series bake-off paper that overviews the current state-of-the-art in (univariate) time series classification (TSC). The new material includes algorithms that were published after the first bake-off, the introduction of the new 30 datasets, and the results of the new experiments. Code, data, and guide to reproduce the experiments are also provided.

Like its predecessor, this paper is a great effort by the authors to capture the current landscape of TSC research. I appreciate the enormous amount of work (data preparation, coding, experiments, etc.) that need to be done for this paper. I strongly believe that, like its predecessor, future research in this area will greatly benefit from this work.

Following are some of my suggestions. I hope it can be helpful to the authors:

“We contribute 30 new datasets”: If i understand correctly, some datasets are donated by the other researchers. Wording it this way can be misunderstood that the authors of the paper are also the authors of all 30 datasets.

“The SharePriceIncrease data was formatted as part of a student project”. Is it possible to acknowledge the student (e.g. with name/link) that led the project ?

Figure 9 shows an example of spectro data where interval-based approaches may be superior. On the other hand, table 18 presents a different story where Quant (the best interval-based method) has the worst rank on the spectro column. While I agree that intuitively interval-based methods work best with phase dependent data, I don’t think this is a good example.

I also would like to add some thoughts on the classifying of the TSC algorithms. I find the categories (shaplet-based, feature-based, interval-based, etc.) proposed by the authors nicely capture the main characteristics of the algorithms. However, I would say some algorithms can cross the line between the categories. For example MrSQM/MrSEQL are shapelet methods but they are also very similar to dictionary-based methods (BOSS, WEASEL etc.) in that they also discretise the time series. The only difference is that , instead of frequency (histogram), they use binary occurrence information. Some classifiers also make use of feature-based techniques after transforming the time series (e.g. Multi Rocket early version tried catch22 features if I remember correctly). The algorithms across the categories also share some mechanisms (dilation, first difference, fourier transform, multi domain, multi view, multi resolution etc.). In this regard, I would love to learn the opinions of the authors on what seem to be the most effective ideas/themes in the current state of research and what would be the potential directions going forward in the area.  A table similar to table 3 in the first bake-off would be nice. Also, I think the name “ensemble” is better at capturing the characteristics of the last group than “hybrid”.

Figure 38 vs Figure 40 vs Figure 41 show the interesting results on the 112, 30, and 142 datasets that I think require deeper discussion. It looks like the results from the 112 datasets generalise but not strongly (e.g MR-H looks a bit better and HC2 looks a bit worse in the 30 new datasets). Would be interesting to see more multiple comparison matrix as well (e.g. fig 42). They can be put in the appendix.

Overall, I find this paper like its predecessor a great contribution to  TSC research. With the recent update of the data archive and the advancement of the state-of-the-art I think this is a timely refresh.

\subsubsection{Reviewer 2}

Summary:

This paper addresses the task of time series classification by presenting a range of algorithms proposed in recent years. These algorithms are categorized from distance-based approaches to hybrid ensemble models. Extensive experiments compare the algorithms within each category and also among the category winners. The paper uses the UCR archive for many TSC applications and introduces new datasets to the field, including corrected versions of unused UCR datasets, adapted versions from regression datasets, and entirely new ones.

Strengths of the paper:

1. Despite the dense accumulation of information, the paper is very well written, making it simple and easy to follow.
2. An extensive number of experiments have been conducted with different resampling on the datasets to ensure a general conclusion about which algorithms are to be used in given applications for future TSC datasets.
3. A detailed statistical and efficiency comparison between models is proposed, adding depth to the analysis.
4. The organization of the paper is very good, aiding in the comprehension of complex material.
5. By using not one, but many approaches to compare models on different datasets, the paper avoids being robust to one technique and one metric.

Weaknesses and comments:

1. On page 4, when defining dilation, the usage of the terms "convolution" and "receptive field" seems a bit out of place. For this reason, I suggest defining those terms in the same paragraph instead of later in the paper.
2. The paper mentions the usage of the Wilcoxon test but does not specify which version, as there are two: one-tailed and two-tailed.
3. The methodology for handling datasets with unequal lengths, particularly the truncation method, is not clear. Can you elaborate more on this?
4. Regarding datasets with missing values, are there methods that proposed imputation algorithms to fix this issue? Were they tried?
5. The SharePriceIncrease dataset is missing a citation for the original Kaggle link.
6. Some feature-based methods can use frequency domain transformation, but what if the dataset is already in a frequency domain?
7. For the ShapeDTW measure, the paper mentions "The descriptors include slope, wavelet transforms, and piecewise approximations," but it's unclear which were used in the experiments.
8. In paragraph 1 of section 4.2.2, references are missing for all the mentioned tests.
9. The paper notes that sBoss was not included in the comparison of dictionary-based approaches due to technical challenges. Could you elaborate on these challenges?
10. In section 4.6, the paper discusses a correlation between convolution and shapelet but it's not clear how this is useful unless more elaboration or a figure is provided to analyze the correlation between both categories in terms of performance.
11. There is a contradiction where the paper criticizes most deep learners for evaluating on a subset of the UCR archive, but in the introduction, it states, "Secondly, it must have been evaluated on some subset of the UCR/UEA datasets." for the selection of algorithms.
12. The paper mentions that most deep learners evaluate on MTS data, but it's noted that deep learning models for time series are not always unified to univariate and multivariate; they are usually proposed for one of them.
13. The paper says, "We have not seen any algorithm that can realistically claim to outperform InceptionTime," then states H-InceptionTime outperforms InceptionTime, creating inconsistency. Also, it mentions "we restrict our attention to three deep learning algorithms" but five were used.
14. The title of section 4.7.1 should end with "Networks" instead of "Network."
15. On page 41, a key difference between ResNet and Inception related to the multiple convolution of different kernel sizes on each input is mentioned later when explaining the LITE architecture; this should be corrected for consistency.
16. In section 4.7.6 on page 42, did you mean "Litetime" instead of "LiteInception"?
17. On page 47, "IT" and "InceptionTime" are used when referring to the figure, but it should be "HInceptionTime," right?
18. A link to the numba package should be added in the footnote for completeness.
19. Adding deep learners to the time comparison may not be logical due to GPU training, but when comparing with the number of FLOPS, which is independent of the hardware, it should be considered separately from the training time comparison figure.
20. In the conclusion, the paper mentions, "It may be because the problems in the archives are relatively small compared to other archives used for deep learning evaluation." This observed phenomenon in Table 16, where H-InceptionTime goes from being ranked 6th to 3rd when train size is increased, should be correlated in the paper to support this conclusion.
21. The tool mentioned in the paper for heatmap generation has an associated paper as shown in the ReadMe of the GitHub, it could also be cited.

\subsubsection{Reviewer 3}
The review paper offers an extensive analysis of time series classification, expanding on the authors' previous review in 2017. It summarizes six years of progress in the field, discussing new methods, introducing datasets, and extending the analysis and comparison among recent methods. Similar to the previous work, this paper significantly contributes to the field and provides a thorough survey of recent developments. However, several comments, both minor and major, have been noted:

- As a continuation of the previous bake-off paper, this paper should mention other methods discussed in the earlier work in the hierarchy Figures (4,7,11,15,20), even if they are out of scope. Linking between the previously discussed methods and new methods would help establish a more comprehensive view.
- While the paper introduces an extended set of TSC algorithms, it overlooks some techniques that meet the criteria, such as [2,4]. Additionally, important concepts like motifs and discords should be mentioned.
- Although 1-NN DTW serves as a baseline, it is underperforming. Complementing it with the best-performing method from the previous bake-off paper for each category would enhance the analysis.
- The paper discusses different design pipelines for TSC algorithms, but further exploration of their advantages, disadvantages, and linkage with the results and analysis is needed.
- The analysis primarily focuses on accuracy, with train time mentioned in Figure 43 and Table 19. However, a more comprehensive analysis should include thorough time and space complexity evaluations.
- The results showed a comparison between methods for each category and then a comparison across categories with the best chosen from each one. Although that is a sensible approach to take, it can overlook the fact that multiple techniques of the same category can outperform the best in another one. Hence, that might introduce bias and inaccuracy in the results.
- The process for converting the four datasets from regression to classification lacks clarity regarding threshold and class determination.
- Many UCR datasets were created using a shape-based measure, potentially introducing bias. Including more realistic case studies without tuning to UCR settings would provide insights into algorithm performance in real scenarios.
- Parameters used for each method should be clearly stated in the appendix for transparency.
- While deep learning methods focus on CNN-based approaches, consideration should be given to RNN and LSTM-based methods as well, such as [4,5].
- The tool used for Figure 42 was not published at the time of the review, and additional information on how these numbers represent a fair comparison should be discussed.
- Adding methods that are across categories, such as [1,3], would enrich the survey.

Minor comments:
- The category of distance-based methods can be confusing; using the term "whole-series" or "shape-based" approach would clarify.
- An introduction of convolution filters on page 4 is recommended.
- Some parts of the paper and figures are repeated from the original bake-off paper such as Figure 9.
- Using Bold to highlight high performance in tables such as Table 3-12 would Enhance readability.
- More discussion on bias in newly introduced datasets and Figure 2 would aid in result interpretation.

[1] Alaee, S., Mercer, R., Kamgar, K., and Keogh, E. (2021). Time series motifs discovery under DTW allows more robust discovery of conserved structure. Data Mining and Knowledge Discovery, 35, 863-910.
[2] Karim, F., Majumdar, S., Darabi, H., and Chen, S. (2017). LSTM fully convolutional networks for time series classification. IEEE access, 6, 1662-1669.
[3] Wang, J., Yang, C., Jiang, X., and Wu, J. (2023, August). When: A wavelet-dtw hybrid attention network for heterogeneous time series analysis. In Proceedings of the 29th ACM SIGKDD Conference on Knowledge Discovery and Data Mining (pp. 2361-2373).
[4] Fawaz, H. I., Forestier, G., Weber, J., Idoumghar, L., and Muller, P. A. (2019, July). Deep neural network ensembles for time series classification. In 2019 International Joint Conference on Neural Networks (IJCNN) (pp. 1-6). IEEE.
[5] Lee, D., Lee, S., and Yu, H. (2021, May). Learnable dynamic temporal pooling for time series classification. In Proceedings of the AAAI Conference on Artificial Intelligence (Vol. 35, No. 9, pp. 8288-8296).

\subsubsection{Reviewer 4}
I do see value in “bake-offs” and Bagnall is the master. On that basis this paper may be useful.

However, I would like the authors to address the “O Rei vai nu”. As you note, there are dozens to hundreds of papers on time series classification. Each seems to produce better results, which is somewhat implausible (hence the utility of a bakeoff).
However, if you look at papers that actually do time series classification to solve a real problem (not just publish an incremental paper pushing up the average score), then 99\% of them just use simple nearest neighbor [a][b][c][d][e][f].
This needs to be explained. Some possibles..
1)      The apparent progress of sophisticated  time series classification algorithms is mostly an illusion caused by overfitting on UCR data.
2)      Perhaps people are not overfitting on UCR data, but UCR data is contrived in a way that it is a bad proxy for real problems that people try to solve.
3)      Maybe the sophisticated  time series classification algorithms really are better than NN with DTW. But “normal” researchers are too lazy or too dumb to figure out how to use them, and just use simple algorithms. If that was true, there is a real problem to solve, how to communicate the wonders of  sophisticated  algorithms to the general population.
4)      ???
Note that ‘1’ and ‘2’ are both consistent with one of your other papers in which you say.. “we conclude that 1-NN with Euclidean distance is fairly easy to beat but 1-NN with DTW is not, if window size is set through cross validation.”
I really do think you need to discuss address this, it is a stunning disconnect.

(SpatialBOSS is not included because of technical challenges). This makes little sense, if the current team of authors cannot overcome technical challenges of a BOSS variant, no one could.

“The SharePriceIncrease data was formatted as part of a student project”.
Can we have a more detailed acknowledgement of the students work?

we restrict our attention to three deep learning algorithms  TYPO?  I counted five

[a] On the Distribution of Muscle Signals: A Method for Distance-Based Classification of Human Gestures Jonas Große Sundrup,

[b] Ego-Network Transformer for Subsequence Classification in Time Series Data Chin-Chia Michael Yeh

[c] Mahato V.,. An evaluation of Data-driven gantry health monitoring and process status identification based on texture extraction [d]  Fan S.K.S., Hsu C.Y., Tsai D.M., He F., Cheng C.C. Data-driven approach for fault detection and diagnostic in semiconductor manufacturing [e]Carvalho B.G.Flow instability detection in offshore oil wells with multivariate time series machine [f]Neugebauer J., Improving cascade classifier precision by instance selection and outlier generation ICAART 2016

\subsection{Response to Reviewers}
\newcounter{comments}[section]

\newcommand{\rcomment}[1]
{
\stepcounter{comments}
\addcontentsline{toc}{subsection}{Reviewer \arabic{section} Comment \arabic{comments}}
\begin{tcolorbox}[colback=blue!5,colframe=white!45!black,title=Reviewer \arabic{section} Comment \arabic{comments}]
#1
\end{tcolorbox}
}

\section*{Reviewer 1}
\setcounter{section}{1}

\rcomment{
The paper is a revisit of the time series bake-off paper that overviews the current state-of-the-art in (univariate) time series classification (TSC). The new material includes algorithms that were published after the first bake-off, the introduction of the new 30 datasets, and the results of the new experiments. Code, data, and guide to reproduce the experiments are also provided.

Like its predecessor, this paper is a great effort by the authors to capture the current landscape of TSC research. I appreciate the enormous amount of work (data preparation, coding, experiments, etc.) that need to be done for this paper. I strongly believe that, like its predecessor, future research in this area will greatly benefit from this work.
}

\textbf{Response}

We would like to thank reviewer 1 for these positive comments and we hope the paper has the impact they predict.

\rcomment{“We contribute 30 new datasets”: If i understand correctly, some datasets are donated by the other researchers. Wording it this way can be misunderstood that the authors of the paper are also the authors of all 30 datasets.}

\textbf{Response}

Thank you, it was not our intention to imply this. We have reworded the text as follows

\begin{quote}
    \textcolor{blue}{
We release $30$ new univariate datasets donated by various researchers through the TSC GitHub repository\footnote{https://github.com/time-series-machine-learning/tsml-repo} and compare the best in category on these new datasets.
}
\end{quote}

\rcomment{“The SharePriceIncrease data was formatted as part of a student project”. Is it possible to acknowledge the student (e.g. with name/link) that led the project?}

\textbf{Response}

We have added an acknowledgement to the student in question. The data was collected in 2018 and I am no longer in contact with the student, so cannot provide a link to them. We have added a link to Kaggle in response to reviewer 2 comment 6.

\begin{quote}
    \textcolor{blue}{
The \textbf{SharePriceIncrease} data was formatted by Vladislavs Pazenuks as part of their 2018 undergraduate student project.
    }\\
\end{quote}

\rcomment{Figure 9 shows an example of spectro data where interval-based approaches may be superior. On the other hand, table 18 presents a different story where Quant (the best interval-based method) has the worst rank on the spectro column. While I agree that intuitively interval-based methods work best with phase dependent data, I don’t think this is a good example.}

\textbf{Response}

The ethanol level example we use is not meant to suggest interval based algorithms are the best approach for all spectrogram classification problems, which may have been formatted differently. They do however work well on the EthanolLevel problem. QUANT has an average accuracy of 82.71 \% as opposed to 81.55\% from HIVE-COTE 2.0 and 72.13\% from MultiRocket Hydra.

\rcomment{I also would like to add some thoughts on the classifying of the TSC algorithms. I find the categories (shaplet-based, feature-based, interval-based, etc.) proposed by the authors nicely capture the main characteristics of the algorithms. However, I would say some algorithms can cross the line between the categories. For example MrSQM/MrSEQL are shapelet methods but they are also very similar to dictionary-based methods (BOSS, WEASEL etc.) in that they also discretise the time series. The only difference is that , instead of frequency (histogram), they use binary occurrence information. Some classifiers also make use of feature-based techniques after transforming the time series (e.g. Multi Rocket early version tried catch22 features if I remember correctly). The algorithms across the categories also share some mechanisms (dilation, first difference, fourier transform, multi domain, multi view, multi resolution etc.). In this regard, I would love to learn the opinions of the authors on what seem to be the most effective ideas/themes in the current state of research and what would be the potential directions going forward in the area.}

\textbf{Response}

Thank you for these comments. There is indeed overlap between categories and any taxonomy is always open to debate. The goal of the categories was to capture the type of discriminatory features used rather than the nature of the algorithm, but it is far from perfect. It does allow us to structure the presentation, rather than presenting results of 40 classifiers at once.

As of now, a prevalent trend is to use dilation, a randomized ensemble of hyper-parameters, first order differences, and the implementation of a linear RIDGE classifier. The key benefit of employing these techniques is their ability to achieve high accuracy while maintaining low training time. However, one drawback associated with these approaches is the memory footprint of the feature space. Notably, the top-performing methods, such as the ROCKET family of approaches, WEASEL 2.0, and RDST, leverage these concepts to their advantage.

In terms of future directions for classification algorithm design, our opinions of the authors may differ. Tony thinks that the scope for improved general accuracy is narrowing, although (perhaps unsurprisingly) he considers there is scope for further heterogeneous ensembles. There is also always the possibility that a deep learning approach will outperform the existing state of the art. He believes that there is more scope for progress with more complex and realistic time series use cases. High dimensional, unequal length, unequal sampling and streaming series classification are not, in his opinion, well understood.

We have expanded the conclusion with the following text:

\begin{quote}
    \textcolor{blue}{Since the original bake off a number of trends have developed in research, and some prior observations remain true. On average, hybrid algorithms still perform better than single domain approaches on the UCR archive.}
    The ROCKET and HIVE-COTE family of classifiers work well because they combine convolution/shapelet approaches with dictionary based ones, i.e. they look for the presence of or the frequency of subseries. A key component of ROCKET based classifiers is dilation,
    We have shown that using dilation has significantly improved the single representation classifiers RDST and WEASEL 2.0. Incorporation of dilation could well benefit other algorithms, such as interval based classifiers.
    \textcolor{blue}{Ensemble algorithms are still effective and popular, but pipeline algorithms combining a transformation with a linear classifier such as ridge regression have shown to be just as competitive. The algorithms using linear classifiers such as ROCKET, WEASEL and RDST have shown to be more scalable than ensembles generally, but cannot produce good probability estimates. More algorithms have begun to incorporate transformed series such as first-order differences and periodograms into their feature extraction. This has been shown to increase accuracy in the majority of the algorithm types we have presented.}
\end{quote}

\rcomment{A table similar to table 3 in the first bake-off would be nice.}

We have included a table listing algorithm characteristics in Appendix B and added the following text in Section 4

\begin{quote}
    \textcolor{blue}{
To try and capture the commonality and differences between algorithms we provide a Table in the appendix B (Table B3) that groups algorithms by whether they employ the following design characteristics: dilation; discretisation; differences/derivatives; frequency domain; ensemble; and linear classification.}
\end{quote}

\rcomment{I think the name “ensemble” is better at capturing the characteristics of the last group than “hybrid”.}

Thanks, we thought long about the names. The problem is most of the classifiers (e.g. BOSS, STC and InceptionTime) are also ensembles. The name "hybrid" is meant to capture that the classifier employs algorithms from different domains. HC2 and TS-CHIEF are different types of ensemble: HC2 is heterogeneous whereas TS-CHIEF is homogeneous. You could argue MR-Hydra is a hybrid, since it employs dictionaries and convolutions. However, that would have put it up against HC2  meaning it would not figure in best of class experiments. We felt that was not representative, and hence included it under convolution based.

The pattern of results is broadly the same, and we need to be careful about over interpreting the actual order of ranks. The general conclusion is that HC2 and MR-Hydra perform the same on average. We have added MCM diagrams for the 30 new datasets and 112 original datasets in Appendix C. These diagrams take up a lot of space, so including more for non-best of class would lengthen an already long paper.

\newpage
\section*{Reviewer 2}
\setcounter{section}{2}
\setcounter{comments}{0}

\rcomment
{
This paper addresses the task of time series classification by presenting a range of algorithms proposed in recent years. These algorithms are categorized from distance-based approaches to hybrid ensemble models. Extensive experiments compare the algorithms within each category and also among the category winners. The paper uses the UCR archive for many TSC applications and introduces new datasets to the field, including corrected versions of unused UCR datasets, adapted versions from regression datasets, and entirely new ones.

Strengths of the paper:

1. Despite the dense accumulation of information, the paper is very well written, making it simple and easy to follow.

2. An extensive number of experiments have been conducted with different resampling on the datasets to ensure a general conclusion about which algorithms are to be used in given applications for future TSC datasets.

3. A detailed statistical and efficiency comparison between models is proposed, adding depth to the analysis.

4. The organization of the paper is very good, aiding in the comprehension of complex material.

5. By using not one, but many approaches to compare models on different datasets, the paper avoids being robust to one technique and one metric.
}

\textbf{Response.} Thank you for identifying these strengths of the paper.

\rcomment{On page 4, when defining dilation, the usage of the terms "convolution" and "receptive field" seems a bit out of place. For this reason, I suggest defining those terms in the same paragraph instead of later in the paper.}

\textbf{Response.} We have moved this introduction to Section 2 as suggested.

\rcomment{The paper mentions the usage of the Wilcoxon test but does not specify which version, as there are two: one-tailed and two-tailed.}

\textbf{Response.}
We use a one sided test. Text changed to

\begin{quote}
    \textcolor{blue}{
We perform pairwise one-sided Wilcoxon signed-rank tests
}
\end{quote}

\newpage
\rcomment{The methodology for handling datasets with unequal lengths, particularly the truncation method, is not clear. Can you elaborate more on this?}

\textbf{Response}

Thank you for pointing this out, the text is actually incorrect. On further investigation, we realised that we did not in fact truncate the series. Some problems have some very short instances (e.g. AllGestureWiimote have cases with just two observations), so truncation to the shortest length would have made the problems meaningless. Instead, we took the approach adopted to produce results for the UCR archive homepage: we pad with the series mean, with added low level Gaussian noise. This is apparent in Figure 3, and we should have noticed. Thank you for highlighting this.

\begin{quote}
    \textcolor{blue}{
Datasets with the suffix \textbf{\_eq} are unequal length series made equal length through padding with the series mean perturbed by low level Gaussian noise.
}
\end{quote}

\rcomment{Regarding datasets with missing values, are there methods that proposed imputation algorithms to fix this issue? Were they tried?}

\textbf{Response}

None of the implementations we used had the capability to handle missing values in time series, and we do not think any of the corresponding publications contained methods for handling missing cases either. We could of course impute missing values. However, the proportion of cases with missing values was low. We feel that simply removing these cases is the most transparent way of handling these datasets. We altered the Section 3.1 text to clarify this:

\begin{quote}
    Four data sets with the suffix \textbf{\_nmv} (no missing values) are datasets where the original contains missing values. These are also from the current archive.
    \textcolor{blue}{We have used the simplest method for processing the data, and}
      removed any cases which contain missing values for these problems (DodgerLoop variants and MelbournePedestrian).
      \textcolor{blue}{The number of cases removed per dataset amounts to 5-15\% of the original size for all four datasets which we deemed acceptable. While there have been imputation methods proposed for time series, the amount of missing values present and their pattern varies. The DodgerLoop datasets have large strings of missing values, while MelbournePedestrian has singular values or small groupings of missing data. }\\
\end{quote}

\rcomment{The SharePriceIncrease dataset is missing a citation for the original Kaggle link.}

\textbf{Response}

We have added a link and acknowledged the student who formatted the data (see reviewer 1 comment 2 response).

\rcomment{Some feature-based methods can use frequency domain transformation, but what if the dataset is already in a frequency domain?}

\textbf{Response}

Taking the Fourier transform twice $\textit{fft}(\textit{fft}(x))$ is equivalent to time inversion, you get $x(-t)$. Those classifiers using both time and frequency domain in their representation, will always capture both, regardless of the input being time or frequency domain.

\rcomment{For the ShapeDTW measure, the paper mentions "The descriptors include slope, wavelet transforms, and piecewise approximations," but it's unclear which were used in the experiments.}

\textbf{Response}

Based on the results from the original paper, we used the raw subsequences and derivative subsequences. Text adjusted to reflect this.

\rcomment{In paragraph 1 of section 4.2.2, references are missing for all the mentioned tests.}

\textbf{Response}

We have added references to these in Section 4.2.2.

\rcomment{The paper notes that sBoss was not included in the comparison of dictionary-based approaches due to technical challenges. Could you elaborate on these challenges?.}

\textbf{Response}

S-BOSS could not complete the full 112 run due to both runtime and memory constraints. Similar to HCTSA and Catch22 from the feature-based section, mentioning it is important for later algorithms which build on it. We would not recommend using it over more recent advances, and feel that comparing them over the full 112 dataset group is more useful. We have expanded the text to provide this reasoning:

\begin{quote}
    \textcolor{blue}{
    SpatialBOSS is not included due to its significant runtime and memory requirements which would require the exclusion of multiple datasets. We believe that comparing more recent advances on the full archive is more valuable than its inclusion, and suggest those interested in SpatialBOSS view the results presented in  Middlehurst et al (2020) which show it is comparable to WEASEL 1.0 in performance.
    }
\end{quote}

\rcomment{In section 4.6, the paper discusses a correlation between convolution and shapelet but it's not clear how this is useful unless more elaboration or a figure is provided to analyze the correlation between both categories in terms of performance.}

\textbf{Response}

The results indicate that there is significant variation in classifier performance between the shapelet and convolution classes. We have added the following text.

\begin{quote}
Shapelets can be realised through a convolution operation, followed by a min-pooling operation on the array of windowed Euclidean distances. This was first observed by [x].    \textcolor{blue}{However, despite this methodological connection, there is significant difference in the results obtained by convolution based and shapelet based approach, as illustrated in the Appendix C, Figure~C5. For example, the ROCKET results are negatively correlated with shapelet based approaches such as STC or RDST.} The main difference between convolutions and shapelets is that shapelets are subseries from the training data whereas convolutions are found from the entire space of possible real-values.
\end{quote}

\rcomment{There is a contradiction where the paper criticizes most deep learners for evaluating on a subset of the UCR archive, but in the introduction, it states, "Secondly, it must have been evaluated on some subset of the UCR/UEA datasets." for the selection of algorithms.}

\textbf{Response}

Giving a clear reason for selecting a subset of datasets (i.e. no capability to handle unequal length or scalability) is acceptable in our opinion. An issue with many of these publications is that they do not present a rationale for selecting datasets, which cannot help but raise the suspicion of cherry-picking. We have updated the introduction text to clarify this:

\begin{quote}
    Secondly, it must have been evaluated on
    \textcolor{blue}{one of the UCR/UEA dataset releases, or on a subset thereof, with reasoning provided for any datasets that are missing.}
\end{quote}

\rcomment{The paper mentions that most deep learners evaluate on MTS data, but it's noted that deep learning models for time series are not always unified to univariate and multivariate; they are usually proposed for one of them.}

\textbf{Response}

Thank you for highlighting this. We have made it clearer that multivariate is beyond the scope of this paper.

\begin{quote}
Most are evaluated only on the multivariate archive. Whilst cherry-picking data is questionable, using just MTSC data is not,  \textcolor{blue}{since deep learning classifiers are usually proposed specifically for MTSC}. However, it puts them beyond the scope of this paper.
\end{quote}

\rcomment{The paper says, "We have not seen any algorithm that can realistically claim to outperform InceptionTime," then states H-InceptionTime outperforms InceptionTime, creating inconsistency. Also, it mentions "we restrict our attention to three deep learning algorithms" but five were used.}

\textbf{Response}

We have fixed the incorrect number of algorithms, thanks. Our statement has been expanded to include H-InceptionTime:

\begin{quote}
    We have not seen any algorithm that can realistically claim to outperform InceptionTime (Fawaz et al, 2020),
    \textcolor{blue}{nor its successor H-InceptionTime (Ismail-Fawaz et al, 2022a).}
\end{quote}

\rcomment{The title of section 4.7.1 should end with "Networks" instead of "Network"
.}

\textbf{Response}

We have fixed this, thanks.

\rcomment{On page 41, a key difference between ResNet and Inception related to the multiple convolution of different kernel sizes on each input is mentioned later when explaining the LITE architecture; this should be corrected for consistency.}

\textbf{Response}

We have fixed this, thanks. Our description of InceptionTime has been extended to include multiplexing convolutions.

\begin{quote}
    It then applies multiple convolutional filters of varying kernel sizes, termed multiplexing
convolution, to capture temporal features at different scales. Key design differences to ResNet are ensembling of models, the use of bottleneck layers, multiplexing convolution using varying kernel sizes, and the use of only two residual blocks, as opposed to three in ResNet.
\end{quote}

\rcomment{In section 4.7.6 on page 42, did you mean "Litetime" instead of "LiteInception}

\textbf{Response}

We have fixed this, thanks.

\rcomment{On page 47, "IT" and "InceptionTime" are used when referring to the figure, but it should be "HInceptionTime," right?}

\textbf{Response}

We have fixed this, thanks.

\rcomment{A link to the numba package should be added in the footnote for completeness.}

\textbf{Response}

Done.

\rcomment{Adding deep learners to the time comparison may not be logical due to GPU training, but when comparing with the number of FLOPS, which is independent of the hardware, it should be considered separately from the training time comparison figure.}

\textbf{Response}

This is a good idea, and we were not aware of it. We have not captured this information in any of our deep learning experiments, which took a very long time to run. In future experiments we will make sure we record the number of FLOPS to make comparison more sensible.

\rcomment{In the conclusion, the paper mentions, "It may be because the problems in the archives are relatively small compared to other archives used for deep learning evaluation." This observed phenomenon in Table 16, where H-InceptionTime goes from being ranked 6th to 3rd when train size is increased, should be correlated in the paper to support this conclusion.}

\textbf{Response}

Thank you, we had not noticed this. We have added the following text to the conclusions

\begin{quote}
It may be because the problems in the archives are relatively small compared to other archives used for deep learning evaluation: \textcolor{blue}{Table~16 shows that H-InceptionTime improves relative to other algorithms as the number of training cases increases.}
\end{quote}

\rcomment{The tool mentioned in the paper for heatmap generation has an associated paper as shown in the ReadMe of the GitHub, it could also be cited.}

\textbf{Response}

Done.

\newpage
\section*{Reviewer 3}
\setcounter{section}{3}
\setcounter{comments}{0}

\rcomment{
The review paper offers an extensive analysis of time series classification, expanding on the authors' previous review in 2017. It summarizes six years of progress in the field, discussing new methods, introducing datasets, and extending the analysis and comparison among recent methods. Similar to the previous work, this paper significantly contributes to the field and provides a thorough survey of recent developments.
}

\textbf{Response}

Thank you for this summary.

\rcomment{As a continuation of the previous bake-off paper, this paper should mention other methods discussed in the earlier work in the hierarchy Figures (4,7,11,15,20), even if they are out of scope. Linking between the previously discussed methods and new methods would help establish a more comprehensive view.}

\textbf{Response}

We have updated Figures 5, 12, 16, 21 and 33 to include approaches from the previous bake off and described this in the captions.

\rcomment{While the paper introduces an extended set of TSC algorithms, it overlooks some techniques that meet the criteria, such as [2,4]. Additionally, important concepts like motifs and discords should be mentioned.}

\textbf{Response}

Thank you for identifying these papers. We considered [2], but we are not convinced as to their experimental regime: there seems to be leakage between train and test and we were not able to reproduce their results. We know of another research group who did the same. Furthermore, even their published results have already been shown to not be competitive in the deep learning bake off (Fawaz et al.). We have talked to one of the authors of [4] and they agree it is an early version of H-InceptionTime and they did not think we should include it. Motifs and discords are  very important time series primitives, but we are not aware of their successful application in classification. The paper is already very long, so we would respectfully decline from including even more background.

\rcomment{Although 1-NN DTW serves as a baseline, it is under performing. Complementing it with the best-performing method from the previous bake-off paper for each category would enhance the analysis.}

\textbf{Response}

We have in fact already done this by category, but obviously we have not made it clear enough. We have added a sentence in the introduction to clarify this
\begin{quote}
We describe the latest TSC algorithms included in this bake off in Section 4. This section also describes the first set of experiments that link to the previous bake off\textcolor{blue}{: for each category of algorithms we compare the latest classifiers with the best in class from Bagnall et. al. (2017)}
\end{quote}

\rcomment{The paper discusses different design pipelines for TSC algorithms, but further exploration of their advantages, disadvantages, and linkage with the results and analysis is needed.}

\textbf{Response}

\begin{figure}[h]
    \centering
   \includegraphics[width=0.6\columnwidth]{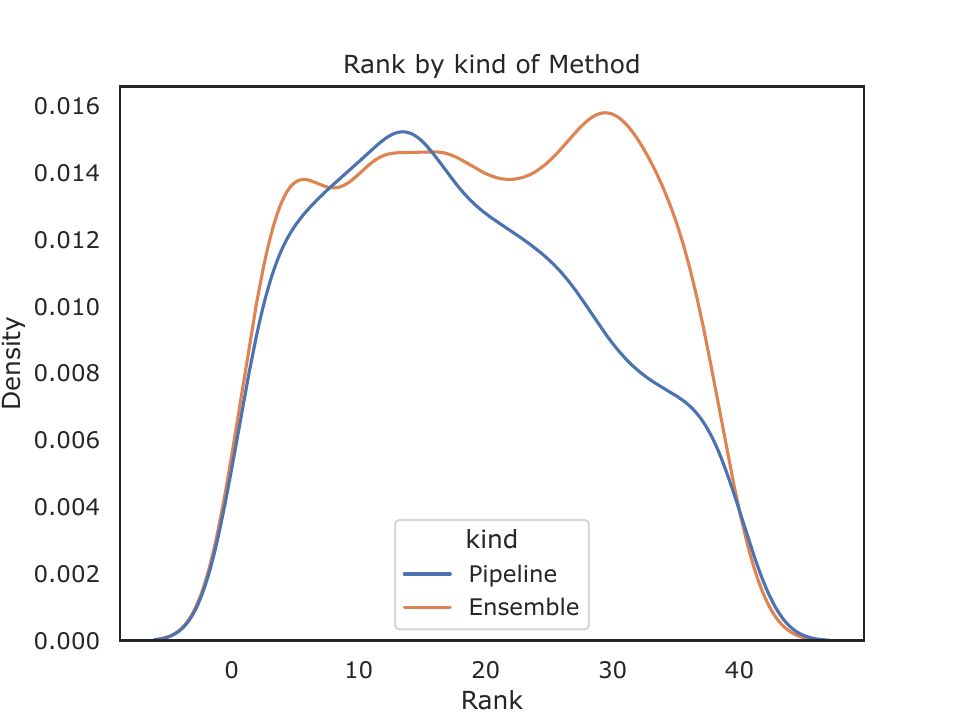}
    \caption{Distribution of ranks of pipeline vs ensemble based classifiers over all datasets.}
    \label{fig:pipeline_vs_ensemble}
\end{figure}

 Drawing general conclusions from comparing pipeline-based and ensemble-based approaches poses challenges. While there is a tendency for pipeline-based approaches to yield lower average ranks when compared (see Figure~\ref{fig:pipeline_vs_ensemble}), it's worth noting that the two top-performing approaches, MR-Hydra and HC2, originate from pipelines and ensembles, respectively. We have included a breakdown by the different factors in table B3 and also included two tables describing the correlation between estimators in the appendix (Figure C3 and C4).

We have added the following text to the analysis section.
\begin{quote}
    We have not seen any algorithm that can realistically claim to outperform InceptionTime (Fawaz et al, 2020).
    \textcolor{blue}{
We explored the effect on performance of the design decisions described in Table B3. If we group average accuracy ranks by each design factor of use of dilation, differences, ensemble, frequency domain and discretisation and perform a one factor ANOVA on each factor, we find a significant difference in rank distribution between those using dilation and those that do not, and those that use differences and those that do not. There was no significant difference in distribution when grouped by frequency, ensemble or discretisation. Care must be taken when interpreting these results since the assumptions behind the tests are not satisfied. However, there is at least some support for the utility of using dilation and differenced series.
 Finally, we have included a comprehensive correlation matrix on average accuracy ranks (see Figure C3 in the appendix). These demonstrate the diversity in performance of these classifiers and show the difference in performance between shapelet based and convolution based algorithms (FigureC4). }
 \end{quote}

\rcomment{The analysis primarily focuses on accuracy, with train time mentioned in Figure 43 and Table 19. However, a more comprehensive analysis should include thorough time and space complexity evaluations.}

\textbf{Response}

Our experiments include evaluations of accuracy, balanced accuracy, log loss, area under the ROC curve and train time. A worst-case complexity runtime analysis is not particularly useful for these classifiers. All of these algorithms have polynomial runtime, and are dominated by the actual machine learner used. Finally, the actual run time is often dependent on data driven stopping criteria. Some (e.g. STC) set run time as a parameter, some depend on the convergence of RIDGE regression. We think that summary measures of observed run time over all problems captures the differences. We agree that a better understanding of how these algorithms' memory scale with problem size would be very useful, but it is not trivial to store the maximum memory usage during a run on our HPC. We can now do so, but we were not able to at the time we ran these experiments. To do so would require we rerun all our experiments, which would take months. We hope the reviewer understands our reluctance to do this.

\rcomment{The results showed a comparison between methods for each category and then a comparison across categories with the best chosen from each one. Although that is a sensible approach to take, it can overlook the fact that multiple techniques of the same category can outperform the best in another one. Hence, that might introduce bias and inaccuracy in the results.}

\textbf{Response}

This is true, but we feel the need to be comprehensive needs to be balanced against the requirement to present results in a comprehensible way. However, in response to these completely valid reviewer comment we have included a new figure C1 in Appendix C which contains a critical difference diagram for all 40 algorithms. As you say, there are some categories (i.e. hybrids which take ranks 1, 4, 5 and 9) which perform better or worse than other categories generally. We include this line in Section 5 pointing to said appendix:

\begin{quote}
    \textcolor{blue}{Further results tables and figures are available in Appendix C, with all results files available on the accompanying website.}
\end{quote}

\rcomment{The process for converting the four datasets from regression to classification lacks clarity regarding threshold and class determination.}

\textbf{Response}

We have expanded the text introducing the discretised regression datasets:

\begin{quote}
    The four datasets ending with \textbf{\_disc} are taken from the TSER archive (Tan et al,
    2021). The continuous response variable was discretised manually for each dataset, the original continuous labels and new class values for each dataset are shown in Figure 2.
    \textcolor{blue}{Both Covid3Month and FloodModeling2 had a minimum label value with many cases. For both of these, this minimum label value has been converted into its own class label. For problems where there are no obvious places where the label can be separated into classes by value (including Covid3Month where the value is greater than 0), a split point was manually selected taking into account the average label value and and the number of cases in each class for a splitting point.}
\end{quote}

\rcomment{Many UCR datasets were created using a shape-based measure, potentially introducing bias. Including more realistic case studies without tuning to UCR settings would provide insights into algorithm performance in real scenarios.}

\textbf{Response}

We acknowledge the weaknesses of the UCR datasets (see response to reviewer 4) and to help mitigate against this we have introduced 30 new data. Case studies are great, but are more useful to highlight the strengths and weaknesses of a particular algorithm on a problem class. In depth case studies would massively lengthen the paper even if we restrict to the eight best in class and we feel for this type of paper it would not add enough to merit the extra pages. We have expanded our characterisation of the datasets to section 3. We also observe that the shape/distance-based group of algorithms we present results for arguably perform the worst out of all groups. Hence the bias introduced by the selection of UCR data may be less important than thought.

\rcomment{Parameters used for each method should be clearly stated in the appendix for transparency.}

\textbf{Response}

This is a good suggestion and we have created a table listing the parameters used for each algorithm. However, given that many algorithms have multiple lines of parameters that would have to be listed, we have elected to put this on the accompanying webpage. Tables such as B2 and B3 take a whole page even with a single line per algorithm, so this table would likely span for multiple pages with all 40 algorithms. We have included the following text in Section 3.2 pointing to the table:

\begin{quote}
    Further guidance on reproducibility,
    \textcolor{blue}{parameterisation of the algorithms used in our experiments}
    and our results files are available in an accompanying webpage.
\end{quote}

\rcomment{While deep learning methods focus on CNN-based approaches, consideration should be given to RNN and LSTM-based methods as well, such as [4,5].}

\textbf{Response}

\begin{figure}[h]
    \centering
   \includegraphics[width=1.0\columnwidth]{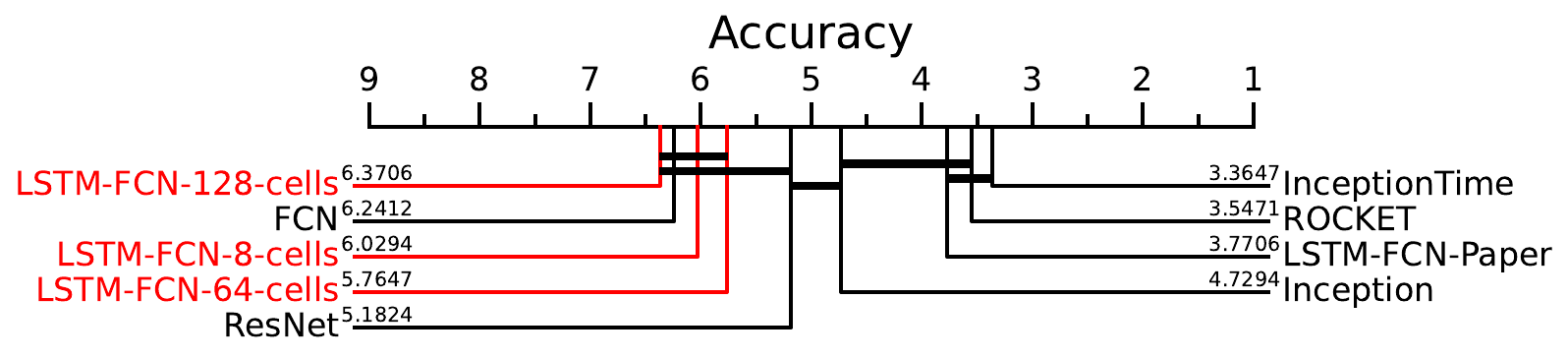}
    \caption{LSTM-FCN rerun.}
    \label{fig:lstm-fcn}
\end{figure}

Thank you for providing the two references. The study referenced in [5] adopts a CNN-based methodology. However, the paper cited in [4], called LSTM-FCN, exhibits several drawbacks [a, b, c, d, e]. Notably, the LSTM layer is applied to the transposed time series, effectively resembling a Fully Connected Approach [a,e]. Furthermore, the authors employ a process of selecting the optimal model based on test data and manually adjusting hyper-parameters (number of cells) for each dataset. Upon re-evaluation (see Figure~\ref{fig:lstm-fcn}), it became evident that the LSTM-FCN study heavily overfit the test data, and its actual rank on accuracy is worse than ResNet. Consequently, we have opted to exclude this paper from our assessment.

We are keen to keep continuity in the field, and none of us directly research deep learning for time series classification. A highly cited 2019 DAMI paper considered many algorithms, including RNN and LSTM, and we build on their conclusions, which found them to perform poorly. We already compare 40 algorithms, double the number in the first bake off, and we need to draw the line somewhere.

\begin{enumerate}
    \item[(a)] \url{https://github.com/titu1994/LSTM-FCN/issues/4}
    \item[(b)] \url{https://github.com/titu1994/LSTM-FCN/issues/6}
    \item[(c)] \url{https://github.com/titu1994/LSTM-FCN/issues/7}
    \item[(d)] \url{https://github.com/titu1994/LSTM-FCN/issues/9}
    \item[(e)] \url{https://github.com/titu1994/LSTM-FCN/issues/11}
\end{enumerate}

\rcomment{The tool used for Figure 42 was not published at the time of the review, and additional information on how these numbers represent a fair comparison should be discussed.}

\textbf{Response}

Thank you very much. We have extended the description within the paper to highlight the utility of the pairwise comparison tool:

\begin{quote}
    \textcolor{blue}{
    Ismail-Fawaz et al (2023a) noted that critical difference diagrams (CD) can be deceptive and lack stability, with the relative ordering being highly sensitive to the selection of comparates included in the comparisons. This sensitivity renders them susceptible to inadvertent manipulation. To circumvent this problem, they propose a bespoke pairwise comparison tool, called multiple comparative matrix (MCM). It shows pairwise comparisons between all comparates, and includes difference in average scores, wins/draws/losses, and Wilcoxon p-values. Colors of the heat map represent mean differences in scores. Red indicates that the comparate in the row wins by more on average than the comparate in the column. Bold text indicates that the difference in significant. Figure 42 summarises the performance of the eight classifiers using the MCM, with comparisons to the 30 new datasets and original 112 datasets available in Appendix C. A notable observation arises when comparing the CD on accuracy in Figure 38 to this MCM on the 112 UCR UTSC. The rankings of WEASEL 2.0, QUANT, H-IT and RDST are deceptive. Despite WEASEL 2.0 demonstrating more pairwise wins compared to RDST or QUANT in the MCM, its ranking appears higher (worse) than both in the CD when all 8 comparates are taken into account. In addition, H-IT has less pairwise wins than RDST in the MCM, yet shows the lower (better) rank in the CD.
    }
\end{quote}

\rcomment{Adding methods that are across categories, such as [1,3], would enrich the survey.}

\textbf{Response}

Whilst we aim to be comprehensive, we need to retain some focus on topic. To cover techniques not directly used in classification properly would take a lot more space, and the paper is already very long (currently 75 pages).

\rcomment{The category of distance-based methods can be confusing; using the term "whole-series" or "shape-based" approach would clarify.}

\textbf{Response}

We can see the source of confusion, distances are used on sub series by many algorithms, but no taxonomy is perfect and is open to debate. The term whole-series was used in the original bake off, but we feel it does not fully capture the category: feature pipelines also use the whole series. In fact, we could argue all classifiers use the whole series, even interval based ones (over members of the ensemble). We think the fact that they use distance functions over the whole series is the defining characteristic.

We have altered the overview text

\begin{quote}
Distance based: classification is based on some time series specific distance measure between     \textcolor{blue}{whole series
}
\end{quote}
and here

\begin{quote}
\textit{Distance based} classifiers use a distance function to measure the similarity between     \textcolor{blue}{whole time series}.
\end{quote}

\rcomment{An introduction of convolution filters on page 4 is recommended.}

\textbf{Response}

We have moved this introduction to Section 2 as suggested.

\rcomment{Some parts of the paper and figures are repeated from the original bake off paper such as Figure 9.}

\textbf{Response}

Whilst there is some similarity, we feel it is warranted in the name of continuity. Given that this is a direct follow-up from the 2017 bake off, there are inevitably going to be parts which are similar and figures which are still relevant. Just as adding our goal for this is as an introduction to current TSC research, and we do not want the previous bake off to be necessary reading to understand this paper. Figure 9 in-particular is still a good example to introduce the concept of interval-based classifiers.

\rcomment{Using Bold to highlight high performance in tables such as Table 3-12 would Enhance readability.}

\textbf{Response}

Done.

\rcomment{More discussion on bias in newly introduced datasets and Figure 2 would aid in result interpretation.}

\textbf{Response}

Thank you for highlighting this. We have included this new paragraph along Figure 3, which shows the characteristics of the 30 new datasets when compared to the existing 112 UCR UTSC datasets.

\begin{quote}
    \textcolor{blue}{
    Figure 3 shows the characteristics of the 30 new datasets when compared to the existing 112 UCR UTSC datasets, across different dimensions including length, train set size, number of classes, and data type. Findings reveal that the new datasets exhibit a broader range of lengths compared to old ones, while showing similar train set size and similar number of classes. It is worth noting that there seems to be a slight bias towards datasets derived from sensor and motion data in the new collection, whereas the majority of older datasets are sourced from the domain of image outlines.
    }
\end{quote}

\newpage
\section*{Reviewer 4}
\setcounter{section}{4}
\setcounter{comments}{0}

\rcomment{I do see value in “bake-offs” and Bagnall is the master. On that basis this paper may be useful.}

\textbf{Response}

Thank you, I will quote that!

\rcomment{However, I would like the authors to address the “O Rei vai nu”. As you note, there are dozens to hundreds of papers on time series classification. Each seems to produce better results, which is somewhat implausible (hence the utility of a bakeoff).
However, if you look at papers that actually do time series classification to solve a real problem (not just publish an incremental paper pushing up the average score), then 99\% of them just use simple nearest neighbor [a][b][c][d][e][f].
This needs to be explained. Some possibles..
1)      The apparent progress of sophisticated  time series classification algorithms is mostly an illusion caused by overfitting on UCR data.
2)      Perhaps people are not overfitting on UCR data, but UCR data is contrived in a way that it is a bad proxy for real problems that people try to solve.
3)      Maybe the sophisticated  time series classification algorithms really are better than NN with DTW. But “normal” researchers are too lazy or too dumb to figure out how to use them, and just use simple algorithms. If that was true, there is a real problem to solve, how to communicate the wonders of  sophisticated  algorithms to the general population.
4)      ???
Note that ‘1’ and ‘2’ are both consistent with one of your other papers in which you say.. “we conclude that 1-NN with Euclidean distance is fairly easy to beat but 1-NN with DTW is not, if window size is set through cross validation.”
I really do think you need to discuss address this, it is a stunning disconnect.

}

\textbf{Response by Tony}: Just to make clear this is my opinion, I cannot speak for the other two authors. Thank you appreciating bake offs and for the links for these papers. Perhaps the fact that so many are using NN classifiers is support for the need for our paper, although to be honest I have not observed this trend of only using 1NN-DTW. The majority of the papers you cite use 1NN in the context of traditional classification, perhaps in a pipeline with a transformation. From my experience [a,b,c,d,e], time series specific classifiers do add value to a range of problem domains.

(1) overfitting on the UCR: we acknowledge this risk and encourage the use of new problems. This is why we included 30 new problems in this paper and are constantly trying to expand the archive via tsc.com. The pattern of performance is the same on the new data. To overfit such a diverse and large collection of problems without explicit leakage between train and test is non trivial. You cannot memorise test instances if you do not see them. This is an argument for the robust and reproducible experimental regime we adopt. We also go to some lengths to be algorithm neutral and not unfairly promote our own contributions. This includes wrapping and evaluating MultiROCKET-Hydra even though the main contribution of their paper is the Hydra algorithm.

(2) UCR not fit for purpose: My standard response to this is to misquote Churchill: Using the UCR problems is the worst form of evaluation for TSC, except for all the others that have been tried. I believe that TSC research is much more thorough than many machine learning fields, where evaluation on twenty or less standard problems such as Iris is still common. Research is often a conversation, and our results are really only a starting point. More effort in understanding when and why a classifier is better is the way forward, but it is hard.

(3) "normal" researchers etc. In my experience [a,b,c,d,e] domain experts tend to just stick to the methods they know. For example, in food science they tend to always smooth spectra then use partial least squares. This may be the best option. However, if you do not even consider alternatives, you risk missing out on an easy improvement to your model. EEG research spans medicine, psychology and BCI. Each set of researchers have their own standard way of doing things, and largely ignore research in other related fields. Understandably, they focus more on the application and interpretation than the machine learning. However, they may be missing out on easy wins for improving their models. Many will still use it with 1-NN because the message is still out there, which is perfectly acceptable. I subscribe to Occam's razor, and always advocate benchmarking against the simplest approached first. However, they should be aware of and have access to the latest research. It improves their work and adds support for classifiers. This is why we provide open source implementations of the latest techniques. One of the main objectives of the aeon project is to make the latest research easily accessible.

Re {\em “we conclude that 1-NN with Euclidean distance is fairly easy to beat but 1-NN with DTW is not, if window size is set through cross validation.”}
I think this was true when I wrote it: in the original bake off only nine of 19 classifiers beat 1-NN DTW. However, in the latest bake off of 40 classifiers, only two failed to be significantly better than DTW 1-NN. The best classifiers are now over 12\% more accurate on average over all problems. The field has advanced. DTW is still a very useful tool, and distance based primitives such as matrix profile have a huge place in problems such as anomaly detection. But I think the evidence is there that DTW with 1-NN is not competitive for time series classification. I don't think DTW is the only tool we need, just as I don't believe that HIVE-COTE is the one model to rule them all. There are use cases for all algorithms. Logistic regression and C4.5 still have a place in the data scientists toolkit. However, more complex models do add value in most cases, and on average.

\begin{enumerate}
    \item [a] Large et al. Detecting forged alcohol non-invasively through vibrational spectroscopy and machine learning, PAKDD 2018
   \item [b] Bagnall et al. Detecting Electric Devices in 3D Images of Bags, ArXiv, 2020
   \item [c] Middlehurst, Detection of nuisance call centres using improved hybrid time series classification algorithms, PhD thesis
   \item [d] Flynn and Bagnall, Classifying flies based on reconstructed audio signals, IDEAL 2019
   \item [e] Rushbrook et al. Time Series Classification of Electroencephalography Data,. LNCS, 202
\end{enumerate}

\textbf{Response by Matthew}: For the most part I agree with the statement by Tony above. While there are issues with the UCR archive and potentially a disconnect between the cleaner data format used in research and the reality of data seen in a lot of use cases, I do not believe the progress made over the past years since the original bakeoff have been an illusion. Adding the capability for these algorithms to tackle more "realistic" problems such as unequal length/streaming and properly evaluating them is just another advancement to be made. I believe a lot of this is a visibility and usability problem similar to some of the points stated above, which is why we put effort into publications such as this to show what is out there. I think the truth is a lot of people will not bother to try a new (to them) approach unless it is low-effort or extremely likely to produce positive results in most cases. I can sympathise with this a bit, given the time investment to implement some of these algorithms and even run some researchers published code. This is the motivation for why myself and many other researchers contribute to toolkits such as aeon and tslearn, to provide implementations of these algorithms in a familiar format in a widely used language in the community.

\textbf{Response by Patrick:}
I believe that people not affine to ML use what is available in libaries such as sklearn. Better algorithms are not visible enough outside our community, or they are not easy to adapt. For example: [a] does not test anything but 1-NN DTW; [d] uses naive bayes and k-nn; and [f] uses classifiers available in sklearn (ocSVM, k-NN, SVM).

\rcomment{SpatialBOSS is not included because of technical challenges). This makes little sense, if the current team of authors cannot overcome technical challenges of a BOSS variant, no one could.}

\textbf{Response}

S-BOSS could not complete the full 112 run due to both runtime and memory constraints, and we feel that comparing recent advances over the full 112 dataset group is more useful. See reviewer 2 comment 10.

\rcomment{“The SharePriceIncrease data was formatted as part of a student project”.
Can we have a more detailed acknowledgement of the students work?}

\textbf{Response}

We have included better acknowledgement for their work. See reviewer 1 comment 3 and reviewer 2 comment 6.

\rcomment{we restrict our attention to three deep learning algorithms  TYPO?  I counted five}

\textbf{Response}

We have fixed this, thanks.

\subsection{Round2}

\subsubsection{Reviewer 4}
I like the formatting of your response.

“Response by Tony:” , “Response by Patrick:” “Response by Matthew:”
I have never seen this before, very strange.
I don’t think you really addressed my question, but I am not going to litigate this here.

“I think this was true when I wrote it: in the original bake off only nine of 19 classifiers beat 1-NN DTW. ..”.   The community polishes the heck out of complicated approaches, but 1-NN can be polished by generalizing to k-NN, smoothing the data, using prefix-suffix invarent DTW etc.  There is a disconnect between what is useful to practitioners and what will get you a paper in NIPS/KDD/ICML

“domain experts tend to just stick to the methods they know.” Perhaps they stick to what works.

\subsubsection{Reviewer 1}
Thank you for your response. I have no further comments.

\subsubsection{Reviewer 3}
The authors effectively addressed all comments with convincing answers, except for minor points that can be overlooked. We thank them for their extensive work presented in this paper.

\end{appendices}

\end{document}